\ifdefined\useorstyle
\documentclass{article}
\pdfoutput=1

\usepackage[numbers]{natbib} 
\usepackage{macros/nips_2024}
\usepackage{booktabs}

\usepackage{macros/statistics-macros}
\usepackage{macros/packages}
\usepackage{macros/editing-macros}
\usepackage{macros/formatting}
\usepackage{macros/statistics-macros}
\usepackage{caption}

\usepackage{algorithm}
\usepackage{algpseudocode}

\usepackage{endnotes}
\let\footnote=\endnote

\usepackage{xspace}

\usepackage{hyperref}

\usepackage{pgfplotstable}
\usepackage{graphicx}
\usepackage{float} 
\usepackage{tabularx}
\usepackage{overpic}
\usepackage{tikz}
\usepackage{rotating}
\usepackage{psfrag}
\usepackage{enumitem}
\usepackage{tikz}
\usetikzlibrary{shapes, arrows, positioning}  
\usepackage{algorithm}
\usepackage{algpseudocode}

\date{\today}

\maketitle

\else

\documentclass[11pt]{article}
\usepackage[numbers]{natbib}
\usepackage{macros/packages}
\usepackage{macros/editing-macros}
\usepackage{macros/formatting}
\usepackage{macros/statistics-macros}
 
\usepackage{algorithm}
\usepackage{algpseudocode}

\usepackage{subcaption}

\usepackage{amsmath}
\usepackage{amssymb}
\usepackage{tikz}
\usetikzlibrary{positioning, arrows.meta, shapes.geometric, calc, fit, backgrounds}
\usepackage[most]{tcolorbox}   
\usepackage{xcolor}            
\usepackage{xparse}            
\usepackage{tabularx}          
\usepackage{booktabs}          
\usepackage{array}             
\usepackage{multicol}          
\usepackage{graphicx}          
\usepackage{amsmath}           
\usepackage{amssymb}           
\usepackage{multirow}
\definecolor{FailRed}{RGB}{220,53,69}
\definecolor{HintOrange}{RGB}{245,158,11}
\definecolor{IgnoreGold}{RGB}{202,138,4}
\definecolor{ConfuseBlue}{RGB}{37,99,235}
\definecolor{WasteTeal}{RGB}{13,148,136}
\definecolor{WrongPurple}{RGB}{147,51,234}
\definecolor{EarlyMagenta}{RGB}{219,39,119}

\usepackage[T1]{fontenc}
\usepackage{listings}
\usepackage{xcolor}

\definecolor{pygblue}{RGB}{0,0,255}        
\definecolor{pyggreen}{RGB}{0,128,0}       
\definecolor{pygstring}{RGB}{186,33,33}    
\definecolor{pygcomment}{RGB}{64,128,128}  

\lstdefinestyle{lgpy}{
  language=Python,
  basicstyle=\scriptsize\ttfamily\color{black},
  breaklines=true,
  showstringspaces=false,
  columns=fullflexible,
  keepspaces=true,
  upquote=true,
  keywordstyle=\color{pyggreen}\bfseries,
  identifierstyle=\color{black},
  stringstyle=\color{pygstring},
  commentstyle=\color{pygcomment}\itshape,
  literate=%
    {latentgym.core.registry}{{\textcolor{pygblue}{latentgym.core.registry}}}{23}
    {.core\_env}{{\textcolor{pygblue}{.core\_env}}}{9}
}

\definecolor{codegreen}{rgb}{0.0,0.5,0.0}
\definecolor{codegray}{rgb}{0.5,0.5,0.5}
\definecolor{codepurple}{rgb}{0.55,0.0,0.55}
\definecolor{codeblue}{rgb}{0.0,0.0,0.7}
\definecolor{backcolour}{rgb}{0.96,0.96,0.96}

\lstdefinestyle{pystyle}{
    language=Python,
    backgroundcolor=\color{backcolour},
    commentstyle=\color{codegray}\itshape,
    keywordstyle=\color{codeblue}\bfseries,
    stringstyle=\color{codegreen},
    basicstyle=\ttfamily\footnotesize,
    breaklines=true,
    captionpos=b,
    keepspaces=true,
    showstringspaces=false,
    frame=single,
    framerule=0.4pt,
    rulecolor=\color{codegray},
    xleftmargin=8pt,
    xrightmargin=8pt,
    aboveskip=6pt,
    belowskip=6pt,
}

\definecolor{GPT4oColor}{RGB}{95, 115, 165}
\definecolor{GeminiColor}{RGB}{95, 115, 165}
\definecolor{ClaudeColor}{RGB}{95, 115, 165}
\definecolor{GPT5MiniColor}{RGB}{95, 115, 165}
\usepackage[T1]{fontenc}

\definecolor{AgentBg}{RGB}{235, 240, 250}      
\definecolor{AgentBorder}{RGB}{70, 105, 170}   
\definecolor{AgentTitle}{RGB}{30, 55, 115}     

\definecolor{GmBg}{RGB}{252, 247, 235}         
\definecolor{GmBorder}{RGB}{180, 130, 50}      
\definecolor{GmTitle}{RGB}{120, 80, 20}        

\definecolor{ThinkBg}{RGB}{246, 240, 248}      
\definecolor{ThinkBorder}{RGB}{140, 100, 160}  
\definecolor{ThinkTitle}{RGB}{75, 50, 100}     

\NewDocumentEnvironment{anecdotebox}{m m}{%
  \begin{tcolorbox}[
    enhanced,
    colback=#2!4,
    colframe=#2!72!black,
    coltitle=white,
    colbacktitle=#2!85!black,
    boxrule=0.7pt,
    arc=2.5pt,
    left=8pt, right=8pt, top=4pt, bottom=4pt,
    fonttitle=\bfseries\small,
    title={#1},
    breakable,
  ]%
  \sloppy\emergencystretch=3em\relax}{
   \end{tcolorbox}
}

\NewDocumentEnvironment{taskbox}{m m}{%
  \par\smallskip%
  \begin{tcolorbox}[
    enhanced,
    colback=white,
    colframe=#2!55!black,
    coltitle=white,
    colbacktitle=#2!70!black,
    boxrule=0.45pt,
    arc=1.8pt,
    left=4pt, right=4pt, top=2pt, bottom=2pt,
    fonttitle=\bfseries\footnotesize,
    title={#1},
    unbreakable,
  ]%
  \sloppy\emergencystretch=4em\relax
  \scriptsize}{
  \end{tcolorbox}%
}

\NewDocumentCommand{\agentmsg}{m m}{%
  \par\smallskip\noindent
  \begin{tcolorbox}[
    enhanced,
    colback=AgentBg,
    colframe=AgentBorder,
    boxrule=0pt,
    leftrule=2pt,
    arc=1.5pt,
    left=4pt, right=4pt, top=1.5pt, bottom=1.5pt,
    unbreakable,
  ]
  \textbf{\textcolor{AgentTitle}{Agent (turn #1):}}\space #2%
  \end{tcolorbox}%
  \par
}

\NewDocumentCommand{\gmmsg}{m}{%
  \par\smallskip\noindent
  \begin{tcolorbox}[
    enhanced,
    colback=GmBg,
    colframe=GmBorder,
    boxrule=0pt,
    leftrule=2pt,
    arc=1.5pt,
    left=4pt, right=4pt, top=1.5pt, bottom=1.5pt,
    unbreakable,
  ]
  \textbf{\textcolor{GmTitle}{Game Master:}}\space #1%
  \end{tcolorbox}%
  \par
}

\NewDocumentCommand{\thinkmsg}{m m}{%
  \par\smallskip\noindent
  \begin{tcolorbox}[
    enhanced,
    colback=ThinkBg,
    colframe=ThinkBorder,
    boxrule=0pt,
    leftrule=2pt,
    arc=1.5pt,
    left=4pt, right=4pt, top=1.5pt, bottom=1.5pt,
    unbreakable,
  ]
  \textbf{\textcolor{ThinkTitle}{Agent (thinking, turn #1):}}\space\textit{#2}%
  \end{tcolorbox}%
  \par
}

\definecolor{quotebg}{RGB}{240, 244, 248}      
\definecolor{quoteborder}{RGB}{52, 95, 140}    

\newtcolorbox{trajectory}{
  enhanced,
  breakable,
  colback=quotebg,
  colframe=quotebg,
  boxrule=0pt,
  arc=1pt,
  outer arc=1pt,
  left=10pt,
  right=8pt,
  top=6pt,
  bottom=6pt,
  borderline west={2.5pt}{0pt}{quoteborder},
  fontupper=\small
}

\newcommand{\titleFont}{\bfseries\fontfamily{qpl}\selectfont}
\newcommand{\authorFont}{\fontfamily{qpl}\selectfont}

\usepackage{xcolor}
\usepackage[most]{tcolorbox}
\usepackage{minted}

\definecolor{codebg}{RGB}{248,248,248}
\definecolor{codeframe}{RGB}{220,220,220}
\definecolor{codetitle}{RGB}{245,245,245}

\newtcblisting{pythonbox}[1]{
  listing engine=minted,
  minted language=python,
  minted options={
    fontsize=\footnotesize,
    breaklines,
    autogobble,
    linenos=false
  },
  listing only,
  title={\texttt{#1}},
  colback=codebg,
  colframe=codeframe,
  coltitle=black,
  colbacktitle=codetitle,
  fonttitle=\footnotesize\ttfamily,
  boxrule=0.4pt,
  arc=2pt,
  left=1.0em,
  right=1.0em,
  top=0.7em,
  bottom=0.7em,
  toptitle=0.35em,
  bottomtitle=0.35em,
}

\begin{document}




\abovedisplayskip=8pt plus0pt minus3pt
\belowdisplayskip=8pt plus0pt minus3pt







\definecolor{boxfill}{RGB}{244,246,254}
\definecolor{boxaccent}{RGB}{40, 55, 100}
\definecolor{titleblue}{HTML}{1a2a6c}

\begin{tcolorbox}[
  enhanced,
  colback=boxfill,
  colframe=boxaccent,
  boxrule=0.6pt,
  arc=6pt,
  left=18pt, right=18pt, top=14pt, bottom=12pt,
  breakable,
]

\begingroup
\centering


{\LARGE\titleFont\color{titleblue}
LatentGym: A Testbed For Cross-Task Experiential \\ Learning
With Controllable Latent Structure \par}
\vspace{0.8em}

{\normalsize \authorFont
\textbf{Daksh Mittal}\textsuperscript{*\,1}\quad
\textbf{Tommaso Castellani}\textsuperscript{*\,1}\quad
\textbf{Thomson Yen}\textsuperscript{*\,1}\\[3pt]
\textbf{Naimeng Ye}\textsuperscript{1}\quad
\textbf{Minghui Chen}\textsuperscript{1}\quad
\textbf{Tiffany Cai}\textsuperscript{1}\quad
\textbf{William Zeng}\textsuperscript{2}
\\[3pt]
\textbf{Emmanouil Koukoumidis}\textsuperscript{2}\quad
\textbf{Fangyu Wu}\textsuperscript{1}\quad
\textbf{Hongseok Namkoong}\textsuperscript{1}\par}
\vspace{0.8em}

{\normalsize \authorFont
\textsuperscript{1}Columbia University\quad
\textsuperscript{2}Oumi\par}
\vspace{0.5em}

{\normalsize\authorFont\hypersetup{urlcolor=titleblue}%
\href{https://dakshmittal30.github.io/LatentGym-Experiential-Learning/}{Blog} \hspace{0.3em} $|$ \hspace{0.3em}
\href{https://github.com/namkoong-lab/LatentGym}{Code} \hspace{0.3em} $|$ \hspace{0.3em}
\href{https://huggingface.co/collections/namkoong-lab/latentgym}{Models}}\\[0.5em]

\endgroup

{\color{boxaccent!50}\hrule height 0.3pt}
\vspace{0.7em}

\begingroup
\authorFont

We envision continually learning agentic systems that become more useful over time: as they encounter sequences of related tasks, they should infer the hidden structure shared across those tasks and use it to improve future decisions. This \emph{cross-task experiential learning} capability is pivotal in domains such as personalization and interactive assistance, but existing training/evaluation frameworks do not provide shared, controllable latent structures and cannot measure whether or why agents improve. We introduce \textsc{LatentGym}: a controllable suite in which each environment is organized around a ground-truth latent variable governing the structure across tasks. Our construction yields metrics that separate exploration (whether the agent's actions gather information about the latent) from exploitation (whether the agent uses what it has gathered). We demonstrate our suite on empirical studies addressing three questions: how and why frontier models fail to adapt across related tasks; whether post-training on related task sequences improves general cross-task adaptation, and where those gains come from; and how design choices such as inter-task feedback shape training dynamics and generalization. Together, these results establish a controlled foundation for studying how LLM agents learn from experience across tasks, and for designing agents that adapt more reliably in sequential, personalized, and interactive settings.

\par
\endgroup

\vspace{0.85em}

\begingroup
\small\authorFont
\noindent
\begin{minipage}[c]{0.68\linewidth}
\textbf{Correspondence:} \texttt{dm3766@columbia.edu}\\[0.5em]
\textsuperscript{*}Equal contribution
\end{minipage}\hfill
\begin{minipage}[c]{0.30\linewidth}
\raggedleft
\raisebox{-0.5\height}{\includegraphics[height=2.2em]{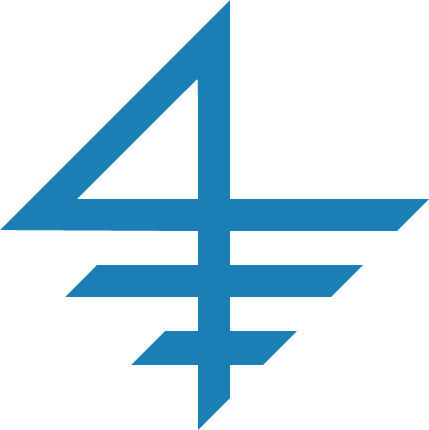}}\hspace{1.0em}
\raisebox{-0.5\height}{\includegraphics[height=2.2em]{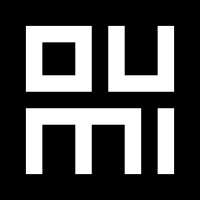}}\\[1.0em]
\end{minipage}
\endgroup

\vspace{0.3em}
\end{tcolorbox}

\vspace{1.0em}


\fi



\section{Introduction}
\label{sec:introduction}

We envision agentic systems that grow more capable the longer they work in a domain. An agent that encounters a sequence of related tasks, such as the tickets handled in customer support, the issues resolved across a software product, or the experiments run within a research area~\citep{prabhakar2025swebenchcl,lu2024aiscientist}, should improve with experience instead of solving each in isolation. It should pick up the shared structure across tasks, such as common knowledge, recurring patterns, and reusable strategies, and turn that experience into better decisions on the tasks that follow~\citep{silver2025experience}. We call this ability \emph{cross-task experiential learning}, the capacity to improve across a sequence of related tasks by recognizing what they share and acting on it. Humans do this routinely; whether today's LLM agents do, and how to build agents that do it reliably, remains largely open~\citep{yang2025lifelongagentbench,nie2025evolve,hu2026memory}.


Rigorously studying whether agents improve across tasks, and why they fail when they do, is hard today.  We can observe an agent's performance rise or fall over a sequence of tasks, but not easily say \emph{why} a given failure occurred: whether the sequence carried too little shared structure to exploit, whether the agent failed to recognize the structure that was there, or whether it recognized the structure but failed to act on it. Nor can we cleanly compare methods meant to build this capability, since we have no setting in which the structure to be learned is known and controllable, making it hard to tell which method improves which part of the capability. 
We draw inspiration from how small, controlled testbeds have repeatedly driven the field forward by stripping away confounders, so that methods can be compared cheaply and their mechanisms understood.  Toy environments such as 
MNIST and CIFAR in computer vision, and tabular gridworlds, control suites, and procedurally generated games in RL~\citep{cobbe2020procgen,yu2020metaworld,nikulin2023xlandminigrid} allow researchers to build understanding before moving to messier and costlier real-world settings. 

\begin{figure}[t]
    \centering
    \includegraphics[width=\linewidth]{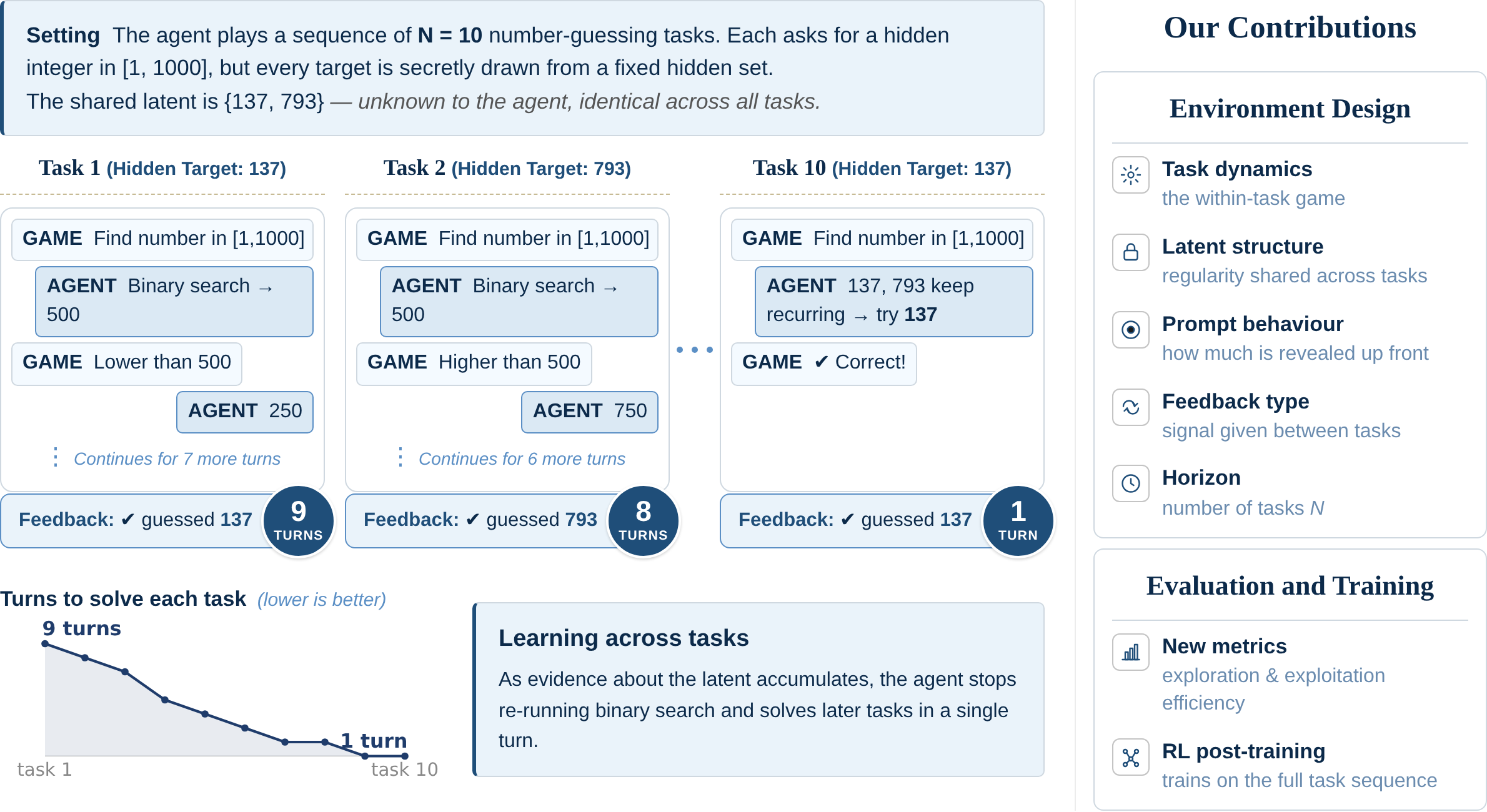}
    \caption{\textbf{Probing Cross-Task Adaptation in LLM Agents through Controllable Latents.}  \textbf{(Left)} Number-guessing illustration. The agent plays a sequence of $N{=}10$ tasks; each task asks for a hidden integer in $[1, 1000]$, but targets are in fact drawn from a fixed hidden set $z=\{137, 793\}$ (the latent shared across tasks). Early on (Task 1), the agent runs binary search and takes 9 turns. By Task 10, having observed that 137 and 793 recur, it solves the task in 1 turn by guessing one of the recurring values directly.  \textbf{(Right)} Each environment in our framework is built around such an explicit, controllable latent and factorizes along axes including task dynamics, prompt behavior, feedback type, and horizon . We use this design to study how frontier LLM agents adapt across tasks, introduce metrics for exploration and exploitation efficiency, and train via RL on full task sequences (Cross-Task RL), demonstrating that this approach can instill general cross-task adaptation.}
    \label{fig:abcd}
\end{figure}

We thus focus on small, fully controllable cross-task experiential learning problems where the experimenter knows the structure to be learned, runs are cheap enough to iterate on, and competing methods can be compared and understood. Such a testbed must make the structure shared across tasks both known to and controllable by the experimenter. We call this shared structure the \emph{latent}: without ground-truth knowledge of it, an apparent failure to adapt cannot be interpreted.  Figure~\ref{fig:abcd} makes this concrete. The agent plays a sequence of number-guessing tasks. Each task asks for a hidden integer said to lie in $[1,1000]$, but the targets are in fact drawn from a fixed hidden set $\{137,793\}$, which is the latent shared across the sequence. Early on the agent has little evidence about the latent, so a near-optimal strategy is binary search over $[1,1000]$. After a few tasks it can notice that the target keeps coming from a small set, and the later tasks test whether it acts on that evidence by guessing one of the recurring targets instead of restarting binary search. When the latent is known to the experimenter, they can see exactly where an agent's adaptation breaks down.
 
We realize this testbed as \textsc{LatentGym}, a single framework that brings together three components needed to  study of cross-task experiential learning (Section~\ref{sec:framework}): controllable environments, diagnostics, and an integrated RL pipeline as we outline below.

\paragraph{Controllable environments.} Each environment is organized around an explicit, controllable latent that defines the regularity shared across a sequence of tasks. Around this latent, each environment has a few additional independent knobs: the within-task dynamics, \emph{prompt type}
(how much information about the latent the prompt reveals), \emph{feedback type} (what feedback the agent
receives between tasks), and \emph{horizon} (how many tasks the agent sees). Together, these knobs
determine both how hard adaptation is and how it can be supported: the prompt sets the agent’s
prior over the latent and feedback governs how quickly evidence accumulates across tasks. Varying them in a controlled way lets us isolate the role of each, both as a difficulty knob at evaluation time and as a design variable at training time.

\paragraph{Diagnostics.}
Beyond measuring whether an agent improves over a sequence, \textsc{LatentGym} measures \emph{how} it adapts. Two agents with the same cumulative reward may behave very differently: one keeps probing the latent while the other exploits what it already knows, and the two call for different remedies. We therefore separate cross-task experiential learning into two quantities: \emph{Exploration efficiency} asks whether the agent's actions produce experience informative about the latent; \emph{exploitation efficiency} asks whether the agent turns that experience into better decisions on later tasks. An interactive dashboard lets us inspect the per-task reward curves and trajectories behind every run.


\paragraph{An integrated RL pipeline.} Environments are integrated with the standard RL training pipeline, so the same environments serves both evaluation and training. Our framework is directly compatible with SkyRL~\citep{cao2025skyrlagent}, which provides a range of standard algorithms out of the box, including SFT, PPO~\citep{schulman2017ppo}, GRPO~\citep{shao2024deepseekmath} etc. New advantage functions are easy to add: SkyRL runs the distributed rollouts and policy optimization, while we add only new estimators and a thin environment adapter rather than rewriting the full training stack. The pipeline also exposes the per-task rewards across a sequence, so advantage functions can be defined over the full sequence rather than a single task.

\vspace{1em}
Environments  (including latents, prompts, feedback) and RL algorithms each registers independently and composes freely; adding a new one requires no changes to the rest. \emph{Combining these into a single, extensible framework is the central contribution of this paper.} We instantiate \textsc{LatentGym} with \emph{seven environments} built on text-based games~\citep{guertler2025textarena}. Using it, we report three findings, each made possible by the controllable latent and each carrying a concrete lesson for how agents learn across tasks:


\begin{enumerate}
\item \textbf{A capability evaluation of frontier models.} We evaluate frontier models (Claude Sonnet 4.6~\citep{anthropic2026claudesonnet46}, GPT-4o~\citep{openai2024gpt4o}, Gemini 2.5 Flash~\citep{google2025gemini25}) and find that they can fail at cross-task adaptation even under simple latent structures (Section~\ref{sec:sota_failure_modes}). Our framework surfaces three failure modes: \emph{adaptation neglect} (failing to look for cross-task structure), \emph{adaptation breakdown} (recognizing structure but failing to act on it reliably), and \emph{adaptation miscalibration} (over-applying explicit information at the cost of task performance).

\item \textbf{Cross-Task RL induces general adaptation.} We show that reliable  adaptation can be induced by applying GRPO~\citep{shao2024deepseekmath} over full sequences of related tasks, a recipe we call \emph{Cross-Task RL} (Section~\ref{sec:experiments}). The gains transfer out of distribution to held-out latents and held-out environments, suggesting that part of what is learned is a general meta-strategy rather than latent or environment-specific cues. Further, Cross-Task RL improves both exploration and exploitation efficiency, though which axis dominates depends on the environment.

\item \textbf{Algorithmic design choices in Cross-Task RL.} We study how feedback type and prompt type during training shape Cross-Task RL (Section~\ref{sec:algorithmic}), and find, counterintuitively, that training under sparser task feedback transfers more robustly across deployment conditions than richer feedback.
\end{enumerate}

 We release \textsc{LatentGym} as open source, including its environments, diagnostics, RL training pipeline, and inspection dashboard. We hope it gives the community a shared, controllable laboratory for studying cross-task experiential learning and for building the agents that achieve it.

\paragraph{Related Work}

Our framework directly exposes controllable latent structure and enables a systematic study of \emph{cross-task} adaptation.  On the other hand, in-context learning~\citep{brown2020gpt3} and test-time scaling~\citep{shao2024deepseekmath,deepseekai2025r1,snell2025scaling}  largely consider a \emph{single task}: demonstrations, reasoning traces, or retrieved context help the model do better on the current query, and each query is treated independently. Existing agentic benchmarks~\citep{paglieri2024balrog,abdulhai2023lmrlgym,liu2023agentbench,jimenez2024swebench,zhou2024webarena} do not expose the shared structure that the experimenter can specify or vary across tasks. 
While recent work has begun to explore harness design~\citep{yao2023react,shinn2023reflexion,wang2023voyager}, meta-RL training, and continual learning for LLM agents~\citep{monea2025lamer, zheng2025lifelongllm, lin2026orbit,yang2025lifelongagentbench}, 
these settings still leave what should be learned across the sequence implicit and outside the experimenter's control. As a result, it remains difficult to diagnose where agents fail at cross-task 
adaptation and, consequently, how they should be trained to improve,
e.g., disambiguating the role of scaffolding vs. memory modules.

Our setup inherits its formal structure from meta-RL~\citep{beck2023survey}, particularly the in-context variants where adaptation occurs through a sequence model's context~\citep{duan2016rl2,wang2016learning,mishra2018snail,rakelly2019pearl,zintgraf2020varibad} rather than through gradient updates~\citep{finn2017maml}. A closely related line studies whether transformers can internalize an RL algorithm in their forward pass~\citep{laskin2023ad,lee2023dpt,grigsby2024amago}, by training on RL trajectories across many task families. 
Classical meta RL benchmarks consider shared latent structure~\citep{brockman2016gym,oel2021xland,bauer2023ada}, but they are non-linguistic and train small policies from scratch, saying little about pretrained LLM agents.  We bring this paradigm into language space like~\citet{monea2025lamer,lin2026orbit} and develop a testbed that allow studying questions unique to LLM agents. 

Another related line of work synthesizes simulators into testbeds for LLM
agents~\citep{cote2018textworld,shridhar2021alfworld,paglieri2024balrog,abdulhai2023lmrlgym,liu2025gem,liu2023agentbench,jimenez2024swebench,zhou2024webarena},
along with recent benchmarks targeting continual and experiential
learning~\citep{yang2025lifelongagentbench,prabhakar2025swebenchcl,zhao2024expel}. They
measure capability or experience reuse on diverse tasks, but without
exposing an explicit, controllable latent shared across tasks within a
single evaluation stream. The closest in spirit is
XLand-MiniGrid~\citep{nikulin2023xlandminigrid}, which factorizes meta-RL
environments along clean axes but in gridworlds rather than language.

\section{A Controllable Suite for Cross-Task Experiential Learning}
\label{sec:framework}

An agent in \textsc{LatentGym} faces a sequence of $N$ related tasks that share a hidden \emph{latent}, the regularity common to all of them. A latent can be a rule, a preference, a mapping, a constraint, a target set, or a temporal pattern. The agent never observes it and must infer it from interaction. Its weights stay fixed across the sequence, so it adapts in-context, carrying the history of earlier tasks forward and conditioning on it when acting, with no gradient updates. We score the per-task reward sequence $r_1,\dots,r_N$ and summarize it by the \emph{cumulative reward} $R=\sum_{i}r_i$; the final-task reward $r_N$ and the improvement $r_N-r_1$ isolate other facets of adaptation. Because the latent governs every task, an information-seeking action that lowers an early $r_i$ can raise many later ones, so the exploration--exploitation trade-off here runs \emph{across} tasks, where evidence gathered early pays off on every task that follows.

Each environment in our framework is built around an explicit, controllable latent. Around this central axis, each environment factorizes
along four further axes for controlled study: \emph{core dynamics} define the within-task decision problem; \emph{task-specific variation} captures instance-level differences across tasks; \emph{prompt type} controls how much information about the latent is revealed to the agent; and \emph{feedback type} determines what information is added to the running history after each task.

Together, these axes let us independently control three orthogonal dimensions of difficulty (Table~\ref{tab:difficulty_axes}): \emph{within-task difficulty} (how hard each task is in isolation), \emph{latent-identification difficulty} (how hard the $\mathrm{Latent}$ is to infer from interaction), and \emph{cross-task difficulty} (how strongly the {latent} governs subsequent tasks once inferred, jointly determined by the prior two). Varying the sequence horizon $N$ controls how many tasks are available for adaptation. Prompt and feedback modulate these axes orthogonally: a richer prompt lowers latent-identification difficulty by revealing partial information about the {latent}, while richer feedback (e.g., revealing ground-truth outcomes regardless of success) accelerates how quickly evidence about the latent accumulates across tasks. Together with horizon $N$, prompt and feedback set how much support the agent receives for inferring and exploiting the latent, while the three difficulty axes set how hard that inference and exploitation actually are.


\begin{table}[h]
\centering
\small
\begin{tabular}{@{}lll@{}}
\toprule
\textbf{Difficulty axis} & \textbf{Set by} & \textbf{Example settings (number guessing)} \\
\midrule
Within-task & Visible range & $[1,100]$,  $[1,1000]$,  $[1,10000]$ \\
Latent-identification & Size of hidden set & $2$, $5$, $10$ \\
Cross-task & Latent $+$ within-task difficulty & size $2$ in $[1,1000]$ vs.\ size $10$ in $[1,100]$ \\
\bottomrule
\end{tabular}
 \caption{\textbf{Difficulty axes in the number-guessing environment.} Within-task difficulty is set by the visible range; latent-identification difficulty by the size of the hidden set. Cross-task difficulty, how much knowing the latent helps on later tasks, depends on both: a hidden set small relative to the visible range makes the latent highly predictive, while one comparable to the range makes it nearly uninformative.}
 \label{tab:difficulty_axes}
\end{table}

\subsection{Composable and extensible software}
\label{sec:software}
 The four axes are modular, swappable components, and not hard-coded options inside a monolithic environment. A fully specified environment is the product of five independently registered components,
\begin{equation*}
  \texttt{FullyDefinedEnv} \;=\; \text{core-env} \,\times\, \text{latent} \,\times\, \text{prompt} \,\times\, \text{feedback} \,\times\, N,
\end{equation*}
The registry resolves any choice of components into a single runnable environment, named by an identifier that records the choice,  (number-guessing/set-of-3/no-info/standard/ep10). The core environment supplies the within-task game, the latent supplies the ground truth shared across tasks, the prompt sets how much of the latent the agent is told, the feedback sets what it observes after each task, and $N$ sets the horizon. Because each component registers on its own, the experimenter can change one axis without touching the others: swapping the prompt yields a hint condition while the latent stays fixed, and swapping the latent raises difficulty while everything else holds.

A new environment is its single-task core dynamics and its registrations.  The single-task dynamics conform to the standard \texttt{latentgym.core-env} interface, written from scratch or wrapped from an existing game. The latents, prompts, and feedbacks for that environment are declared in their own files and register themselves on import.  From these registrations the full set of environments is composed automatically, and the shared machinery for composition, sequence generation, evaluation, and training is reused without change. Figure~\ref{fig:eightlines} (left) shows the whole of number guessing in eight lines. Further, the composition is flexible enough that a single sequence can draw on a different core environment for each task while sharing  some latent.

\begin{figure}[t]
\centering
\tcbset{
  lgcodepanel/.style={
    colback=codebg, colframe=codeframe,
    coltitle=AgentTitle, colbacktitle=AgentBg,
    fonttitle=\footnotesize\bfseries\sffamily,
    boxrule=0.5pt, titlerule=0pt, arc=2.5pt,
    left=0.8em, right=0.8em, top=0.5em, bottom=0.5em,
    toptitle=0.35em, bottomtitle=0.35em,
    enhanced,
    attach boxed title to top left={xshift=0.6em, yshift=-2.6mm},
    boxed title style={colframe=codeframe, arc=2pt},
  }
}
\begin{minipage}[t]{0.49\linewidth}
\begin{tcolorbox}[lgcodepanel, title=Define a new environment, equal height group=lgfig]
\begin{lstlisting}[style=lgpy]
from latentgym.core.registry import register_env
from .core_env import NumberGuessingSingleEpisodeEnv

register_env(
    "number_guessing",
    NumberGuessingSingleEpisodeEnv,
    min_range=1,
    max_range=1000,
    max_turns_per_episode=30,
)
\end{lstlisting}
\end{tcolorbox}
\par\smallskip
{\footnotesize\textbf{(a)}~The within-task game is a \texttt{SingleEpisodeEnv}; its latents, prompts, and feedbacks live in their own files and register on import. Number guessing in eight lines.\par}
\end{minipage}
\hfill
\begin{minipage}[t]{0.49\linewidth}
\begin{tcolorbox}[lgcodepanel, title=Compose a configuration, equal height group=lgfig]
\begin{lstlisting}[style=lgpy]
FullyDefinedEnv(
    "number_guessing",  # core dynamics
    "set_of_3",         # latent
    "no_info",          # prompt
    "standard",         # feedback
    num_episodes=10,    # horizon N
)
\end{lstlisting}
\end{tcolorbox}
\par\smallskip
{\footnotesize\textbf{(b)}~The registry resolves the five axes into one runnable environment. Each axis varies on its own, here the \texttt{latent\_id} alone sets latent-identification difficulty (\texttt{set\_of\_2}, \texttt{set\_of\_3}, \texttt{range\_100}, \ldots), so a difficulty sweep is a handful of configs, not new code.\par}
\end{minipage}
\caption{\textbf{Define once, compose many.} A new environment supplies its core dynamics and registers its components (left). From these registrations, any choice of core dynamics, latent, prompt, feedback, and horizon $N$ composes automatically into a runnable environment (right); changing one axis leaves the others untouched.}
\label{fig:eightlines}
\end{figure}


Difficulty follows directly from the choice of components.  Within-task difficulty follows from the visible range (an environment parameter), and latent-identification difficulty from the size of the hidden set (a latent parameter). The two move independently, so each row of Table~\ref{tab:difficulty_axes} is a different combination of settings rather than new code (Figure~\ref{fig:eightlines}, right), and a difficulty sweep is a handful of such configurations.

One environment object serves both evaluation and training. It implements the standard text-environment interface, reports per-task reward and turn counts, and aggregates rewards as cumulative, terminal, improvement, or per-task. Switching between evaluating a frontier model and training on full task sequences is therefore a change of operating mode, not of environment. For RL training on full task sequences, the framework provides  built-in advantage estimators, and users can specify their own; these plug into SkyRL~\citep{cao2025skyrlagent}, which carries out the rollouts, weight synchronization, and policy optimization. A thin adapter exposes each \textsc{LatentGym} environment to SkyRL, so the framework contributes the estimators and the adapter while reusing a training stack built for that purpose.  For evaluation, an inspection dashboard renders the per-task reward curves and full trajectory conversations behind every run.

\subsection{Diagnosing how an agent adapts}
\label{sec:metrics}
The reward curve $r_1,\dots,r_N$ and its summaries ($R$, $r_N$, $r_N-r_1$) tell us whether an agent improves over a sequence. They do not tell us \emph{how}. Two agents with identical cumulative reward $R$ can differ sharply in whether they probe the latent or exploit what they already know, and the two failures call for different fixes. Telling them apart is hard, because exploration and exploitation are entangled within a single run, and any late-task reward reflects both how informative the early interactions were and how well the agent acted on them.

We separate the two with a counterfactual that the infrastructure supports directly. One agent gathers the experience, and a different agent acts on it. At a switch point $K\in\{1,\dots,N-1\}$, agent $A$ plays tasks $1,\dots,K$ and produces the accumulated history; control then passes to agent $B$, which inherits that same history and plays tasks $K{+}1,\dots,N$. The framework preserves the full running context across the hand-off and attributes each task's reward to whichever agent played it, so a single run yields the post-switch (tail) reward
\[
R_{\mathrm{tail}}(A\!\to\!B)=\sum_{i=K+1}^{N} r_i \qquad\text{($A$ explores, $B$ exploits).}
\]
Comparing two agents then isolates each capability by holding one role fixed at a reference agent $C\in\{A,B\}$. Holding the \emph{exploiter} fixed and swapping the explorer measures \textbf{exploration}; holding the \emph{explorer} fixed and swapping the exploiter measures \textbf{exploitation}:
\[
\mathrm{ExploreGain}_C(B\,\text{vs.}\,A)=R_{\mathrm{tail}}(B\!\to\!C)-R_{\mathrm{tail}}(A\!\to\!C),\]
\[\mathrm{ExploitGain}_C(B\,\text{vs.}\,A)=R_{\mathrm{tail}}(C\!\to\!B)-R_{\mathrm{tail}}(C\!\to\!A).
\]
We report both choices of the reference $C$ to show the gap is robust to the held-fixed role. Sweeping $K$ traces where each capability accrues, since a larger $K$ gives the explorer more tasks to gather evidence while leaving the exploiter fewer tasks to act on.

\subsection{The seven environments}
We instantiate \textsc{LatentGym} with seven environments spanning different task dynamics and latent structures (Table~\ref{tab:environments}), built on top of the TextArena suite of text-based games~\citep{guertler2025textarena}. In TextArena each task is a single-game episode the agent plays in natural language. \textsc{LatentGym} differs in that each environment is a sequence of $N$ games drawn from a shared latent structure: the agent plays multiple games of the same family (latent) in sequence and is rewarded for getting better across them. Each environment supports three prompt conditions (Table~\ref{tab:prompt-types}): no information about the latent, a vague hint that cross-task structure may exist, and full information about the latent. It also supports two feedback conditions: a \emph{standard} condition, in which the agent receives only a binary success/failure signal after each task, and an \emph{information} condition, in which the ground-truth outcome is revealed regardless of whether the agent succeeded. Full environment specifications, including latents of varying difficulty levels, are in Appendix~\ref{app:envs}.

\begin{table}[h!]
\centering
\small
\begin{tabular}{p{0.22\linewidth} p{0.34\linewidth} p{0.36\linewidth}}
\toprule
\textbf{Environment} & \textbf{Core task} & \textbf{Representative latent structures} \\
\midrule
Number Guessing & Identify hidden number & Restricted target set, range pattern \\
Bandits & Select rewarding arm & Shared best arm, reward pattern  \\
Secretary & Accept/reject candidates & Threshold, position pattern \\
Mastermind & Infer hidden code & Code constraint, structural rule \\
Word Ladder & Transform word through graph & Hub word, reusable path structure \\
Wordle & Guess hidden word from feedback & Shared word property, candidate-set \\
Hangman & Reveal word through letter guesses & Shared word category, letter pattern \\
\bottomrule
\end{tabular}
\caption{Environment suite instantiated from the framework. Each environment defines a within-task decision problem and a family of shared latent structures.}
\label{tab:environments}
\end{table}

\begin{table}[h!]
\centering
\small
\begin{tabular}{p{0.16\linewidth} p{0.34\linewidth} p{0.40\linewidth}}
\toprule
\textbf{Prompt type} & \textbf{Information revealed} & \textbf{Purpose} \\
\midrule
No information 
& No indication that tasks share a latent structure 
& Tests whether the agent discovers structure from experience alone \\

Vague hint 
& States that some recurring structure may exist 
& Tests whether the agent identifies the cross-task regularities efficiently \\

Full information 
& Explicitly describes the latent structure 
& Tests whether the agent can use latent information once given \\
\bottomrule
\end{tabular}
\caption{Prompts vary how much information about the shared latent is available before interaction.}
\label{tab:prompt-types}
\end{table}

\section{Failure Modes of Frontier Models in a Minimal Setting}
\label{sec:sota_failure_modes}

A controlled testbed is useful only if existing models struggle on it, exposing where they fall short. Our first demonstration
establishes that \textsc{LatentGym} does this: the environments are deliberately simple, yet frontier models fail to learn across tasks on them, and they fail in interpretable ways, making it a discriminating place to develop and compare methods. We evaluate three representative frontier models (GPT-4o, Claude Sonnet 4.6, and Gemini 2.5 Flash) chosen to span different providers and capability profiles rather than to form an exhaustive leaderboard.  These models fail at cross-task adaptation even under simple latent structures and across varying prompt conditions.
Inspecting individual trajectories surfaces three recurring failure modes.

\paragraph{Adaptation neglect: models repeatedly restart from scratch.}
When models are not explicitly told that a hidden latent pattern exists,
they often solve each task largely independently, without attempting to
infer reusable structure from previous experience. This appears even in
very simple latent structures, and we occasionally observe similar
behavior even when models are told that such a pattern exists. For example, in number guessing (latent: \texttt{set\_of\_3}), the target is always drawn from the same three values, yet GPT-4o, Gemini and Claude repeatedly begin later tasks with the generic binary search strategy rather than testing previously revealed targets.

 \begin{anecdotebox}{Claude Sonnet 4.6 solves each task independently}{ClaudeColor}
\noindent \textbf{Setup:} In each task the agent needs to identify a hidden integer between 1 and 1000; after each guess, the game master replies whether the target is greater than, less than, or equal to the guess. The \textbf{latent} across tasks is = \{592, 781, 926\}, that is the hidden number is actually only 1 out of these 3 numbers. Below we show Task 1, Task 4 and Task 9 out of stream of 10 tasks as solved by Claude. In all the tasks agent starts with binary search without recognizing that actually the hidden numbers are repeated. (Full transcript is given in Appendix~\ref{app:adaptation_neglect})

\noindent\begin{tabularx}{\linewidth}{@{}X@{\hspace{4pt}}X@{\hspace{4pt}}X@{}}
\begin{taskbox}{Task 1}{ClaudeColor}
\agentmsg{1}{I'll use a binary search strategy. [500]}
\par\smallskip\noindent\textcolor{gray}{\scriptsize\itshape \dots\ continues for 6 more turns. }\par
\end{taskbox} & \begin{taskbox}{Task 4}{ClaudeColor}
\agentmsg{1}{Let me start with binary search again. [500]}
\par\smallskip\noindent\textcolor{gray}{\scriptsize\itshape \dots\ continues for 9 more turns.}\par
\end{taskbox} & \begin{taskbox}{Task 9}{ClaudeColor}
\agentmsg{1}{Let me start with binary search again. [500]}
\par\smallskip\noindent\textcolor{gray}{\scriptsize\itshape \dots\ continues for 9 more turns.}\par
\end{taskbox}
\\
\end{tabularx}
\end{anecdotebox}

\begin{figure}[h!]
        \includegraphics[width=\linewidth]{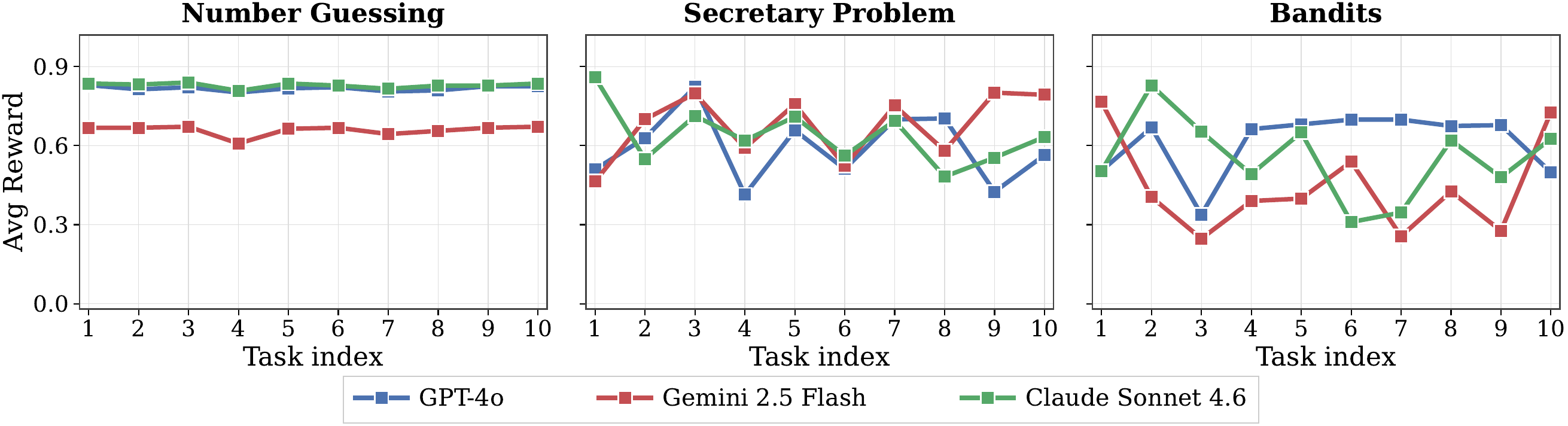}
    \caption{Average reward across tasks under the no-information prompt condition for three environments: number guessing (set of 3), secretary (threshold 0.6), and bandits (best arm 0).}
    \label{fig:no-info-reward-three-envs}
\end{figure}

Similar  results appear  across models and  environments, see Figure \ref{fig:no-info-reward-three-envs} and  Appendix~\ref{app:adaptation_neglect} for representative transcripts. 
At first glance, this behavior may appear reasonable: a well-instructed agent should not impose hidden assumptions on a task. However, the pattern is more concerning in an adaptation setting. Real deployments contain recurring regularities that are not specified; an agent that only acts on explicitly stated structure will fail to improve with experience.

\paragraph{Adaptation breakdown: models notice structure but fail to use it reliably.}
When models are told that some pattern may exist, they often show partial but ineffective adaptation. Sometimes they ignore the hint and continue acting myopically (Appendix~\ref{app:adaptation_breakdown_models_ignore}); sometimes they search for a pattern but fail to identify the correct one (Appendix~\ref{app:adaptation_breakdown_models_fail_identify_pattern}); sometimes they correctly infer part of the structure but do not act on it (Appendix~\ref{app:adaptation_breakdown_models_fail_to_act}); and sometimes they try too hard to fit a pattern and drift away from the original objective of solving the task (Appendix~\ref{app:adaptation_breakdown_model_tries_hard}).

\begin{anecdotebox}{Gemini 2.5 Flash recognizes the pattern but does not act upon it}{ClaudeColor}
\noindent Agent task is to identify a hidden integer between 365 and 10365; after each guess, the game master replies whether the target is greater than, less than, or equal to the guess. The \textbf{hidden latent} is that the numbers actuall fall in a range covers [1400, 2400]. Below we 
show Task 7 and Task 10 out of 10 tasks.
\setlength{\columnsep}{6pt}
\begin{multicols}{2}\raggedcolumns
\begin{taskbox}{Task 7}{ClaudeColor}
\agentmsg{1}{ I've begun the process by defining the initial boundaries: 365 to 10365. My strategy remains binary search \ldots [5365] }
\par\smallskip\noindent\textcolor{gray}{\scriptsize\itshape \dots\ continues for 12 more turns.}\par
\end{taskbox}
\columnbreak
\begin{taskbox}{Task 10}{ClaudeColor}
\agentmsg{1}{
 I've reviewed the previous nine games. The numbers, ranging from 1461 to 2278, are well below the upper bound of 10365, although I'll continue using binary search \ldots [5365]}
\par\smallskip\noindent\textcolor{gray}{\scriptsize\itshape \dots\ continues for 12 more turns.}\par
\end{taskbox}
\end{multicols}
\end{anecdotebox}

These failures point to a missing exploration--exploitation policy. The models are not simply incapable of forming hypotheses: in many trajectories, they explicitly propose plausible cross-task patterns. However, they do not know when a vague signal warrants further search, when enough evidence has been collected, or how strongly a hypothesis should influence future actions. 
More broadly, these cases suggest that recognizing latent structure and exploiting it are separable capabilities: a model can partially succeed at one while failing at the other.

\begin{figure}[h!]
        \centering
        \includegraphics[width=\linewidth]{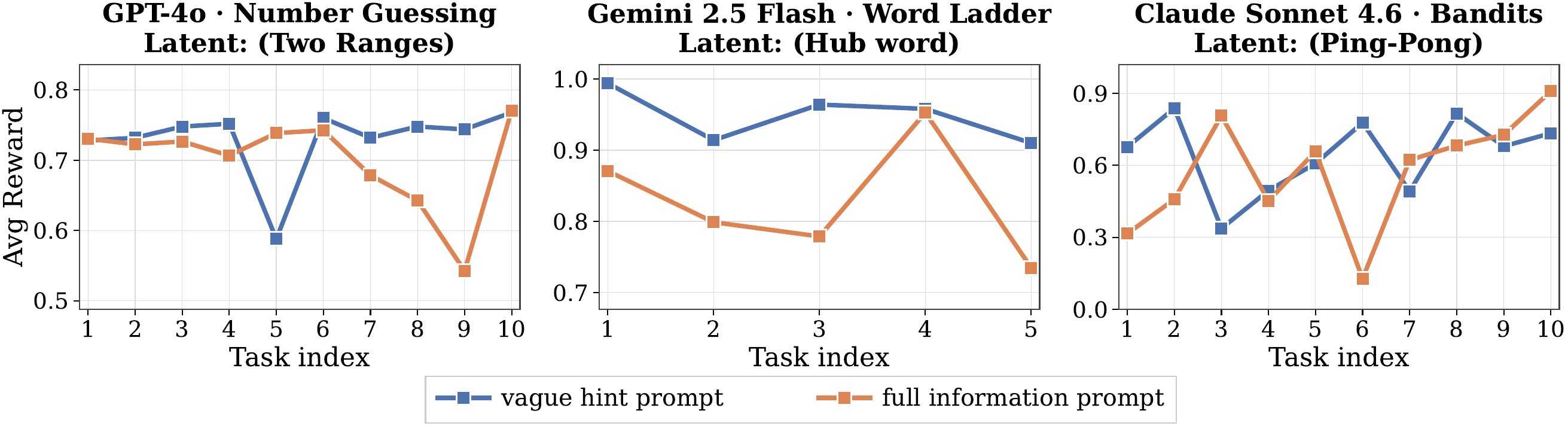}
    \caption{Examples where full information harms performance.}
    \label{fig:full-info-harms-examples}
\end{figure}

\paragraph{Adaptation miscalibration: more information can make performance worse.}
Even when the relevant pattern is explicitly provided, models can suffer from miscalibrated adaptation and their performance might degrade  relative to weaker description of the task (Figure~\ref{fig:full-info-harms-examples}).  
 The explicit description appears to introduce a competing sub-goal: instead of using the latent as a calibrated prior, the model optimizes for demonstrating the stated pattern, over-applying the rule or choosing actions that satisfy its interpretation while violating the operational constraints of the game. Examples include proposing off-graph words in word ladder, choosing the wrong sub-range in number guessing, committing to a single arm in alternating bandits, or accepting too early in secretary-style tasks. We observe this across \texttt{number\_guessing/two\_ranges}, \texttt{secretary/increasing\_position}, \texttt{wordladder/hub\_word}, and \texttt{bandits/ping\_pong}; representative transcripts are in Appendix~\ref{app:failure-anecdotes-miscalibrated-adaptation}. These cases show that in-context adaptation requires more than access to the right information: the model must also decide how strongly to condition on it.

These failure modes show what frontier models lack: a robust cross-task strategy for deciding when to search for structure, when to commit, and how strongly to condition on explicit information without overfitting to it.
  This is precisely the gap our benchmark is designed to expose.

\section{Demonstration: Cross-task RL for In-context Adaptation}
\label{sec:experiments}
 
Our second demonstration uses \textsc{LatentGym} to ask whether RL fine-tuning can instill general cross-task
experiential learning. We compare two recipes. \emph{Single-task RL} is the standard approach: fine-tune the policy on isolated tasks, with the reward depending on the reward $r$ of that single task. \emph{Cross-task RL} fine-tunes end-to-end on full sequence of $N$ related tasks under a shared latent $z$, with the reward depending on the entire sequence of rewards $r_1, \ldots, r_N$. Under this signal, the policy is rewarded for using early tasks to infer  the latent and act more effectively on later ones. 

We address three research questions: 
\begin{enumerate}
    \item Is cross-task RL necessary, or does single-task RL on the same task family suffice? 
    \item Do the strategies learned by cross-task RL generalize beyond the training distribution? 
    \item Which capability drives cross-task RL's lift, is it better at exploration or exploitation?
\end{enumerate}

\paragraph{Experimental Setup.}
For the experiments, we fine-tune Qwen3-8B on sequences sampled from five environments described in Section~\ref{sec:framework}: Number Guessing, Mastermind, Hangman, Word Ladder, and Secretary. For each environment, we train on a fixed set of latent structures. Each training sequence consists of $N{=}10$ tasks, with the policy carrying its history as context across tasks. We compare three variants: the \emph{base} instruction-tuned model; \emph{single-task RL}, which fine-tunes with $N{=}1$ and improves raw task ability; and \emph{cross-task RL}, which fine-tunes on full sequences. We report the three cross-task metrics defined in Section~\ref{sec:framework}: cumulative reward $R = \sum_i r_i$, cross-task gain $r_N - r_1$, and final-task reward $r_N$. We fine-tune the models with Group Relative Policy Optimization (GRPO)~\citep{shao2024deepseekmath}, using the standard clipped objective with a KL penalty against a frozen reference $\pi_{\mathrm{ref}}$.  Each rollout over a sequence receives a single reward summarizing its per task rewards; we use the cumulative reward $R =\sum_i r_i$ that for finetuning. 
Complete training and evaluation details including hyperparameters, and per-environment latent structures are given in Appendix~\ref{app:experimental-details}.

\paragraph{Cross-Task RL induces adaptation.} We evaluate the base, single-task RL, and cross-task RL variants on five environments. Each RL variant is trained separately on each of six environments on sequences of $N{=}10$ tasks under standard feedback. On cumulative reward $R = \sum_i r_i$ (Table~\ref{tab:study1-numbers}), cross-task RL leads on every environment, beating the base by an average of $99\%$ and single-task RL by $39\%$, with the largest gains over single-task RL on  Wordle ($+91\%$) and Number Guessing ($+72\%$). Single-task RL closes part of the gap on Word Ladder ($+5\%$) and
Secretary ($+8\%$), where the per-task problem is the main bottleneck,
but trails cross-task RL elsewhere. The Gain column $(r_N - r_1)/r_1$ gives the clearest picture of \emph{cross-task} adaptation: cross-task RL is the only variant with consistently positive Gain across environments (largest on Number Guessing, $+75\%$, and Secretary, $+24\%$), while single-task RL is negative on five of six environments (e.g., $-61\%$ on Wordle, $-18\%$ on Mastermind), indistinguishable from the base. Final-task reward $r_N$ tracks the
same ranking; cross-task RL improves $r_N$ by an average of $55\%$ over
single-task RL, with the largest single-environment gains on Wordle
($2.5\times$) and Number Guessing ($2.0\times$). This shows that cross-task adaptation
requires a cross-task RL training.

To further test whether this finding reflects a learned adaptation strategy rather than environment-specific cues, we train a single model on sequences drawn from \emph{multiple} environments (Number Guessing, Secretary, Hangman, and Word Ladder, with two latents each). Notably, this multi-env cross-task model exceeds the single-env cross-task models trained directly on each environment, by an average of $19\%$ in cumulative reward with the largest gains on Secretary ($33\%$)(Table~\ref{tab:study2-numbers}). This suggests cross-task RL may be learning a general adaptation strategy rather than environment-specific cues, and the policy may benefit from training on a mixture of environments.

\begin{table}[h!]
\centering
\small
\setlength{\tabcolsep}{4pt}
\renewcommand{\arraystretch}{1.1}
\begin{tabular}{l ccc ccc ccc}
\toprule
& \multicolumn{3}{c}{Cross-task RL} & \multicolumn{3}{c}{Single-task RL} & \multicolumn{3}{c}{Base (Qwen3-8B)} \\
\cmidrule(lr){2-4} \cmidrule(lr){5-7} \cmidrule(lr){8-10}
Environment
  & \makecell{Cum.\\Reward} & \makecell{Gain\\ (\%)} & \makecell{Final\\Reward}
  & \makecell{Cum.\\Reward} & \makecell{Gain\\ (\%)} & \makecell{Final\\Reward}
  & \makecell{Cum.\\Reward} & \makecell{Gain\\ (\%)} & \makecell{Final\\Reward} \\
\midrule
Number guessing & $\mathbf{5.78}$ & $\mathbf{+75\%}$  & $\mathbf{0.63}$ & $3.36$ & $-11\%$ & $0.31$ & $2.23$ & $-18\%$ & $0.18$ \\
Mastermind      & $\mathbf{7.13}$ & $\mathbf{+3\%}$   & $\mathbf{0.70}$ & $5.12$ & $-18\%$ & $0.51$ & $4.32$ & $-4\%$  & $0.44$ \\
Hangman         & $\mathbf{6.48}$ & $\mathbf{+5\%}$   & $\mathbf{0.65}$ & $5.53$ & $-5\%$  & $0.55$ & $3.47$ & $-3\%$  & $0.33$ \\
Wordladder      & $\mathbf{9.20}$ & $\mathbf{+1\%}$   & $\mathbf{0.93}$ & $8.80$ & $-2\%$  & $0.88$ & $5.35$ & $-5\%$  & $0.54$ \\
Secretary       & $\mathbf{7.60}$ & $\mathbf{+24\%}$  & $\mathbf{0.78}$ & $7.07$ & $+3\%$  & $0.72$ & $4.58$ & $+15\%$ & $0.46$ \\
Wordle          & $\mathbf{6.23}$ & $\mathbf{+11\%}$  & $\mathbf{0.61}$ & $3.26$ & $-61\%$ & $0.24$ & $2.54$ & $+8\%$  & $0.26$ \\
\bottomrule
\end{tabular}
\caption{\textbf{Cross-task RL is necessary for in-context adaptation.}
Comparison of Cross-task RL, single-task RL, and the base (Qwen3-8B) model across the six environments. \emph{Cum. Reward} is the total reward over the sequence ($R = \sum_i r_i$); \emph{Gain} is the cross-task improvement from first to last as a percentage of the initial reward ($(r_N - r_1)/r_1$); \emph{Final Reward} is the reward on the last task ($r_N$). Each entry averages over the three in-distribution training latents and over $50$ trajectories per latent.}
\label{tab:study1-numbers}
\end{table}

\begin{table}[h!]
\centering
\footnotesize
\setlength{\tabcolsep}{3pt}
\renewcommand{\arraystretch}{1.1}
\resizebox{\columnwidth}{!}{%
\begin{tabular}{l ccc ccc ccc ccc}
\toprule
& \multicolumn{3}{c}{Cross-task (multi env)} & \multicolumn{3}{c}{Single (multi env)} & \multicolumn{3}{c}{Cross-task (one env)} & \multicolumn{3}{c}{Base (Qwen3-8B)} \\
\cmidrule(lr){2-4} \cmidrule(lr){5-7} \cmidrule(lr){8-10} \cmidrule(lr){11-13}
Environment
  & \makecell{Cum.\\Reward} & \makecell{Gain\\ (\%)} & \makecell{Final\\Reward}
  & \makecell{Cum.\\Reward} & \makecell{Gain\\ (\%)} & \makecell{Final\\Reward}
  & \makecell{Cum.\\Reward} & \makecell{Gain\\ (\%)} & \makecell{Final\\Reward}
  & \makecell{Cum.\\Reward} & \makecell{Gain\\ (\%)} & \makecell{Final\\Reward} \\
\midrule
Number guessing & $\mathbf{7.92}$ & $+25\%$          & $\mathbf{0.86}$ & $6.73$ & $+14\%$ & $0.67$          & $6.28$ & $\mathbf{+44\%}$ & $0.62$          & $2.97$ & $-45\%$          & $0.18$ \\
Hangman         & $\mathbf{8.55}$ & $+14\%$          & $\mathbf{0.89}$ & $5.63$ & $+2\%$  & $0.55$          & $7.25$ & $+14\%$          & $0.75$          & $4.68$ & $\mathbf{+33\%}$ & $0.48$ \\
Wordladder      & $\mathbf{8.89}$ & $\mathbf{+2\%}$  & $\mathbf{0.90}$ & $8.61$ & $-4\%$  & $0.87$          & $8.80$ & $+1\%$           & $0.90$          & $4.58$ & $-11\%$          & $0.47$ \\
Secretary       & $\mathbf{9.54}$ & $\mathbf{+33\%}$ & $\mathbf{0.93}$ & $6.35$ & $-43\%$ & $0.36$          & $7.19$ & $+36\%$          & $0.75$          & $4.76$ & $+24\%$          & $0.47$ \\
\bottomrule
\end{tabular}%
}
\caption{\textbf{Multi-environment cross-task RL.}
The multi-env cross-task model outperforms single-environment cross-task RL, with an average improvement of 19\% on cumulative reward.}
\label{tab:study2-numbers}
\end{table}

\paragraph{Do strategies learned generalize beyond the training distribution?}
Next we test whether cross-task RL's gains generalize beyond the training distribution. A policy that has learned a transferable adaptation strategy should generalize beyond the specific latents and environments seen during training. We test this at two levels: held-out latents within trained environments (OOD-1), and entirely held-out environments (OOD-2).

For \textbf{OOD-1}, we hold out a latent within each environment and evaluate cross-task RL on sequences generated under those held-out latents: the environment is fixed, only the latent shifts. Cross-task RL retains a clear advantage on cumulative reward in five of six environments (Table~\ref{tab:study3-numbers}), beating the base by an average of $55\%$ and single-task RL by $26\%$, with the largest gains over single-task RL on Wordle ($+65\%$) and Number Guessing ($+52\%$). The Gain column $(r_N - r_1)/r_1$ remains positive for cross-task RL in five of six environments, and cross-task RL achieves the highest Final Reward $r_N$ in five of six environments.

For \textbf{OOD-2}, we use a leave-one-out protocol over the same four environments as the multi-env experiment above: we train on sequences from three of them and test on the held-out fourth, so the policy has not seen the target environment in any latent during training. The leave-one-out model beats the base on cumulative reward on all four held-out environments by an average of $14\%$, with the largest gains on Hangman ($21\%$) and Number Guessing ($20\%$) (Table~\ref{tab:study4-numbers}); $r_N$ is also higher on every environment. The Gain results are mixed: the leave-one-out model beats the base on three of four environments but trails on Hangman, where the base shows a large positive Gain ($+33\%$) driven partly by a low initial reward ($r_1 \approx 0.36$, vs. $r_1 \approx 0.55$ for the leave-one-out model).

Since training optimizes cumulative reward $\sum_i r_i$, generalization on that metric is the expected effect of training, and this is indeed what we observe at both OOD levels. More encouragingly, we also see some evidence of generalization on Gain and on $r_N$, neither of which is directly optimized. Gain should be interpreted with caution, however: a model already strong on the initial task has limited room to improve (ceiling effect), while a model that is weak initially can post a large Gain simply from a low starting point (floor effect). Final Reward $r_N$ should therefore be considered alongside Gain when interpreting generalization, and we do see cross-task RL's $r_N$ exceed the base in almost all OOD cases (five of six environments under OOD-1, all four under OOD-2). Since generalization tends to improve with data scale in LLM fine-tuning more broadly, scaling these experiments to a wider range of environments (the analogue of data scaling in our setting) along with more elaborate studies such as varying the reward functional $\Phi$ or controlling for changes in per-task ability, is left to future work.

\begin{table}[h!]
\centering
\small
\setlength{\tabcolsep}{4pt}
\renewcommand{\arraystretch}{1.1}
\begin{tabular}{l ccc ccc ccc}
\toprule
& \multicolumn{3}{c}{Cross-task RL} & \multicolumn{3}{c}{Single-task RL} & \multicolumn{3}{c}{Base (Qwen3-8B)} \\
\cmidrule(lr){2-4} \cmidrule(lr){5-7} \cmidrule(lr){8-10}
Environment
  & \makecell{Cum.\\Reward} & \makecell{Gain\\ (\%)} & \makecell{Final\\Reward}
  & \makecell{Cum.\\Reward} & \makecell{Gain\\ (\%)} & \makecell{Final\\Reward}
  & \makecell{Cum.\\Reward} & \makecell{Gain\\ (\%)} & \makecell{Final\\Reward} \\
\midrule
Number guessing & $\mathbf{6.11}$ & $\mathbf{+23\%}$ & $\mathbf{0.59}$ & $4.03$          & $-9\%$           & $0.39$          & $2.82$ & $-49\%$          & $0.18$ \\
Mastermind      & $\mathbf{4.29}$ & $\mathbf{+10\%}$ & $\mathbf{0.43}$ & $3.48$          & $-10\%$          & $0.35$          & $3.46$ & $-13\%$          & $0.34$ \\
Hangman         & $\mathbf{6.84}$ & $+30\%$          & $\mathbf{0.70}$ & $5.90$          & $+17\%$          & $0.61$          & $5.54$ & $\mathbf{+88\%}$ & $0.60$ \\
Wordladder      & $\mathbf{9.35}$ & $\mathbf{-1\%}$  & $\mathbf{0.93}$ & $9.25$          & $-4\%$           & $0.91$          & $7.42$ & $-6\%$           & $0.74$ \\
Secretary       & $4.31$          & $+6\%$           & $0.38$          & $\mathbf{4.43}$ & $+5\%$           & $\mathbf{0.43}$ & $4.33$ & $\mathbf{+30\%}$ & $0.43$ \\
Wordle          & $\mathbf{5.07}$ & $\mathbf{+21\%}$ & $\mathbf{0.57}$ & $3.07$          & $-53\%$          & $0.26$          & $2.13$ & $-41\%$          & $0.19$ \\
\bottomrule
\end{tabular}
\caption{\textbf{Generalization under latent shift (OOD-1).}
Cross-task RL beats the base by an average of 55\% and single-task RL by 26\% on cumulative reward. 
}
\label{tab:study3-numbers}
\end{table}

\begin{table}[h!]
\centering
\small
\setlength{\tabcolsep}{4pt}
\renewcommand{\arraystretch}{1.1}
\begin{tabular}{l ccc ccc}
\toprule
& \multicolumn{3}{c}{Cross-task RL (LOO)} & \multicolumn{3}{c}{Base (Qwen3-8B)} \\
\cmidrule(lr){2-4} \cmidrule(lr){5-7}
Environment
  & \makecell{Cum.\\Reward} & \makecell{Gain\\ (\%)} & \makecell{Final\\Reward}
  & \makecell{Cum.\\Reward} & \makecell{Gain\\ (\%)} & \makecell{Final\\Reward} \\
\midrule
Number guessing & $\mathbf{3.57}$ & $\mathbf{-9\%}$  & $\mathbf{0.32}$ & $2.97$ & $-45\%$          & $0.18$ \\
Hangman         & $\mathbf{5.65}$ & $+5\%$           & $\mathbf{0.58}$ & $4.68$ & $\mathbf{+33\%}$ & $0.48$ \\
Wordladder      & $\mathbf{5.15}$ & $\mathbf{+4\%}$  & $\mathbf{0.51}$ & $4.58$ & $-11\%$          & $0.47$ \\
Secretary       & $\mathbf{4.84}$ & $\mathbf{+32\%}$ & $\mathbf{0.49}$ & $4.76$ & $+24\%$          & $0.47$ \\
\bottomrule
\end{tabular}
\caption{\textbf{Generalization under environment shift (OOD-2).}
A leave-one-out cross-task model, trained on all environments \emph{except} the evaluation one, beats the base on every held-out environment, with an average improvement of 14\% on cumulative reward. 
}
\label{tab:study4-numbers}
\end{table}

\begin{figure}[h!]
    \centering
    \includegraphics[width=\linewidth]{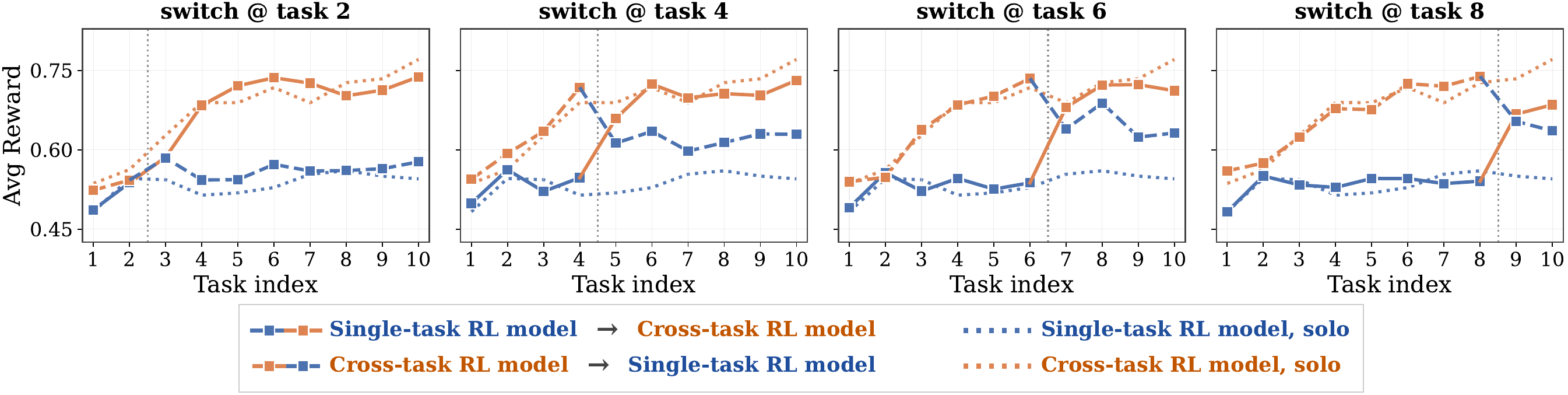}
    \caption{\textbf{Agent-switching: cross-task RL vs.\ single-task RL on Number Guessing.} At a switch point $k \in \{2, 4, 6, 8\}$, control is handed over to a different agent that inherits the cross-task context $h_{<k}$ accumulated by the first agent. Solid/Dashed lines: trajectories of the two switch directions. Dotted lines: each agent running the full sequence alone (\emph{solo}).}
    \label{fig:switch_streamRL_vs_singleRL}
\end{figure}

\paragraph{Exploration vs.\ exploitation efficiency.}
The aggregate gains in Tables~\ref{tab:study1-numbers}--\ref{tab:study2-numbers} do not reveal whether Cross-Task RL's lift comes from better exploration, better exploitation, or both. We isolate the two on Number Guessing with the agent-switching experiment from Section~\ref{sec:framework} (Figure~\ref{fig:switch_streamRL_vs_singleRL}): at a switch point $K \in \{2, 4, 6, 8\}$, control is handed from one agent (the \emph{explorer}, running on tasks $1,\ldots,K$) to a second agent (the \emph{exploiter}, running on tasks $K+1,\ldots,N$), which inherits the running history accumulated by the first. Writing $\mathrm{CT}$ and $\mathrm{ST}$ for the Cross-Task and Single-task RL agents and using the post-switch reward $R_{\mathrm{tail}}(\cdot\!\to\!\cdot)$ of Section~\ref{sec:framework}, we report the exploration and exploitation gains of $\mathrm{CT}$ over $\mathrm{ST}$ with the reference role held at $C\in\{\mathrm{CT},\mathrm{ST}\}$:
\[
\mathrm{ExploreGain}_C = \frac{R_{\mathrm{tail}}(\mathrm{CT}\!\to\!C) - R_{\mathrm{tail}}(\mathrm{ST}\!\to\!C)}{R_{\mathrm{tail}}(\mathrm{ST}\!\to\!C)},\]
\[
\mathrm{ExploitGain}_C = \frac{R_{\mathrm{tail}}(C\!\to\!\mathrm{CT}) - R_{\mathrm{tail}}(C\!\to\!\mathrm{ST})}{R_{\mathrm{tail}}(C\!\to\!\mathrm{ST})},
\]
each a percentage of the reference's tail reward. $\mathrm{ExploreGain}_C$ holds the exploiter at $C$ and swaps the explorer; $\mathrm{ExploitGain}_C$ holds the explorer at $C$ and swaps the exploiter (Table~\ref{tab:ng-collection-utilization}). Cross-Task RL has a positive exploration advantage at every $K$, but it is small: between $0.7\%$ and $17.4\%$ depending on the exploiter, with a mild increasing trend in $K$ as more of the sequence is built before the switch. The exploitation advantage is far larger and present at every $K$: holding the explorer fixed, switching the exploiter from Single-task RL to Cross-Task RL adds between $13\%$ and $30\%$ in tail reward, with values decreasing in $K$ as the post-switch tail shrinks and there is less room for exploitation to compound. On Number Guessing, the exploitation gap therefore dominates the exploration gap by a factor of roughly $2$--$30\times$ and is the larger driver of Cross-Task RL's lift. The same overall pattern, Cross-Task RL gaining on both axes simultaneously, holds across the other environments, though which axis dominates varies by environment (Appendix~\ref{app:data_collection_utilization}).

\begin{table}[h!]
\centering
\small
\setlength{\tabcolsep}{8pt}
\renewcommand{\arraystretch}{1.15}
\begin{tabular}{c c cc}
\toprule
Switch point $K$ & Reference $C$ & $\mathrm{ExploreGain}_C$ & $\mathrm{ExploitGain}_C$ \\
\midrule
\multirow{2}{*}{2} & $C=\mathrm{CT}$   & $+0.7\%$  & $+25.3\%$ \\
                   & $C=\mathrm{ST}$  & $+4.4\%$  & $+30.1\%$ \\
\midrule
\multirow{2}{*}{4} & $C=\mathrm{CT}$   & $+2.5\%$  & $+16.4\%$ \\
                   & $C=\mathrm{ST}$  & $+14.2\%$ & $+29.5\%$ \\
\midrule
\multirow{2}{*}{6} & $C=\mathrm{CT}$   & $+2.9\%$  & $+13.0\%$ \\
                   & $C=\mathrm{ST}$  & $+16.9\%$ & $+28.5\%$ \\
\midrule
\multirow{2}{*}{8} & $C=\mathrm{CT}$   & $+11.6\%$ & $+16.3\%$ \\
                   & $C=\mathrm{ST}$  & $+17.4\%$ & $+22.7\%$ \\
\bottomrule
\end{tabular}
\caption{\textbf{Cross-Task RL vs.\ Single-task RL on \texttt{number\_guessing}.}
Each cell reports $\mathrm{ExploreGain}_C$ or $\mathrm{ExploitGain}_C$: the relative tail-reward advantage of the Cross-task RL agent ($\mathrm{CT}$) over the Single-task RL agent ($\mathrm{ST}$), evaluated at switch point $K$. $\mathrm{ExploreGain}_C$ holds the exploiter fixed at the reference agent $C$ and swaps the explorer; $\mathrm{ExploitGain}_C$ holds the explorer fixed at $C$ and swaps the exploiter. We report both choices $C \in \{\mathrm{CT},\,\mathrm{ST}\}$ to show the gap is robust to the held-fixed role. All values positive: Cross-task RL is both a better explorer and a better exploiter than Single-task RL.}
\label{tab:ng-collection-utilization}
\end{table}

\section{Discussion}
\label{sec:conclusion}
We have presented \textsc{LatentGym}, a small-scale, controllable laboratory for cross-task experiential learning, and used it in three studies. The frontier-model study showed that a deliberately simple setting still exposes interpretable failures (Section~\ref{sec:sota_failure_modes}). The Cross-Task RL study showed that training can induce the capability and that it generalizes to held-out latents and environments (Section~\ref{sec:experiments}). The study below shows that the inter-task feedback used during training has a large and counterintuitive effect on generalization. This shows that   a controllable latent, together with one environment object shared across evaluation, diagnosis, and training, makes questions that are otherwise hard to pose directly answerable. We see \textsc{LatentGym} as an instrument for developing methodology, a place to prototype and compare algorithms cheaply before scaling to costlier, real-world settings.

\subsection{A worked example: training-time feedback shapes generalization}
\label{sec:algorithmic}
 
\begin{figure}[h!]
    \centering
    \includegraphics[width=0.5\linewidth]{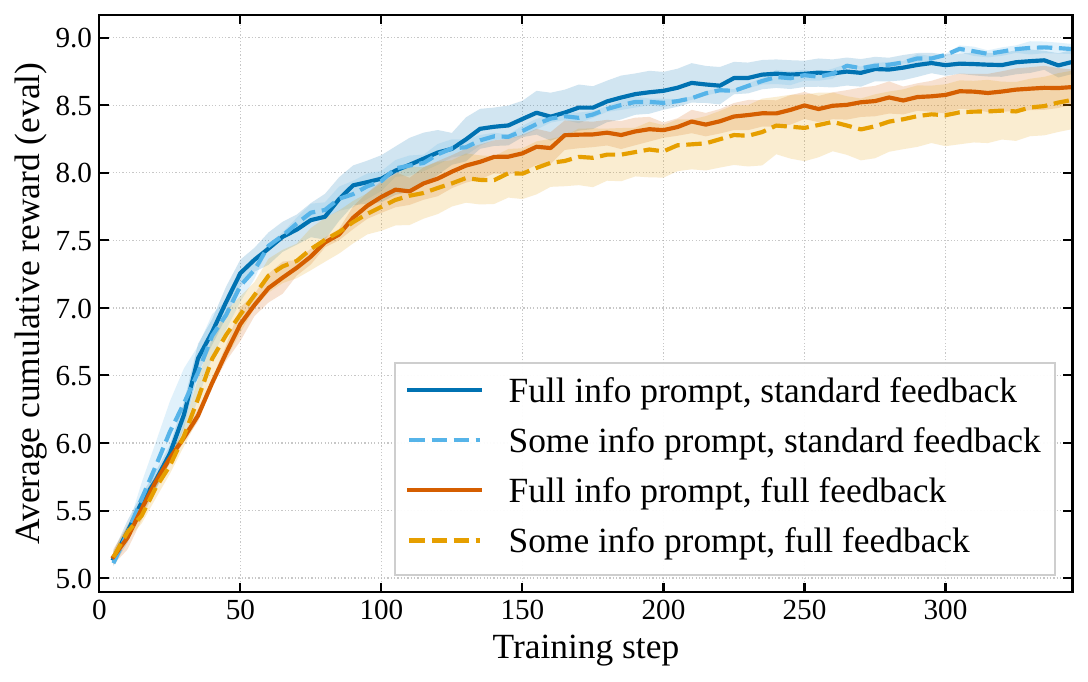}
    \caption{Average cumulative reward on the evaluation set during training, across the four prompt/feedback combinations used at training time.  
    Standard-feedback outperform full-feedback regardless of training prompt while prompt type produces a smaller gap.}
    \label{fig:feedback-eval}
\end{figure}

As an example of the kind of question the suite makes answerable, we ask how the \emph{in-context feedback} used during training affects training dynamics and generalization. We train models under combinations of standard vs.\ full feedback between tasks and some-info vs.\ full-info prompt at training time, while keeping the cumulative-reward objective $R=\sum_i r_i$ fixed, and evaluate each across all four prompt/feedback combinations at deployment. Counterintuitively, standard feedback at training time outperforms full feedback under both prompt regimes (Figure~\ref{fig:feedback-eval}), despite providing strictly less information between tasks. The advantage comes from robustness to the eval-feedback shift rather than from faster learning. Full-feedback-trained models drop by $0.509$ points when the eval feedback switches to standard, whereas standard-feedback-trained models hold up (Appendix~\ref{app:feedback_impact}). Prompt richness produces no substantive difference in aggregate, but the aggregate hides an asymmetry. Full-info-trained models score $0.36$ higher on full-info eval than on some-info eval, whereas some-info-trained models are essentially indifferent to the eval prompt. Some-info training thus yields a more uniformly transferable model, at the cost of the in-distribution advantage full-info training enjoys on full-info eval. This is the kind of training-time question a controllable testbed can answer. The same sequence-level rollouts also make the per-task rewards $r_1,\dots,r_N$ available within a single rollout, which opens up algorithm variants that use them for better credit assignment and sample efficiency.

\subsection{Research agenda}
Our results open several directions for future work.

\begin{enumerate}
    \item \emph{Scaling environments and latents.} Generalization in LLM fine-tuning tends to improve with data scale; the analogue in our setting is the diversity of environments and latent families used during training. Expanding the testbed beyond our seven text-based environments, particularly to settings closer to real deployments such as customer support, personalization, and tool-use agents, would test whether the meta-strategies learned by Cross-Task RL transfer to more naturalistic forms of shared structure.

    \item \emph{Algorithmic design.} Our study fixes the objective at the cumulative reward $R=\sum_i r_i$; alternative summaries (final-task reward, monotonicity-rewarded variants, or risk-averse aggregations) may induce qualitatively different adaptation policies, and a systematic study of how this choice shapes the exploration vs.\ exploitation trade-off remains open. Our finding that sparser training feedback transfers more robustly also suggests revisiting other forms of training-time information shaping, including curricula over latents, prompt schedules, and staged feedback regimes.

    \item \emph{Scaling the model.} Our experiments use Qwen3-8B; whether Cross-Task RL's gains compound or saturate with larger backbones, and how its sample efficiency scales, are natural next questions.

    \item \emph{Diagnostic use of the framework.} Beyond training, the controllable-latent design supports targeted diagnostic studies of any agent. By varying latent difficulty, prompt informativeness, and feedback density independently, one can localize where a specific agent fails, whether in gathering evidence, in identifying the latent, or in acting on it, in a way that aggregate benchmark scores cannot.
\end{enumerate}

We hope the framework, together with the open-source training and evaluation pipeline, serves as a starting point for principled study of cross-task experiential learning in LLM agents.

\section*{Acknowledgements}

This research used resources of the Oak Ridge Leadership Computing Facility (OLCF) and Argonne Leadership Computing Facility (ALCF)] which are a DOE Office of Science User Facility. This work was supported by an award from the ASCR Leadership Computing Challenge (ALCC) under project ERCAP0034861.

\bibliographystyle{abbrvnat}

\ifdefined\useorstyle
\setlength{\bibsep}{.0em}
\else
\setlength{\bibsep}{.7em}
\fi


\newpage
\begin{appendix}

\newpage


\section{Environment Details}
\label{app:envs}

We use seven text-game environments, each instantiated as a
multi-task sequence by sampling a latent $z\sim p(z)$ once per sequence
and then sampling task-specific variation $\xi_i$ for each task
$\tau_i^z$ in the sequence. For every environment we list (i) the
single-task game; (ii) the multi-task / sequenceing version; (iii) the
full latent catalogue, and the prompt template
shown to the agent at task start. Reward, observation, and action
space conventions are identical across latents within an environment.


\subsection{Number Guessing}
\label{app:env-ng}

\paragraph{Single-task game.}
In this game the agent's task is to identify a hidden integer
$y \in [\ell, u]$ that the environment has chosen uniformly at
random. On each turn the agent submits a guess $g_t \in [\ell, u]$
and is told whether the hidden number is greater than, less than, or
equal to the guess. This comparison is the only intermediate signal
the agent receives. The episode ends as soon as the agent guesses
correctly; otherwise it terminates after $T_{\max}=30$ turns.

\paragraph{Reward.}
The per-task reward is decay-shaped so that the agent is
incentivised to solve the puzzle quickly. If the agent guesses
correctly on turn $t$, it receives
\[
    R \;=\; \max\!\bigl(0,\; 1 - 0.020 \cdot t\bigr),
\]
and zero if it never guesses correctly within the turn cap. With
$T_{\max}=30$ and the per-turn decay of $0.020$, the reward is at
least $0.40$ for any solution found within the cap and approaches
$1.0$ as $t \to 1$.

\paragraph{Multi-task version.}
In the multi-task version the agent plays a sequence of $N=10$ number
guessing tasks back-to-back. A latent variable $z$, sampled once at
the beginning of the sequence and held fixed throughout, governs how
the targets $y_1, \dots, y_N$ are correlated across tasks (for
instance, all targets may be drawn from the same fixed three-element
set). The agent's context carries the full cross-task history
$h_{<i}$ — every guess–feedback pair from the preceding tasks
together with the inter-task transition messages — so an attentive
agent can in principle infer $z$ within a few tasks and exploit it
on the rest.

\paragraph{Latents.}
We use five latents in our experiments:
\begin{itemize}
    \item \texttt{set\_of\_2} — all targets in the sequence are drawn
          from the same set of two specific numbers in $[1, 1000]$.
    \item \texttt{set\_of\_3} — all targets in the sequence are drawn
          from the same set of three specific numbers in
          $[1, 1000]$.
    \item \texttt{range\_100} — all targets fall within a fixed
          contiguous $100$-number sub-range of $[1, 1000]$.
    \item \texttt{dynamic\_range} — targets fall within a contiguous
          $1000$-number sub-range of a dynamic outer range that
          shifts deterministically across tasks.
    \item \texttt{two\_ranges} — targets are drawn from two
          disjoint $500$-number ranges within $[1, 10000]$.
\end{itemize}

\paragraph{Additional latents.}
For completeness, the codebase also registers two further latents
that we do not evaluate in this paper:
\texttt{range\_1000} (targets fall within a fixed contiguous
$1000$-number range of $[1, 10000]$) and
\texttt{dynamic\_full\_range} (the full $[\ell, u]$ prompt range
itself slides as $[n, n+1000]$ where $n$ varies across tasks in the
sequence). We retain these in the framework so that future work can
study cross-task adaptation when the prompt-visible range itself
becomes part of the latent.

\begin{tcolorbox}[colback=gray!5,colframe=gray!50,
                  title={Environment Prompt (no info)},
                  fonttitle=\bfseries,breakable]
\small
You are playing 10 number guessing games sequentially. In each game,
you need to guess an integer (whole number, not a decimal) between
1 and 1000 (inclusive).

\medskip
For each guess, I will tell you if the number is greater than, less
than, or equal to your guess. Your goal is to guess the number
correctly in as few turns as possible.

\medskip
Your guess must be wrapped in square brackets. Format your guess as:
\texttt{[number]} (e.g., \texttt{[500]}).
\end{tcolorbox}

\noindent\textbf{some\_info adds:} The numbers in these games
might follow a pattern. Pay attention to the numbers you encounter
across different games, as this might help you improve your
strategy.\\
\textbf{full\_info adds (set\_of\_3 example):} IMPORTANT: All
the numbers in this series of games are drawn from a set of 3
specific numbers. Use the information from previous games to identify
this set and improve your strategy.

\subsection{Bandits}
\label{app:env-bandits}

\paragraph{Single-task game.}
In this game the agent's task is to identify the highest-reward arm
out of $K=5$ buttons (named \texttt{red}, \texttt{blue},
\texttt{green}, \texttt{yellow}, \texttt{purple}). Each button has
a hidden Bernoulli reward probability $p_a \in [0, 1]$ that is
fixed within a task. On each turn the agent can either
\textbf{explore} a button — pulling it and observing a binary
reward $r_t \sim \mathrm{Bern}(p_{a_t})$ — or \textbf{commit} by
selecting a button as its final answer. The intermediate signal the
agent receives is the per-turn binary reward from each pull. The
episode ends as soon as the agent commits, or after $T_{\max}=30$
turns, in which case the last explored button is treated as the
commitment.

\paragraph{Reward.}
The per-task reward depends only on the correctness of the
commitment, with a per-turn decay that incentivises early
commitment. If the committed button is the true highest-probability
arm at the time of commitment $t$, the agent receives
\[
    R \;=\; \max\!\bigl(0,\; 1 - 0.015 \cdot t\bigr),
\]
and zero otherwise. The intermediate Bernoulli rewards observed
during exploration are informative but do not directly contribute
to the per-task score.

\paragraph{Multi-task version.}
In the multi-task version the agent plays a sequence of $N=10$ bandit
tasks back-to-back. The latent $z$, sampled once per sequence,
governs how the arm-reward vector $(p_0, \dots, p_4)$ is correlated
across tasks (for instance, the same arm may always be the best, or
the best arm may rotate through the indices in a predictable
order). The agent's context retains every prior pull-and-reward
pair across tasks, so an attentive agent can in principle infer $z$
from a few tasks and commit much earlier on the rest.

\paragraph{Latents.}
We use four latents in our experiments:
\begin{itemize}
    \item \texttt{loyal\_favorite\_0} — a single fixed arm is the
          highest-reward arm in every task of the sequence.
    \item \texttt{top\_two\_fixed} — the same pair of arms always
          occupies the top two positions; the agent only needs to
          explore between them.
    \item \texttt{even\_indices\_only} — the best arm is always at
          an even index ($0$, $2$, or $4$).
    \item \texttt{ping\_pong} — the best arm oscillates between the
          first and last index across tasks
          ($0\!\to\!4\!\to\!0\!\to\!4\,\dots$).
\end{itemize}

\paragraph{Additional latents.}
For completeness, the codebase registers a much larger family of
bandit latents that we do not evaluate in this paper. These include
position-fixed variants (\texttt{fixed\_probabilities},
\texttt{one\_hot}, \texttt{bottom\_excluded},
\texttt{odd\_indices\_only}); rotation/cycle variants
(\texttt{clockwise\_rotation}, \texttt{counterclockwise\_rotation},
\texttt{swap\_top\_two}, \texttt{cycle\_length\_5});
binary-switch variants (\texttt{binary\_switch\_0\_1},
\texttt{binary\_switch\_0\_last}); cross-episode memory variants
(\texttt{shadow}, \texttt{cold\_hand}, \texttt{hot\_hand},
\texttt{random\_walk}, \texttt{mirror\_mode}); mathematical-sequence
variants (\texttt{skip\_2}, \texttt{skip\_3}, \texttt{fibonacci},
\texttt{prime\_indices}, \texttt{triangular}); and the
\texttt{same\_ranking} variant in which the ordering of arms is
fixed across the sequence while their exact probabilities vary. We
retain these in the framework so that future work can study
cross-task adaptation in richer bandit settings.

\begin{tcolorbox}[colback=gray!5,colframe=gray!50,
                  title=Environment Prompt (no info),
                  fonttitle=\bfseries,breakable]
\small
You are playing a Multi-Armed Bandit game with 5 buttons (red, blue,
green, yellow, purple). Each button has a hidden probability of
giving a reward (0 or 1). You have up to 30 turns.

\medskip
On each turn you can either:
\begin{itemize}
\item Explore: \texttt{[button\_name]} (e.g., \texttt{[red]}) ---
press a button and observe the reward.
\item Select: \texttt{[select button\_name]} (e.g.,
\texttt{[select red]}) --- lock in your final answer.
\end{itemize}

Once you select, the game ends immediately. Your goal is to identify
the button with the highest probability. Selecting correctly earlier
gives a higher reward. If you don't select by the last turn, your
last explore action counts as your selection.

\medskip
IMPORTANT: Only use square brackets \texttt{[...]} to submit your
action for this turn (either \texttt{[button\_name]} to explore or
\texttt{[select button\_name]} to finalize). Do not use square
brackets anywhere else in your response.

\medskip
You will play 10 rounds of this game sequentially.
\end{tcolorbox}

\noindent\textbf{some\_info adds:} The reward probabilities
across these rounds may share a hidden pattern or constraint. Pay
attention to which buttons perform best across games.\\
\textbf{full\_info adds (loyal\_favorite\_0 example):}
IMPORTANT: There is a consistent pattern of buttons across
games. Pay attention to which button wins.

\subsection{Secretary Problem}
\label{app:env-secretary}

\paragraph{Single-task game.}
In this game the agent's task is to pick the best of $n=20$
candidates that arrive one at a time. Each candidate $t$ is
revealed with a numeric score $s_t \in [0, 1]$, and after seeing it
the agent must irrevocably either accept the candidate (ending the
episode immediately) or continue to the next one. Rejected
candidates cannot be revisited; if the agent reaches the end of the
sequence without having accepted, it is forced to accept the last
candidate. The intermediate signal the agent receives during a task
is exactly the running sequence of revealed scores
$s_1, s_2, \dots$ — there is no comparison feedback or hint about
the underlying score distribution.

\paragraph{Reward.}
The per-task reward depends on the score of the accepted candidate
relative to the true maximum of the full sequence
$s_{\max} = \max_t s_t$. If the agent accepts the maximum, it
receives the full reward of $1.0$; otherwise it receives a partial
credit proportional to how close the accepted score is to the
maximum:
\[
    R \;=\; \begin{cases}
        1.0 & \text{if } s_{\text{accepted}} = s_{\max}, \\
        0.5 \cdot \dfrac{s_{\text{accepted}}}{s_{\max}}
            & \text{otherwise}.
    \end{cases}
\]
A random-stopping baseline averages roughly $R \approx 0.26$ under
this scheme; an agent that always commits on a near-best candidate
($s_{\text{accepted}} / s_{\max} \approx 0.95$) earns $R \approx 0.47$
when it is wrong, well below the $1.0$ available for picking the
true maximum.

\paragraph{Multi-task version.}
In the multi-task version the agent plays a sequence of $N=10$
secretary tasks back-to-back. The latent $z$, sampled once per
sequence, governs how the position or score of the winning candidate
is correlated across tasks (for instance, the winner may always
appear at a fixed position, or its score may always exceed a hidden
threshold). The agent's context retains the full
score-sequence-and-decision history from preceding tasks, so an
attentive agent can in principle infer $z$ within a few tasks and
exploit it on the rest.

\paragraph{Latents.}
We use four latents in our experiments:
\begin{itemize}
    \item \texttt{inverse\_order} — candidate scores arrive in
          strictly decreasing order, so the first candidate is
          always the best (decreasing).
    \item \texttt{fixed\_position\_2} — the winning candidate
          always appears at position 2 of the sequence.
    \item \texttt{threshold\_06} — the maximum is always above
          $0.6$ and all other candidates are below it; the agent
          learns to accept any value above the threshold.
    \item \texttt{sorted\_order} — candidate scores arrive in
          strictly increasing order; the optimal policy is to
          always continue to the last candidate (used as the OOD
          held-out latent).
\end{itemize}

\paragraph{Additional latents.}
For completeness, the codebase registers a much larger family of
secretary latents that we do not evaluate in this paper, covering
position-fixed, threshold, position-shift, cycle, structural,
relational, cross-episode memory, and mathematical-position
families. We retain these in the framework so that future work can
study cross-task adaptation in richer secretary settings.

\begin{tcolorbox}[colback=gray!5,colframe=gray!50,
                  title=Enviornment Prompt (no info),
                  fonttitle=\bfseries,breakable]
\small
You are playing the Secretary Problem (optimal stopping). You will
see 20 values one at a time. For each value, decide:
\texttt{[accept]} to claim it, or \texttt{[continue]} to skip and
see the next. Once you accept, the game ends. If you reach the last
value, you must accept it. You win (reward 1.0) if the value you
accept is the maximum of all values. Otherwise you get partial
credit based on how close your accepted value is to the maximum.

\medskip
IMPORTANT: Output your action as \texttt{[accept]} or
\texttt{[continue]}. Square brackets must be used ONLY for the final
action --- never elsewhere in your response.

\medskip
You will play 10 rounds of this game sequentially.
\end{tcolorbox}

\noindent\textbf{some\_info adds:} The value sequences may
have a pattern in where the maximum appears. Pay attention to
previous rounds to improve your strategy.\\
\textbf{full\_info adds (threshold\_06 example):} IMPORTANT:
Across games there is a consistent pattern. There is a value
threshold --- the maximum is always above it and other values are
below. Discover the threshold to accept immediately.

\subsection{Mastermind}
\label{app:env-mastermind}

\paragraph{Single-task game.}
In this game the agent's task is to crack a hidden 4-digit code,
where each digit is independently drawn from $\{1, \dots, 6\}$
(repeats allowed). On each turn the agent submits a length-4 guess
and receives the standard Mastermind feedback as the intermediate
signal: a count of \emph{black pegs} (digits that are correct in
both value and position) and a count of \emph{white pegs} (digits
that are correct in value but in the wrong position). The episode
ends as soon as the agent submits the exact code (all four black
pegs) or after $T_{\max}=10$ guesses.

\paragraph{Reward.}
The per-task reward measures how close the final guess was to the
hidden code, computed as the fraction of pegs that were correct in
both value and position at the time the episode terminated:
\[
    R \;=\; \frac{\text{black pegs on final guess}}{4}.
\]
A correct solve gives $R = 1.0$; a complete miss gives $R = 0.0$.
Crucially, this reward depends only on the \emph{final} guess and
not on the trajectory of intermediate guesses, so the agent is
rewarded for converging to the correct code rather than for
incremental progress along the way.

\paragraph{Multi-task version.}
In the multi-task version the agent plays a sequence of $N=10$
Mastermind tasks back-to-back. The latent $z$, sampled once per
sequence, constrains the family of codes that appear across the
sequence — for instance, all codes might be strictly ascending
sequences, or all might begin with a fixed digit. The agent's
context retains the full guess-and-feedback history from preceding
tasks, so an attentive agent can in principle infer $z$ within a
few tasks and use it to drastically narrow the search space on the
remaining ones.

\paragraph{Latents.}
We use four latents in our experiments:
\begin{itemize}
    \item \texttt{strictly\_ascending} — every hidden code is a
          strictly ascending sequence of four distinct digits.
    \item \texttt{strictly\_descending} — every hidden code is a
          strictly descending sequence of four distinct digits.
    \item \texttt{first\_is\_6} — the first digit of every hidden
          code is fixed to $6$.
    \item \texttt{has\_pair} — every hidden code contains at least
          one repeated digit (used as the OOD held-out latent).
\end{itemize}

\paragraph{Additional latents.}
For completeness, the codebase registers a much larger family of
mastermind latents that we do not evaluate in this paper, covering
weakly monotonic patterns, palindromic and alternating structures,
parity constraints, digit-set constraints, repeat / no-repeat
constraints, and adjacency rules. We retain these in the framework
so that future work can study cross-task adaptation under richer
code constraints.

\begin{tcolorbox}[colback=gray!5,colframe=gray!50,
                  title=Environment Prompt (no info),
                  fonttitle=\bfseries,breakable]
\small
You are playing Mastermind. A secret code of 4 digits has been
chosen, each digit between 1 and 6. Each turn, guess the code using
format \texttt{[1 2 3 4]}. You'll receive feedback: black pegs
(correct digit in correct position) and white pegs (correct digit in
wrong position). Win by finding the exact code.

\medskip
IMPORTANT: Only use square brackets \texttt{[X X X X]} for your
actual guess submission. Do not use square brackets when discussing
or referencing previous guesses.

\medskip
You will play 10 rounds of Mastermind sequentially.
\end{tcolorbox}

\noindent\textbf{some\_info adds:} The secret codes may share
a common pattern. Pay attention to the codes you discover across
games.\\
\textbf{full\_info adds (strictly\_ascending example):} All
secret codes share a hidden structural pattern (e.g., ascending
order, palindrome, specific digit constraint). Discover this pattern
to crack codes faster in later rounds.

\subsection{Word Ladder}
\label{app:env-wordladder}

\paragraph{Single-task game.}
In this game the agent's task is to transform a given start word
into a given target word of the same length by changing exactly one
letter at a time, where every intermediate word in the chain must
itself be a valid English word from the supplied dictionary. On
each turn the agent submits the next word in the chain; the
intermediate signal it receives is whether the proposed word is
accepted into the chain (i.e.\ differs from the previous word by
exactly one letter and is in the dictionary) along with the
updated current word. The episode ends as soon as the agent reaches
the target word or after $T_{\max}=20$ turns.

\paragraph{Reward.}
The per-task reward distinguishes solved from unsolved tasks. If
the agent reaches the target word, it receives a decay-shaped
reward that penalises taking more turns than the optimal solution:
\[
    r \;=\; \max\!\bigl(0,\; 1 - 0.03 \cdot (t - t^*)\bigr),
\]
where $t$ is the number of turns used and $t^*$ is the length of
the shortest valid chain from the start word to the target. A
solution that matches the optimal length earns $R = 1.0$; each
extra turn beyond optimal costs $0.03$, so for example a 3-turn
overshoot earns $R = 0.91$ and a 10-turn overshoot earns
$R = 0.70$. If the agent fails to reach the target within the turn
cap, the reward is the fraction of letter positions in the current
final word that match the target:
\[
    r \;=\; \frac{\text{positions matching target}}{|\text{target}|}.
\]
This partial credit gives a smooth signal even on failed attempts.

\paragraph{Multi-task version.}
In the multi-task version the agent plays a sequence of $N=10$ Word
Ladder tasks back-to-back. The latent $z$, sampled once per sequence,
constrains either the family of valid intermediate words or the
structure of the optimal solution path (for instance, every optimal
chain may pass through a fixed pivot word, or every chain may only
substitute consonants). The agent's context retains the full
solution-chain history from preceding tasks, so an attentive agent
can in principle infer $z$ within a few tasks and use it to
shortcut the search on the rest.

\paragraph{Latents.}
We use four latents in our experiments:
\begin{itemize}
    \item \texttt{hub\_word\_3letter} — every (start, target) pair
          in the sequence admits a shortest-path solution that passes
          through a fixed 3-letter ``hub'' word. Discovering the
          hub splits each puzzle into two easier sub-problems.
    \item \texttt{hub\_word\_4letter} — same as above, with a fixed
          4-letter hub word.
    \item \texttt{order\_outside\_in} — at least one optimal path
          changes outer letter positions first and inner positions
          last.
    \item \texttt{order\_left\_to\_right} — at least one optimal
          path changes letter positions strictly left-to-right
          (used as the OOD held-out latent).
\end{itemize}

\paragraph{Additional latents.}
For completeness, the codebase registers a much larger family of
word-ladder latents that we do not evaluate in this paper, covering
restricted-vocabulary paths, alternative position-order strategies
(right-to-left), substitution-type constraints (vowel-only,
consonant-only, alternating, phonetic-group), word-family
constraints (paths that stay within the words containing a given
substring or sharing a fixed suffix/prefix), and skeleton
constraints on the start/target pair. We retain these in the
framework so that future work can study cross-task adaptation under
richer Word Ladder structures.

\begin{tcolorbox}[colback=gray!5,colframe=gray!50,
                  title= Environment Prompt (no info),
                  fonttitle=\bfseries,breakable]
\small
You are playing a series of 10 Word Ladder games. In each game, you
are given a start word and a target word. Transform the start word
into the target word by changing exactly one letter at a time. Each
intermediate word must be a valid English word.

\medskip
You may reason about your strategy, but your answer must be wrapped
in square brackets. Format your guess as: \texttt{[word]} (e.g.,
\texttt{[cold]}).
\end{tcolorbox}

\noindent\textbf{some\_info adds:} The word pairs and their
solutions across games may share common properties or patterns. Pay
attention to the words that appear in your solutions --- this might
help you solve future games faster.\\
\textbf{full\_info adds (hub\_word\_4letter example):}
IMPORTANT: There is a pattern across games. All the word pairs
in this series can be solved by going through a specific intermediate
`hub' word. Discover this hub word and use it to split every puzzle
into two easier sub-problems.

\subsection{Wordle}
\label{app:env-wordle}

\paragraph{Single-task game.}
In this game the agent's task is to identify a hidden 5-letter
English word that the environment has chosen from a fixed
dictionary. On each turn the agent submits a 5-letter guess; the
intermediate signal it receives is the standard Wordle feedback,
returned per-letter as one of three colours: green (the letter is
correct in both identity and position), yellow (the letter appears
in the target but at a different position), or grey (the letter
does not appear in the target at all). The episode ends as soon as
the agent submits the exact target word or after $T_{\max}=6$
guesses.

\paragraph{Reward.}
The per-task reward is binary at the trajectory level and depends
on whether the agent identified the target within the turn cap. If
the agent guesses the target exactly, it receives the full reward
of $1.0$; otherwise it receives a partial credit equal to the
fraction of letters that ended up correctly identified at the
correct position by the final guess (the per-letter
\emph{percentage completion} returned by the underlying TextArena
environment). A complete miss yields $R = 0$.

\paragraph{Multi-task version.}
In the multi-task version the agent plays a sequence of $N=10$ Wordle
puzzles back-to-back. The latent $z$, sampled once per sequence,
constrains the family of target words used across the sequence — for
instance, every target may use only high-frequency English letters,
or every target may avoid repeated letters. The agent's context
retains the full guess-and-feedback history from preceding tasks,
so an attentive agent can in principle infer $z$ within a few
tasks and use it to seed strong opening guesses on the rest.

\paragraph{Latents.}
We use four latents in our experiments:
\begin{itemize}
    \item \texttt{letter\_freq\_high} — every target word uses only
          high-frequency English letters (average letter frequency
          $> 7\%$).
    \item \texttt{letter\_freq\_medium} — every target word uses
          letters of roughly average English frequency (4--7\%).
    \item \texttt{mixed\_balanced} — every target word uses a
          balanced mix of vowels and consonants drawn evenly from
          across the alphabet.
    \item \texttt{no\_repeated\_letters} — every target word
          contains five distinct letters.
\end{itemize}

\paragraph{Additional latents (designed but not used).}
For completeness, the codebase registers a much larger family of
Wordle latents that we do not evaluate in this paper, covering
vowel-count and vowel-position constraints, position-fixed letter
constraints, alphabet-half restrictions, consonant-vs-vowel pattern
templates (CVCVC, CVCCV, VCCVC), specific-letter ``contains'' /
``starts-with'' / ``ends-with'' patterns, double-letter
constraints, and semantic-category restrictions. We retain these
in the framework so that future work can study cross-task
adaptation under richer Wordle target distributions.

\begin{tcolorbox}[colback=gray!5,colframe=gray!50,
                  title= Environment Prompt  (no info),
                  fonttitle=\bfseries,breakable]
\small
You are playing Wordle. A secret 5-letter word has been chosen. You
have 6 attempts to guess it.

\medskip
For each guess, wrap your word in square brackets (e.g.,
\texttt{[apple]}). Feedback for each letter:
\begin{itemize}
\item G (green): correct letter in the correct position.
\item Y (yellow): letter exists in the word but in the wrong position.
\item X (wrong): letter is not in the word.
\end{itemize}

IMPORTANT: Output your guess as \texttt{[word]}. Square brackets must
be used ONLY for the final guess --- never elsewhere in your
response.

\medskip
You will play 10 rounds of this game sequentially.
\end{tcolorbox}

\noindent\textbf{some\_info adds:} The secret words across
rounds may share a hidden pattern. Pay attention to common features
of the target words.\\
\textbf{full\_info adds (letter\_freq\_high example):}
IMPORTANT: Average letter frequency $>7\%$. Use this knowledge
to guess more efficiently in later rounds. (Hint string is the
\texttt{letter\_freq\_high} latent's \texttt{description} field, fetched
via \texttt{get\_latent("wordle", latent\_id).description}.)

\subsection{Hangman}
\label{app:env-hangman}

\paragraph{Single-task game.}
In this game the agent's task is to identify a hidden English word
that the environment has chosen from a fixed dictionary. On each
turn the agent submits either a single letter or a complete word
guess. The intermediate signal is the partially-revealed board: a
guessed letter that appears in the target is unmasked at every
matching position, while a guess that is not in the target costs
the agent one of its $T_{\max}=6$ allowed wrong guesses. A
complete word guess wins immediately if correct or burns one wrong
attempt if not. The episode ends as soon as the entire word is
revealed, the agent submits the correct word, or the wrong-guess
budget reaches zero.

\paragraph{Reward.}
The per-task reward is binary on a clean win and a fraction of
revealed letters otherwise. If the agent reveals the entire word
(letter-by-letter or via a correct full-word guess), it receives
the full reward of $1.0$. If it exhausts the wrong-guess budget,
it receives partial credit equal to the fraction of unique target
letters that had been correctly revealed by the time the episode
ended,
\[
    r \;=\; \frac{\text{distinct target letters revealed}}{\text{total distinct letters in target}}.
\]
This gives a smooth signal even on failed attempts and rewards the
agent for narrowing down the target before running out of guesses.

\paragraph{Multi-task version.}
In the multi-task version the agent plays a sequence of $N=10$
Hangman tasks back-to-back. The latent $z$, sampled once per
sequence, constrains the family of hidden words that appear across
the sequence — for instance, every target may share a suffix, a
vowel count, or a semantic category. The agent's context retains
the full guess-and-board history from preceding tasks, so an
attentive agent can in principle infer $z$ within a few tasks and
use it to seed strong opening guesses on the rest.

\paragraph{Latents.}
We use four latents in our experiments:
\begin{itemize}
    \item \texttt{vowel\_count\_4} — every target word contains
          exactly four vowels.
    \item \texttt{ending\_ABLE} — every target word ends in the
          suffix ``ABLE''.
    \item \texttt{has\_double\_letter} — every target word contains
          a pair of consecutive identical letters
          (e.g.\ \texttt{LL}, \texttt{EE}, \texttt{SS}).
    \item \texttt{starts\_with\_S} — every target word begins with
          the letter \texttt{S}.
\end{itemize}

\paragraph{Additional latents (designed but not used).}
For completeness, the codebase registers a much larger family of
Hangman latents that we do not evaluate in this paper, covering
word-length constraints, alternate vowel counts and vowel/consonant
ratios, semantic categories (e.g.\ ``physical objects''), letter-
frequency-score classes, ``contains-letter'' / ``avoids-letter''
constraints, alternate suffix and consonant-cluster patterns, and
overall difficulty bands (easy/medium/hard). We retain these in
the framework so that future work can study cross-task adaptation
under richer Hangman target distributions.

\begin{tcolorbox}[colback=gray!5,colframe=gray!50,
                  title= Environment Prompt (no info),
                  fonttitle=\bfseries,breakable]
\small
You are playing Hangman. A secret word has been chosen. Each turn,
guess a letter using \texttt{[L]} format (e.g., \texttt{[A]}) or
guess the full word using \texttt{[WORD]} format (e.g.,
\texttt{[APPLE]}). Correct letters are revealed on the board. Wrong
guesses cost an attempt. Win by revealing all letters or guessing the
word before running out of attempts.

\medskip
IMPORTANT: Only use square brackets \texttt{[...]} to submit your
actual guess. Do not use square brackets anywhere else in your
response.

\medskip
You will play 10 rounds of Hangman sequentially.
\end{tcolorbox}

\noindent\textbf{some\_info adds:} The words may share common
properties. Pay attention to patterns across games to improve your
guessing strategy.\\
\textbf{full\_info adds (ending\_ABLE example):} All target
words share a hidden property (e.g., same length, same starting
letter, same vowel pattern). Discover this property from early rounds
to guess more efficiently in later rounds.



\subsection{Feedback types (orthogonal to prompt).}
After every task in the sequence the agent receives one of:
\begin{itemize}
    \item \texttt{standard} --- only the per-task reward signal
          (success / fraction-correct), identical to what the agent
          already observes during the task.
    \item \texttt{information} --- the per-task reward \emph{plus}
          the ground-truth target / code / winning candidate from the
          just-finished task. This lets the agent verify or revise
          its hypothesis about the latent~$z$.
\end{itemize}
The feedback string is appended to the cross-task context $h_{<i}$
before task $i{+}1$ begins; the prompt condition (no / some / full
info) remains unchanged across tasks within a sequence.

\newpage
\section{Experimental Details}
\label{app:experimental-details}

\paragraph{Base model and infrastructure.}
We use \texttt{Qwen3-8B} as our base model for all RL
fine-tunes. Our training pipeline is built on top of SkyRL, with rollouts generated through a colocated
\texttt{vLLM} inference server. Each run uses $4\times$ NVIDIA
A100 80GB GPUs. We optimise the policy with GRPO, using the
standard clipped objective and a KL penalty against a frozen
reference policy $\pi_{\text{ref}}$.

\paragraph{Training-time prompt and feedback condition.}
Unless stated otherwise, all RL fine-tunes in this paper are run
under the \texttt{some\_info} prompt condition (vague hint that a
cross-task pattern may exist across tasks) and the
\texttt{standard} feedback condition (the agent receives only the
per-task reward between tasks). 

\paragraph{Shared hyperparameters across all RL runs.}
We tune the GRPO hyperparameters once on the Number Guessing
environment and then fix them for every other experiment in this
paper. Final values are listed in Table~\ref{tab:rl-hp}. The only
across-experiment variation is the sequence length per rollout
($N{=}10$ tasks for cross-task runs vs.\ $N{=}1$ for single-task
runs).

\begin{table}[h]
\centering
\small
\caption{Shared GRPO hyperparameters across all RL fine-tunes.}
\label{tab:rl-hp}
\begin{tabular}{lc}
\toprule
Hyperparameter & Value \\
\midrule
Optimizer                            & AdamW ($\beta_1{=}0.9$, $\beta_2{=}0.999$) \\
Policy learning rate                 & $5\times10^{-7}$ \\
Weight decay                         & $0.01$ \\
KL coefficient $\beta$               & $0.04$ \\
Clip range $\epsilon$                & $0.2$ \\
Group size (rollouts per prompt)     & $8$ \\
Train batch size (\texttt{tbs})      & $32$ prompts \\
Mini-batch size (\texttt{mbs})       & $16$ \\
Samples per prompt (\texttt{nsp})    & $8$ \\
Max response tokens                  & $256$ \\
sequence length per rollout            & $N{=}10$ tasks (\texttt{ne=10}) or $N{=}1$ task (\texttt{ne=1}) \\
Reward functional $\Phi$             & $\sum_i G_i$ (cumulative return) \\
Reference policy                     & frozen Qwen3-8B \\
Sampling temperature (training)      & $0.8$ \\
Sampling top-$p$ (training)          & $0.95$ \\
Sampling temperature (eval)          & $0.70$ \\
Sampling top-$p$ (eval)              & $1.0$ \\
Random seed                          & $263\ ([263, 42, 1729] \ \text{if multiple})$ \\
\bottomrule
\end{tabular}
\end{table}

\subsection{Single-environment Experiments}

\paragraph{Training.}
For each of the five core environments (Number Guessing, Mastermind,
Hangman, Word Ladder, Secretary) we train two variants from the
shared Qwen3-8B base: a cross-task model
(\texttt{ne=10}) and a single-task model (\texttt{ne=1}). Both
variants train on the three in-distribution latents listed below
for each environment.

\begin{itemize}
    \item Number Guessing — \texttt{set\_of\_3},
          \texttt{set\_of\_2}, \texttt{dynamic\_range}.
    \item Mastermind — \texttt{strictly\_ascending},
          \texttt{strictly\_descending}, \texttt{first\_is\_6}.
    \item Hangman — \texttt{vowel\_count\_4},
          \texttt{has\_double\_letter}, \texttt{starts\_with\_S}.
    \item Word Ladder — \texttt{hub\_word\_4letter},
          \texttt{hub\_word\_3letter}, \texttt{order\_outside\_in}.
    \item Secretary — \texttt{inverse\_order},
          \texttt{fixed\_position\_2}, \texttt{sorted\_order}.
\end{itemize}

Each training step samples a batch of $32$ prompts and generates
$8$ rollouts per prompt, where one rollout is a complete sequence of
$N$ tasks. At every step the latent of each prompt is uniformly
sampled from the three training latents. cross-task models
therefore see $32 \times 8 \times 10 = 2{,}560$ task-episodes per
step, while single-task models see $32 \times 8 = 256$. Training
proceeds for $\sim$$1{,}000$ GRPO steps per run; we checkpoint
every 15 steps and select the best checkpoint by validation
cumulative return on a held-out seed.

\paragraph{Evaluation.}
We evaluate each fine-tuned variant — and the base
Qwen3-8B model for comparison — under the same conditions
used during training (\texttt{some\_info} prompt and
\texttt{standard} feedback). For every environment we evaluate on
four latents: the three in-distribution latents seen during
training, together with one held-out (OOD-1) latent. The held-out
latents are:
\begin{itemize}
    \item Number Guessing — \texttt{range\_100}.
    \item Mastermind — \texttt{has\_pair}.
    \item Hangman — \texttt{ending\_ABLE}.
    \item Word Ladder — \texttt{order\_left\_to\_right}.
    \item Secretary — \texttt{best\_is\_last}.
\end{itemize}
We run $50$ trajectories of task-sequences per (model, environment,
latent) cell, for a total of $200$ trajectories per environment for each model.

\subsection{Multi-environment Experiments}
\label{app:multi-env}
\paragraph{Training.}
A single cross-task model
and a
single single-task model
are trained on
sequences sampled from the union of four core environments: Number
Guessing, Hangman, Word Ladder, and Secretary. Each sequence's
environment is held fixed across its $N=10$ tasks (within-sequence
variation is over the latent's task-specific $\xi_i$, not over the
environment), and the sequence's latent is uniformly sampled from
that environment's two in-distribution training latents listed
below:
\begin{itemize}
    \item Number Guessing — \texttt{set\_of\_3},
          \texttt{range\_100}.
    \item Hangman — \texttt{vowel\_count\_4},
          \texttt{ending\_ABLE}.
    \item Word Ladder — \texttt{hub\_word\_4letter},
          \texttt{hub\_word\_3letter}.
    \item Secretary — \texttt{inverse\_order},
          \texttt{fixed\_position\_2}.
\end{itemize}
Total training latents: $4 \text{ envs} \times 2 \text{ latents} =
8$. Training proceeds for $\sim$$850$ GRPO steps with the same
shared hyperparameters.

\paragraph{Evaluation.}
We evaluate the two multi-environment fine-tuned variants — and the
base Qwen3-8B model for comparison — with \texttt{some\_info} prompt and
\texttt{standard} feedback. Note that these are the same training conditions as the model trained on $N=10$ while the single-task model is trained on \texttt{some\_info} given that it does not access the previous history. Each model is evaluated on every one
of the four core environments separately, on the two
in-distribution training latents per environment listed above, with
$50$ trajectories of $N=10$ tasks per (model, environment, latent)
cell, for a total of $100$ trajectories per environment.

\subsection{Multi-environment Prompt and Feedback Sweep Experiments}
\label{app:prompt-feedback}

\paragraph{Training.}
We use the same setup as the multi-environment experiments above
(four core environments — Number Guessing, Hangman, Word Ladder,
Secretary — with the same two in-distribution training latents per
environment listed in Section~\ref{app:multi-env}, for a total of
$4 \text{ envs} \times 2 \text{ latents} = 8$ training latents).
We train only cross-task models (\texttt{ne=10}); for each of the
four (prompt, feedback) combinations
\[
(\text{prompt}, \text{feedback}) \in
  \{\texttt{some\_info}, \texttt{full\_info}\}
  \times
  \{\texttt{standard}, \texttt{information}\},
\]
we train three independent models with seeds
$\{42, 263, 1729\}$, for a total of
$4 \text{ combos} \times 3 \text{ seeds} = 12$ models. Training
proceeds for $\sim$$450$ GRPO steps per model with the same shared
hyperparameters.

\paragraph{Evaluation.}
Each fine-tuned model — and the base Qwen3-8B for comparison — is
evaluated on every one of the four prompt $\times$ feedback
combinations (yielding both in-distribution and cross-condition
cells), on every one of the four core environments, on the two
in-distribution training latents per environment, with $50$
trajectories of $N=10$ tasks per
(model, environment, latent, eval-prompt, eval-feedback) cell.
This gives $4 \text{ envs} \times 2 \text{ latents} \times 4 \text{ eval combos}
= 32$ eval cells per model and a total of $1{,}600$ trajectories per
model.

\subsection{Leave-one-out Experiments}

\paragraph{Training.}
We use the same setup as the multi-environment experiments above,
except that one environment $e$ is held out at a time. For each
held-out $e \in \{$NG, hangman, wordladder, secretary$\}$ we train
one cross-task model
on sequences sampled from the \emph{other three} environments, using
the same two in-distribution training latents per training
environment listed in Section~\ref{app:multi-env} (total
$3 \text{ envs} \times 2 \text{ latents} = 6$ training latents per
leave-one-out model). The held-out environment $e$ itself is never
seen at training time. 
Training proceeds for $\sim$$650$
GRPO steps per leave-one-out model with the same shared
hyperparameters.

\paragraph{Evaluation.}
We use the same evaluation setup as the multi-environment
experiments above (\texttt{some\_info} prompt, \texttt{standard}
feedback, $50$ trajectories of $N=10$ tasks per cell). The only
change is that each leave-one-out model — and the base
Qwen3-8B for comparison — is evaluated only on its
held-out environment $e$, on the two in-distribution latents of
that environment, for a total of $100$ trajectories per held-out
environment. This is the OOD-2 setting reported in the paper.


\subsection{Two-latent Single-environment Prompt Experiments}
\label{app:two-latent}

\paragraph{Training.}
For each of three environments — Number Guessing, Word Ladder,
Secretary — we train one cross-task model (\texttt{ne=10}) on
sequences sampled from \emph{two} in-distribution training latents
per environment (six models total, but each trained
independently on its own environment). The two-latent sets are:
\begin{itemize}
    \item Number Guessing — \texttt{set\_of\_3}, \texttt{range\_100}.
    \item Word Ladder — \texttt{hub\_word\_4letter},
          \texttt{restricted\_vocab\_4letter}.
    \item Secretary — \texttt{fixed\_position\_2},
          \texttt{increasing\_position}.
\end{itemize}
At every training step the latent of each prompt is uniformly
sampled from the two training latents of the corresponding
environment. Training uses the \texttt{no\_info} prompt and
\texttt{standard} feedback, and proceeds for $\sim$$320$ GRPO
steps per model with the same shared hyperparameters.

\paragraph{Evaluation.}
We evaluate each fine-tuned model — and the base Qwen3-8B for
comparison — under the same conditions used during training
(\texttt{no\_info} prompt, \texttt{standard} feedback), on the
same two in-distribution latents seen during training, with $50$
trajectories of $N=10$ tasks per (model, latent) cell, for a
total of $100$ trajectories per environment.

\subsection{Data Collection vs.\ Data Utilization Experiments}

This experiment uses the  trained models from the
multi-environment experiments (Section~\ref{app:multi-env}) — the
base model, the cross-task multi-env model, and the single-task
multi-env model — without further fine-tuning.

\paragraph{Evaluation.}
We use the same evaluation setup as the multi-environment
experiments (Section~\ref{app:multi-env}; \texttt{some\_info}
prompt, \texttt{standard} feedback, the two in-distribution
latents per environment, $50$ trajectories of $N=10$ tasks per
cell). For each model pair $(A, B) \in \{$(base, single-task),
(base, cross-task), (single-task, cross-task)$\}$ and each
switch point $K \in \{2, 4, 6, 8\}$, we run two directional
rollouts:
\begin{itemize}
    \item $A \rightarrow B$: model $A$ acts during episodes
          $1, \ldots, K$ and collects the trajectory; model $B$
          inherits the cross-task context $h_{<K}$ and acts during
          episodes $K{+}1, \ldots, 10$.
    \item $B \rightarrow A$: the reverse.
\end{itemize}
Each model pair, switch point, and direction is evaluated on every
one of the four core environments separately. Solo baselines come
from each model's standalone single-agent eval on the same grid.
All reported $\Delta$ values average across the two
in-distribution latents within each environment.

\subsection{Compute (Qwen3-8B Training and Evaluation)}
\label{app:compute}

We report cumulative wall-clock compute for all Qwen3-8B
fine-tunes and post-training evaluations in A100 80GB GPU-hours.

\paragraph{Training (RL fine-tunes).}
\begin{itemize}
    \item Per-environment experiments — $5$ environments $\times$
          $2$ variants (\texttt{ne=10} and \texttt{ne=1}) $=$ $10$
          fine-tunes. Total: $\sim$$1{,}630$ A100-hours.
    \item Multi-environment experiments — $2$ fine-tunes (one
          cross-task, one single-task). Total: $\sim$$330$
          A100-hours.
    \item Leave-one-out experiments — $4$ leave-one-out fine-tunes.
          Total: $\sim$$270$ A100-hours.
    \item Prompt and feedback sweep experiments
          (Section~\ref{app:prompt-feedback}) — $4$ prompt $\times$
          feedback combinations $\times$ $3$ seeds $=$ $12$
          cross-task fine-tunes. Total: $\sim$$2{,}300$ A100-hours.
    \item \textbf{Training subtotal:} $\sim$$4{,}530$ A100-hours.
\end{itemize}

\paragraph{Evaluation (Qwen3-8B post-training).}
Each fine-tuned variant and the base Qwen3-8B are
evaluated on their respective grids ($50$ trajectories per cell,
$N=10$ tasks per trajectory):
\begin{itemize}
    \item Per-environment evals ($10$ fine-tuned variants $+$ base
          across $5$ environments): $\sim$$190$ A100-hours.
    \item Multi-env, leave-one-out, and switch-point sweeps:
          $\sim$$140$ A100-hours.
    \item \textbf{Evaluation subtotal:} $\sim$$330$ A100-hours.
\end{itemize}

\paragraph{Total.}
Training and evaluation of all Qwen3-8B models in this paper
consumed approximately $\sim$$5{,}860$ A100 80GB GPU-hours.


\subsection{Frontier-Model Evaluation}
\label{app:frontier-eval}

\paragraph{Models.}
We evaluate three production frontier models accessed through
OpenRouter:
\begin{itemize}
    \item \texttt{openai/gpt-4o},
    \item \texttt{anthropic/claude-sonnet-4.6},
    \item \texttt{google/gemini-2.5-flash}.
\end{itemize}
All three are queried at sampling temperature $0.7$,
\texttt{max\_tokens}$=4096$ per response, with default top-$p$.
Routing to OpenRouter lets us evaluate every model through a single
unified API and avoids per-provider client differences.

\paragraph{Evaluation grid.}
Unlike the trained-model evaluations, the frontier-model evaluation
exercises the full prompt $\times$ feedback grid in order to test
each model under all of \texttt{no\_info}, \texttt{some\_info}, and
\texttt{full\_info} prompts and both \texttt{standard} and
\texttt{information} feedback conditions. For every (model,
environment) pair we evaluate on $4$ latents $\times$ $3$ prompts
$\times$ $2$ feedbacks $\times$ $5$ trajectories
$=$ $120$ trajectories per environment. The four eval latents are the four canonical latents
per environment listed in Section~\ref{app:envs}.

\newpage

\section{Additional Results}
\label{app:additional_results}


\subsection{ Exploration efficiency vs. Exploitation efficiency}
\label{app:data_collection_utilization}
 
In this appendix we show additional results for the \emph{exploration vs.\ exploitation efficiency} analysis in Section~\ref{sec:experiments} for Number Guessing. For all experiments we follow the same agent-switching protocol. First, we compare Cross-Task RL against the \emph{base} Qwen3-8B model on Number Guessing, alongside the corresponding agent-switch trajectory plot. Second, we report the Cross-Task RL vs.\ Single-task RL agent-switching tables for three additional environments (Hangman, Secretary, and Word Ladder) to test whether the exploration/exploitation decomposition behaves consistently across the benchmark.

\begin{figure}[h!]
    \centering
    \includegraphics[width=\linewidth]{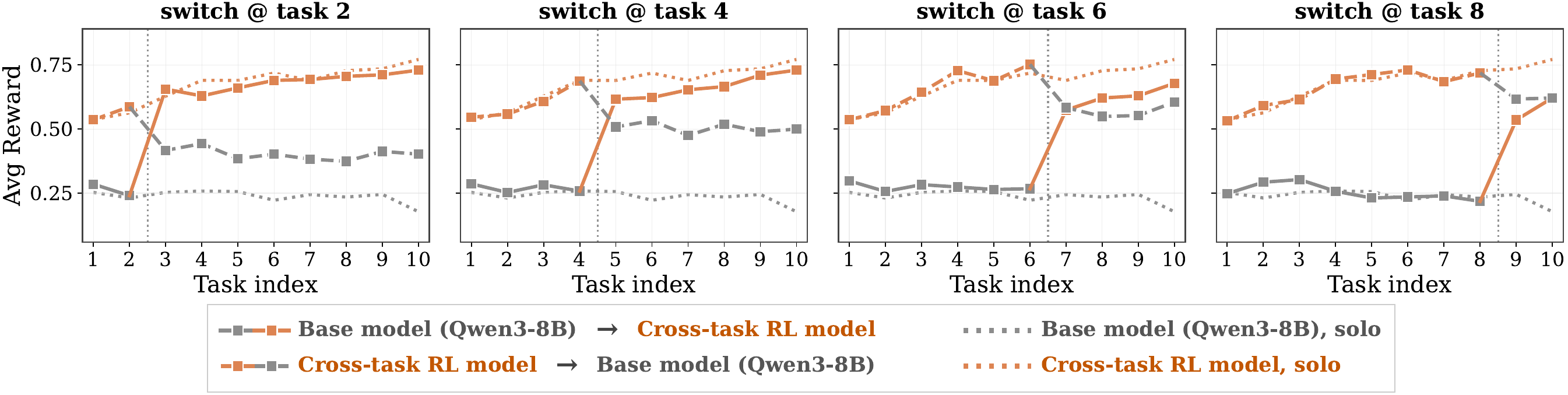}
    \caption{\textbf{Agent-switching: sequence RL vs.\ base Qwen3-8B on Number Guessing.} The same asymmetry holds: sequence RL recovers most of its solo trajectory even when the early context was built by the base model, while the base model cannot exploit a context built by sequence RL. }
    \label{fig:switch_sequenceRL_vs_base}
\end{figure}

\paragraph{Number Guessing: Cross-Task RL vs.\ base.} Table~\ref{tab:ng-sequence-vs-base} reports the same protocol as Section~\ref{sec:experiments} but with the base Qwen3-8B model in place of Single-task RL. Both gaps are dramatically larger than against Single-task RL, reaching $+192.8\%$ on exploration and $+189.6\%$ on exploitation. Figure~\ref{fig:switch_sequenceRL_vs_base} shows the corresponding trajectories.

\paragraph{Hangman.} Table~\ref{tab:hangman-collection-utilization} shows that on Hangman, Cross-Task RL leads on both axes, with exploration advantages of $+20\%$ to $+62\%$ and exploitation advantages of $+6\%$ to $+34\%$. Unlike Number Guessing, exploration is the larger of the two on this environment, and it grows with $K$ while exploitation shrinks.

\paragraph{Secretary.} Table~\ref{tab:secretary-collection-utilization} shows a more balanced picture: exploration advantages of $+9\%$ to $+40\%$ and exploitation advantages of $+12\%$ to $+33\%$, with neither axis dominating consistently across switch points.

\paragraph{Word Ladder.} Table~\ref{tab:wordladder-collection-utilization} shows that on Word Ladder both gaps are small (under $\pm 4\%$), reflecting the convergence of Cross-Task RL and Single-task RL on this environment noted in Section~\ref{sec:experiments}.

Two patterns are consistent across environments. First, Cross-Task RL generally posts non-negative advantages on \emph{both} exploration and exploitation, demonstrating that it improves both capabilities. Second, which axis dominates is environment-dependent: exploitation on Number Guessing, exploration on Hangman, mixed on Secretary, neither on Word Ladder. This suggests that Cross-Task RL addresses different bottlenecks depending on the structure of the task.

\begin{table}[h!]
\centering
\small
\setlength{\tabcolsep}{8pt}
\renewcommand{\arraystretch}{1.15}
\begin{tabular}{c c cc}
\toprule
Switch point $K$ & Reference $C$ & $\mathrm{ExploreGain}_C$ & $\mathrm{ExploitGain}_C$ \\
\midrule
\multirow{2}{*}{2} & $C=\mathrm{CT}$    & $+3.2\%$   & $+75.5\%$  \\
                   & $C=\mathrm{Base}$  & $+70.1\%$  & $+189.4\%$ \\
\midrule
\multirow{2}{*}{4} & $C=\mathrm{CT}$    & $+8.3\%$   & $+43.2\%$  \\
                   & $C=\mathrm{Base}$  & $+119.1\%$ & $+189.6\%$ \\
\midrule
\multirow{2}{*}{6} & $C=\mathrm{CT}$    & $+16.8\%$  & $+27.6\%$  \\
                   & $C=\mathrm{Base}$  & $+153.9\%$ & $+177.5\%$ \\
\midrule
\multirow{2}{*}{8} & $C=\mathrm{CT}$    & $+30.2\%$  & $+21.7\%$  \\
                   & $C=\mathrm{Base}$  & $+192.8\%$ & $+173.6\%$ \\
\bottomrule
\end{tabular}
\caption{\textbf{Cross-Task RL vs.\ Base model on \texttt{number\_guessing}} (4-latent average). Cross-Task RL is dramatically better at both: large positive values throughout, with the largest gaps appearing when the Base is held fixed in the other role.}
\label{tab:ng-sequence-vs-base}
\end{table}

\begin{table}[h!]
\centering
\small
\setlength{\tabcolsep}{8pt}
\renewcommand{\arraystretch}{1.15}
\begin{tabular}{c c cc}
\toprule
Switch point $K$ & Reference $C$ & $\mathrm{ExploreGain}_C$ & $\mathrm{ExploitGain}_C$ \\
\midrule
\multirow{2}{*}{2} & $C=\mathrm{CT}$   & $+20.2\%$ & $+26.6\%$ \\
                   & $C=\mathrm{ST}$  & $+27.1\%$ & $+33.8\%$ \\
\midrule
\multirow{2}{*}{4} & $C=\mathrm{CT}$   & $+30.2\%$ & $+12.2\%$ \\
                   & $C=\mathrm{ST}$  & $+44.7\%$ & $+24.7\%$ \\
\midrule
\multirow{2}{*}{6} & $C=\mathrm{CT}$   & $+36.9\%$ & $+7.6\%$  \\
                   & $C=\mathrm{ST}$  & $+53.0\%$ & $+20.3\%$ \\
\midrule
\multirow{2}{*}{8} & $C=\mathrm{CT}$   & $+48.0\%$ & $+5.8\%$  \\
                   & $C=\mathrm{ST}$  & $+61.6\%$ & $+15.5\%$ \\
\bottomrule
\end{tabular}
\caption{\textbf{Cross-Task RL vs.\ Single-task RL on \texttt{hangman}} (4-latent average). All values positive: Cross-Task RL is both a better explorer and a better exploiter than Single-task RL, with the exploration advantage growing in $K$ and the exploitation advantage shrinking as the post-switch tail shrinks.}
\label{tab:hangman-collection-utilization}
\end{table}

\begin{table}[h!]
\centering
\small
\setlength{\tabcolsep}{8pt}
\renewcommand{\arraystretch}{1.15}
\begin{tabular}{c c cc}
\toprule
Switch point $K$ & Reference $C$ & $\mathrm{ExploreGain}_C$ & $\mathrm{ExploitGain}_C$ \\
\midrule
\multirow{2}{*}{2} & $C=\mathrm{CT}$   & $+12.2\%$ & $+29.1\%$ \\
                   & $C=\mathrm{ST}$  & $+8.6\%$  & $+25.0\%$ \\
\midrule
\multirow{2}{*}{4} & $C=\mathrm{CT}$   & $+18.2\%$ & $+23.6\%$ \\
                   & $C=\mathrm{ST}$  & $+15.4\%$ & $+20.7\%$ \\
\midrule
\multirow{2}{*}{6} & $C=\mathrm{CT}$   & $+21.5\%$ & $+12.1\%$ \\
                   & $C=\mathrm{ST}$  & $+31.6\%$ & $+21.4\%$ \\
\midrule
\multirow{2}{*}{8} & $C=\mathrm{CT}$   & $+23.0\%$ & $+16.7\%$ \\
                   & $C=\mathrm{ST}$  & $+39.9\%$ & $+32.7\%$ \\
\bottomrule
\end{tabular}
\caption{\textbf{Cross-Task RL vs.\ Single-task RL on \texttt{secretary}} (4-latent average). All values positive: Cross-Task RL is both a better explorer and a better exploiter than Single-task RL, with the exploration advantage growing steadily in $K$.}
\label{tab:secretary-collection-utilization}
\end{table}

\begin{table}[h!]
\centering
\small
\setlength{\tabcolsep}{8pt}
\renewcommand{\arraystretch}{1.15}
\begin{tabular}{c c cc}
\toprule
Switch point $K$ & Reference $C$ & $\mathrm{ExploreGain}_C$ & $\mathrm{ExploitGain}_C$ \\
\midrule
\multirow{2}{*}{2} & $C=\mathrm{CT}$   & $-0.1\%$ & $+2.4\%$ \\
                   & $C=\mathrm{ST}$  & $-0.3\%$ & $+2.2\%$ \\
\midrule
\multirow{2}{*}{4} & $C=\mathrm{CT}$   & $-0.5\%$ & $+1.4\%$ \\
                   & $C=\mathrm{ST}$  & $+0.7\%$ & $+2.6\%$ \\
\midrule
\multirow{2}{*}{6} & $C=\mathrm{CT}$   & $-0.2\%$ & $+2.8\%$ \\
                   & $C=\mathrm{ST}$  & $-0.2\%$ & $+2.8\%$ \\
\midrule
\multirow{2}{*}{8} & $C=\mathrm{CT}$   & $-1.6\%$ & $+3.3\%$ \\
                   & $C=\mathrm{ST}$  & $-1.2\%$ & $+3.7\%$ \\
\bottomrule
\end{tabular}
\caption{\textbf{Cross-Task RL vs.\ Single-task RL on \texttt{wordladder}} (4-latent average). Differences are uniformly small on this environment ($\leq 4\%$ throughout) and exploration efficiencies are near zero or slightly negative: both RL variants converge to similar trajectories, leaving little room for cross-task adaptation to matter.}
\label{tab:wordladder-collection-utilization}
\end{table}

\newpage

\subsection{Feedback impact on training and generalization}
\label{app:feedback_impact}
We train Qwen3-8B on four (prompt, feedback) combinations:
\{{full info},{some info}\}$\times$\{information, standard\}, with three random seeds each (12 runs total) and evaluate every checkpoint on the full $4\times 8$ grid of (eval prompt, eval feedback)$\times$(8 env--latents). All curves below are mean $\pm$ 1 standard error of the mean across seed samples; the eval cells contributing to each curve are averaged together as described in the captions. Tables~\ref{tab:v1_4envs_pf_sweep_prompt_2x2} and~\ref{tab:v1_4envs_pf_sweep_feedback_2x2} report the corresponding final-step values.

\begin{figure}[h!]
\centering
\includegraphics[width=0.65\textwidth]{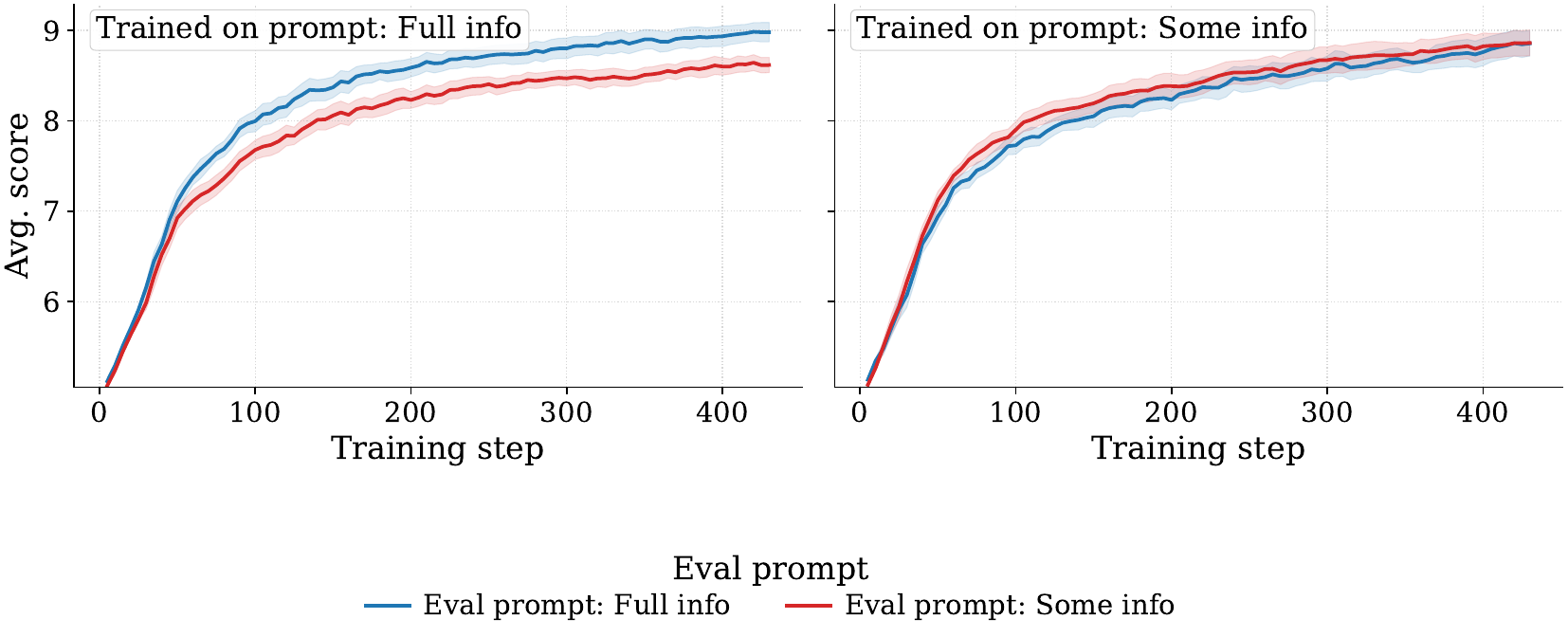}
\caption{For each training prompt (panels), eval-time score under the two eval
prompts (lines), pooled over both the feedbacks. Lines are mean $\pm$ 1 SEM
across $n=6$ training runs (2~feedbacks $\times$ 3~seeds).}
\label{fig:pf_sweep_prompt_by_train}
\end{figure}

\begin{figure}[h!]
\centering
\includegraphics[width=0.65\textwidth]{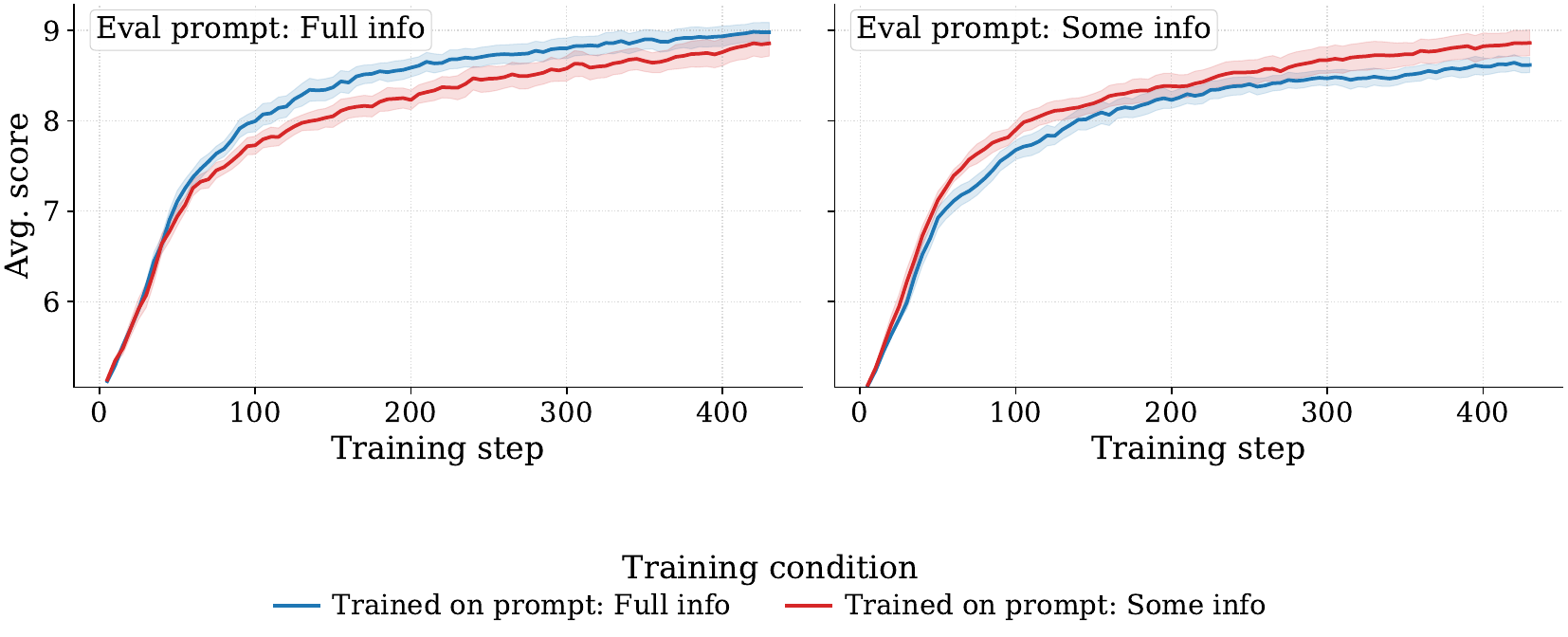}
\caption{Roles reversed from Figure~\ref{fig:pf_sweep_prompt_by_train}: each panel
fixes an eval prompt and compares the two training prompts. Same aggregation
and band semantics.}
\label{fig:pf_sweep_prompt_by_eval}
\end{figure}

\paragraph{Effect of the training prompt.}
Figures~\ref{fig:pf_sweep_prompt_by_train} and~\ref{fig:pf_sweep_prompt_by_eval} present the same data with the panel/line roles swapped. In Figure~\ref{fig:pf_sweep_prompt_by_train} each panel fixes a training prompt and overlays the two eval prompts; full-info-trained models display a clear in-distribution advantage (the blue ``Eval: Full info'' curve sits well above ``Eval: Some info''), whereas some-info-trained models track essentially the same trajectory under both eval prompts. In Figure~\ref{fig:pf_sweep_prompt_by_eval} each panel fixes an eval prompt and overlays the two training prompts; full-info-trained models lead on full-info eval, while on some-info eval the some-info-trained model is slightly ahead. The takeaway is that training under the more informative prompt buys a sizeable in-distribution gain but a modest cross-prompt deficit, while training under the less informative prompt produces a more robust agent that pays only a small price under the matched-prompt condition.

\begin{table}[h]
\centering
\begin{tabular}{l|cc}
\toprule
 & \multicolumn{2}{c}{\textbf{Eval prompt}} \\
\textbf{Train prompt} & Full info & Some info \\
\midrule
Full info  & $8.979 \pm 0.107$ & $8.616 \pm 0.084$ \\
Some info  & $8.855 \pm 0.140$ & $8.860 \pm 0.143$ \\
\bottomrule
\end{tabular}
\caption{Average score (cumulative reward summed over 10 episodes) at final
training step, broken down by training prompt vs.\ evaluation prompt.
Each cell is the mean $\pm$ 1 standard error  across $n=6$
independent training runs (2 training feedbacks $\times$ 3 seeds), averaged
over both eval feedbacks and 8 env--latents.}
\label{tab:v1_4envs_pf_sweep_prompt_2x2}
\end{table}

\begin{figure}[h!]
\centering
\includegraphics[width=0.65\linewidth]{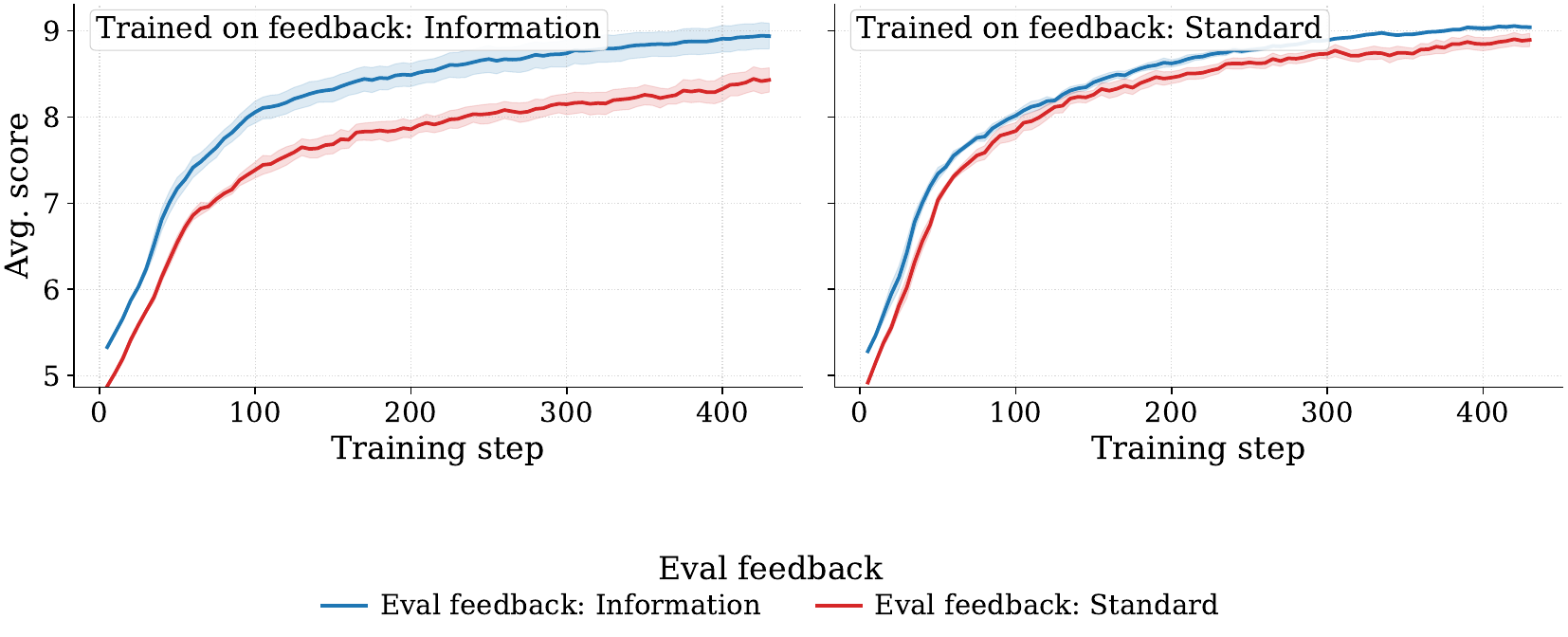}
\caption{For each training feedback (panels), eval-time score under the two eval feedbacks (lines), pooled over both training prompts, both eval prompts, and 8 env--latents. Lines are mean $\pm$ 1 standard error across $n=6$ training runs.}
\label{fig:pf_sweep_fb_by_train}
\end{figure}

\begin{figure}[h!]
\centering
\includegraphics[width=0.65\linewidth]{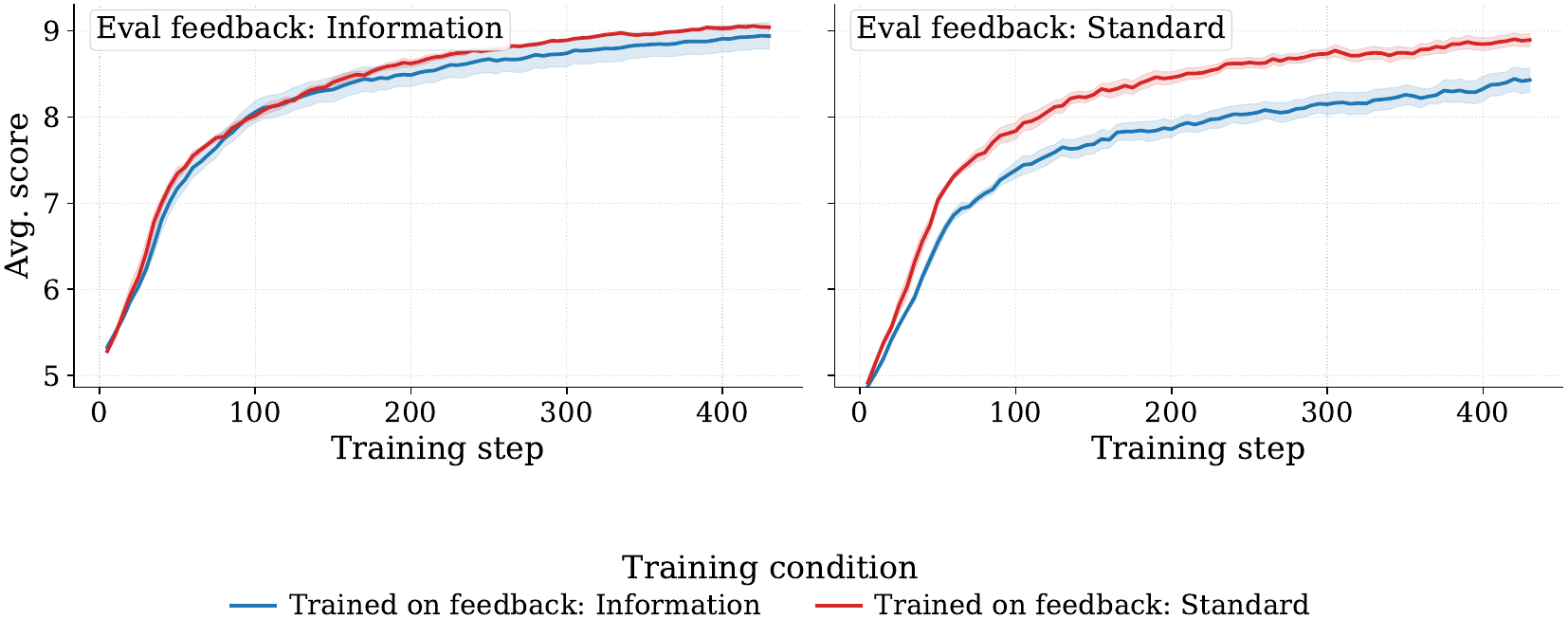}
\caption{Roles reversed: each panel fixes an eval feedback and compares the two training feedbacks. Same aggregation and band semantics.}
\label{fig:pf_sweep_fb_by_eval}
\end{figure}

\paragraph{Effect of the training feedback.}
Figures~\ref{fig:pf_sweep_fb_by_train} and~\ref{fig:pf_sweep_fb_by_eval} repeat the analysis along the feedback dimension. The pattern is asymmetric: under both eval feedbacks, the standard-feedback-trained model is above the information-feedback-trained model, and the gap between matched- and mismatched-eval-feedback within an information-trained model ($8.940 \to 8.431$) is much larger than within a standard-trained model ($9.044 \to 8.895$). Training with extra information feedback therefore looks helpful when the same feedback is available at eval time but degrades transfer to the standard-feedback regime, whereas training under the standard feedback transfers well.

\begin{table}[h]
\centering
\begin{tabular}{l|cc}
\toprule
 & \multicolumn{2}{c}{\textbf{Eval feedback}} \\
\textbf{Train feedback} & Information & Standard \\
\midrule
Information & $8.940 \pm 0.149$ & $8.431 \pm 0.140$ \\
Standard    & $9.044 \pm 0.020$ & $8.895 \pm 0.078$ \\
\bottomrule
\end{tabular}
\caption{Average score at final training step, broken down by training feedback
vs.\ evaluation feedback. Each cell is the mean $\pm$ 1 standard error across $n=6$ independent training runs (2 training prompts $\times$ 3
seeds), averaged over both eval prompts and 8 env--latents.}
\label{tab:v1_4envs_pf_sweep_feedback_2x2}
\end{table}

\begin{figure}[h!]
\centering
\includegraphics[width=0.75\linewidth]{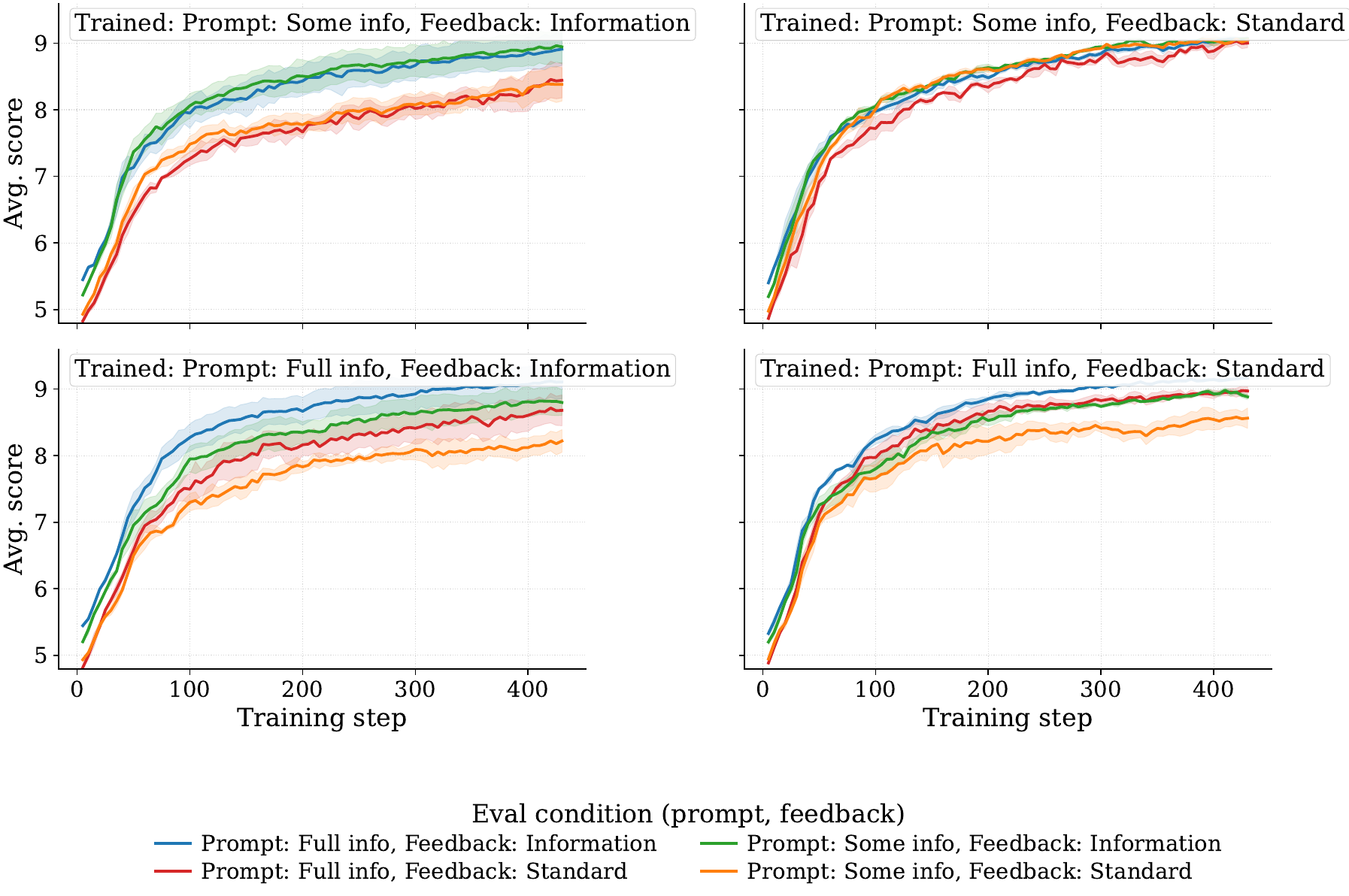}
\caption{For each training (prompt, feedback) combination (panels), eval-time
score under all four eval (prompt, feedback) combinations (lines), averaged
over the 8 env--latents. Lines are mean $\pm$ 1 standard error across $n=3$ seeds.}
\label{fig:pf_sweep_combo_by_train}
\end{figure}

\begin{figure}[h!]
\centering
\includegraphics[width=0.75\linewidth]{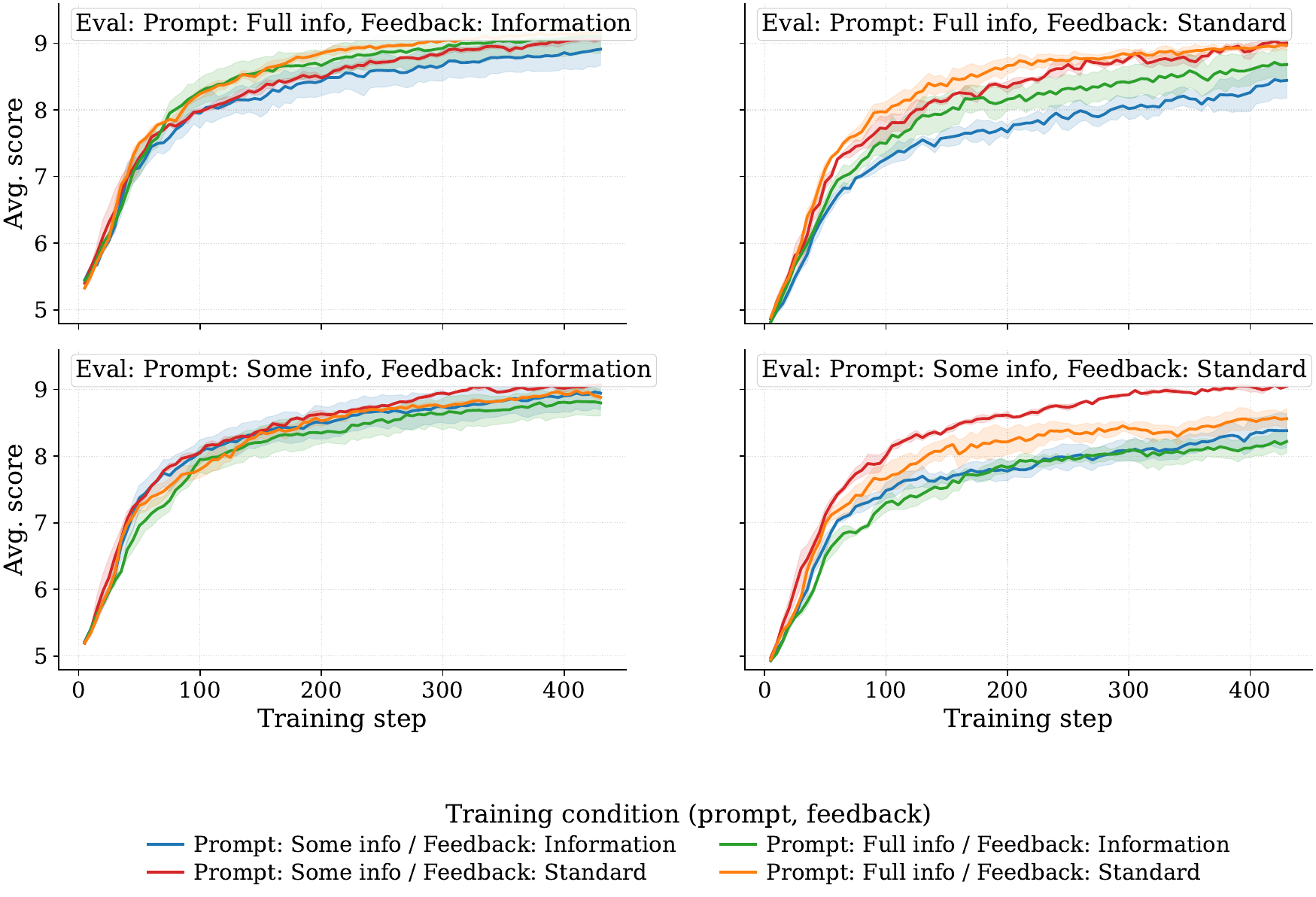}
\caption{Roles reversed from Figure~\ref{fig:pf_sweep_combo_by_train}: each panel
fixes an eval (prompt, feedback) combination and compares the four training
combinations. Same aggregation and band semantics.}
\label{fig:pf_sweep_combo_by_eval}
\end{figure}

\paragraph{Joint (prompt, feedback) view.}
Figures~\ref{fig:pf_sweep_combo_by_train} and~\ref{fig:pf_sweep_combo_by_eval} avoid pooling over either training axis and present the full $4\times 4$ table of training$\times$eval (prompt, feedback) combinations. Figure~\ref{fig:pf_sweep_combo_by_train} shows the four-way trajectory of a single trained model across all eval conditions; on the other hand Figure~\ref{fig:pf_sweep_combo_by_eval} shows, for a fixed eval condition, which training condition reaches the highest score. The dominant factor is the eval feedback itself: information-feedback
evaluation produces higher scores than standard-feedback evaluation for every
training combination, with gaps ranging from $+0.04$ to $+0.52$. Holding the eval condition fixed, no
single training condition wins everywhere, but {(some info, standard)}
is the top-ranked training condition in three of the four eval panels
(Figure~\ref{fig:pf_sweep_combo_by_eval}); the exception is the
{(full info, information)} eval panel, where \textit{(full info,
standard)} ranks first. Prompt alignment has a secondary effect
(matched-prompt cells average $0.18$ points above mismatched-prompt
cells). 

\clearpage

\newpage
\section{Anecdotes for Failures of SOTA models}
\label{app:anecdotes}

In this appendix we present a representative set of anecdotes in
which state-of-the-art models fail. Each anecdote, organised under
its failure category, describes the agent's task, the hidden
latent variable, the ground-truth values, trajectory-level metrics
(number of turns and rewards), and the portion of the conversation
that exposes the failure. The accompanying multi-turn transcript
is colour-coded across three panels: the agent's chain of thought
(when exposed) in purple, the agent's action in blue, and the
game master in green. The anecdotes cover failures across three
models: GPT-4o, Gemini~2.5~Flash, and Claude~Sonnet~4.6.


\newpage
\subsection{Adaptation Neglect}
\label{app:adaptation_neglect}

\begin{anecdotebox}{GPT-4o does not try to identify any pattern}{GPT4oColor}
\noindent\textbf{Category:} Failure to adapt across tasks\par
\noindent\textbf{Model:} GPT-4o\par
\noindent\textbf{Configuration:} Environment: number\_guessing. Latent: set\_of\_3. Prompt: no\_info. Feedback: standard. Trajectory: 0.\par\smallskip
\textbf{Setup.} Targets are drawn from a fixed three-element set on every task. After 2--3 tasks a model could open with one of the previously revealed targets. GPT-4o opens tasks 1, 5 and 9 with the same midpoint guess.\par\smallskip
\textbf{What went wrong.} No use of cross-task evidence; first guess is always the midpoint of the full search range.\par\smallskip
\textbf{Per-task summary} (\textit{$\star$ marks tasks whose transcript is shown below}).\par
\noindent Agent task is to identify a hidden integer between 1 and 1000; after each guess, the game master replies whether the target is greater than, less than, or equal to the guess.\par
\par\noindent\textit{Hidden latent: set = \{592, 781, 926\}}\par\smallskip
\noindent\footnotesize
\begin{tabularx}{\linewidth}{@{}>{\bfseries}p{1.6cm}>{\raggedright\arraybackslash}X>{\raggedright\arraybackslash}X>{\raggedright\arraybackslash}X>{\raggedright\arraybackslash}X>{\raggedright\arraybackslash}X>{\raggedright\arraybackslash}X>{\raggedright\arraybackslash}X>{\raggedright\arraybackslash}X>{\raggedright\arraybackslash}X>{\raggedright\arraybackslash}X@{}}
\toprule
Task & \textbf{$\star$1} & \textbf{2} & \textbf{3} & \textbf{$\star$4} & \textbf{5} & \textbf{6} & \textbf{7} & \textbf{8} & \textbf{$\star$9} & \textbf{10} \\
\midrule
Answer & 781 & 592 & 926 & 592 & 926 & 592 & 926 & 926 & 592 & 781 \\
Turns & 5 & 10 & 8 & 9 & 8 & 9 & 8 & 8 & 9 & 5 \\
Reward & 0.90 & 0.80 & 0.84 & 0.82 & 0.84 & 0.82 & 0.84 & 0.84 & 0.82 & 0.90 \\
\bottomrule
\end{tabularx}
\normalsize
\par\smallskip
\par\smallskip\noindent\textit{Showing Task 1, Task 4 and Task 9 out of 10 tasks.}\par\smallskip
\noindent\begin{tabularx}{\linewidth}{@{}X@{\hspace{4pt}}X@{\hspace{4pt}}X@{}}
\begin{taskbox}{Task 1}{GPT4oColor}
\agentmsg{1}{Let's start with a guess right in the middle to narrow it down quickly: [500].}
\gmmsg{The number is greater than 500.}
\agentmsg{2}{Let's try the midpoint of the remaining range: [750].}
\gmmsg{The number is greater than 750.}
\par\smallskip\noindent\textcolor{gray}{\scriptsize\itshape \dots\ continues (showing 2 of 5 turns)}\par
\end{taskbox} & \begin{taskbox}{Task 4}{GPT4oColor}
\agentmsg{1}{Let's start with the middle: [500].}
\gmmsg{The number is greater than 500.}
\agentmsg{2}{Let's try [750].}
\gmmsg{The number is less than 750.}
\par\smallskip\noindent\textcolor{gray}{\scriptsize\itshape \dots\ continues (showing 2 of 9 turns)}\par
\end{taskbox} & \begin{taskbox}{Task 9}{GPT4oColor}
\agentmsg{1}{Let's start with the middle: [500].}
\gmmsg{The number is greater than 500.}
\agentmsg{2}{Let's try [750].}
\gmmsg{The number is less than 750.}
\par\smallskip\noindent\textcolor{gray}{\scriptsize\itshape \dots\ continues (showing 2 of 9 turns)}\par
\end{taskbox}
\\
\end{tabularx}
\end{anecdotebox}

\newpage
\begin{anecdotebox}{Gemini does not try to identify any pattern}{GeminiColor}
\noindent\textbf{Category:} Failure to adapt across tasks\par
\noindent\textbf{Model:} Gemini 2.5 Flash\par
\noindent\textbf{Configuration:} Environment: number\_guessing. Latent: range\_100. Prompt: no\_info. Feedback: standard. Trajectory: 0.\par\smallskip
\textbf{Setup.} Range is fixed across tasks. Gemini's openings on tasks 1, 5 and 9 do not narrow.\par\smallskip
\textbf{What went wrong.} Cross-task range structure not internalized.\par\smallskip
\textbf{Per-task summary} (\textit{$\star$ marks tasks whose transcript is shown below}).\par
\noindent Agent task is to identify a hidden integer between 1 and 1000; after each guess, the game master replies whether the target is greater than, less than, or equal to the guess.\par
\par\noindent\textit{Hidden latent: range = [1, 100]}\par\smallskip
\noindent\footnotesize
\begin{tabularx}{\linewidth}{@{}>{\bfseries}p{1.6cm}>{\raggedright\arraybackslash}X>{\raggedright\arraybackslash}X>{\raggedright\arraybackslash}X>{\raggedright\arraybackslash}X>{\raggedright\arraybackslash}X>{\raggedright\arraybackslash}X>{\raggedright\arraybackslash}X>{\raggedright\arraybackslash}X>{\raggedright\arraybackslash}X>{\raggedright\arraybackslash}X@{}}
\toprule
Task & \textbf{$\star$1} & \textbf{2} & \textbf{3} & \textbf{4} & \textbf{$\star$5} & \textbf{6} & \textbf{7} & \textbf{$\star$8} & \textbf{9} & \textbf{10} \\
\midrule
Answer & 689 & 630 & 596 & 680 & 607 & 682 & 595 & 675 & 690 & 672 \\
Turns & 10 & 9 & 9 & 7 & 10 & 9 & 10 & 10 & 9 & 6 \\
Reward & 0.80 & 0.82 & 0.82 & 0.86 & 0.80 & 0.82 & 0.80 & 0.80 & 0.82 & 0.88 \\
\bottomrule
\end{tabularx}
\normalsize
\noindent\begin{tabularx}{\linewidth}{@{}X@{\hspace{4pt}}X@{\hspace{4pt}}X@{}}
\begin{taskbox}{Task 1}{GeminiColor}
\thinkmsg{1}{**Analyzing Game Parameters**  I've started to unpack the number-guessing game's constraints. Considering a binary search... {\\\relax}\  **Calculating First Move**  Now, I'm setting the first guess. Based on the rules, and the initial range, I'm leaning toward 500.…}
\agentmsg{1}{[500]}
\gmmsg{The number is greater than 500.}
\thinkmsg{2}{**Calculating Next Guess**   I've narrowed the range down after the initial guess of 500.}
\agentmsg{2}{[750]}
\gmmsg{The number is less than 750.}
\par\smallskip\noindent\textcolor{gray}{\scriptsize\itshape \dots\ continues (showing 2 of 10 turns)}\par
\end{taskbox} & \begin{taskbox}{Task 5}{GeminiColor}
\thinkmsg{1}{**Initiating Game Guess** {\\\relax}\  {\\\relax}\ My binary search begins anew! I'm targeting Game 5, and I'm starting from scratch with the entire range of numbers, 1 to 1000. My initial guess will be the midpoint, 500. Let's see where this takes us.}
\agentmsg{1}{[500]}
\gmmsg{The number is greater than 500.}
\agentmsg{2}{[750]}
\par\smallskip\noindent\textcolor{gray}{\scriptsize\itshape \dots\ continues (showing 2 of 10 turns)}\par
\end{taskbox} & \begin{taskbox}{Task 8}{GeminiColor}
\thinkmsg{1}{**Initiating Game 8 Analysis** {\\\relax}\  {\\\relax}\ Okay, I'm kicking things off with Game 8. My initial strategy? A binary search. To maximize efficiency right from the get-go, I'm setting my first guess at 500, the midpoint of the 1-1000 range. That's the optimal starting point for narrowing down the possibilities.}
\agentmsg{1}{[500]}
\gmmsg{The number is greater than 500.}
\agentmsg{2}{[750]}
\gmmsg{The number is less than 750.}
\par\smallskip\noindent\textcolor{gray}{\scriptsize\itshape \dots\ continues (showing 2 of 10 turns)}\par
\end{taskbox}
\\
\end{tabularx}
\end{anecdotebox}

\newpage

\begin{anecdotebox}{Claude does not try to identify any pattern}{ClaudeColor}
\noindent\textbf{Category:} Failure to adapt across tasks\par
\noindent\textbf{Model:} Claude Sonnet 4.6\par
\noindent\textbf{Configuration:} Environment: number\_guessing. Latent: set\_of\_3. Prompt: no\_info. Feedback: standard. Trajectory: 0.\par\smallskip
\textbf{Setup.} Same three-element target set on every task. Claude opens tasks 1, 5 and 9 with the same midpoint guess and the same binary-search progression.\par\smallskip
\textbf{What went wrong.} Previously revealed targets are not consulted as priors.\par\smallskip
\textbf{Per-task summary} (\textit{$\star$ marks tasks whose transcript is shown below}).\par
\noindent Agent task is to identify a hidden integer between 1 and 1000; after each guess, the game master replies whether the target is greater than, less than, or equal to the guess.\par
\par\noindent\textit{Hidden latent: set = \{592, 781, 926\}}\par\smallskip
\noindent\footnotesize
\begin{tabularx}{\linewidth}{@{}>{\bfseries}p{1.6cm}>{\raggedright\arraybackslash}X>{\raggedright\arraybackslash}X>{\raggedright\arraybackslash}X>{\raggedright\arraybackslash}X>{\raggedright\arraybackslash}X>{\raggedright\arraybackslash}X>{\raggedright\arraybackslash}X>{\raggedright\arraybackslash}X>{\raggedright\arraybackslash}X>{\raggedright\arraybackslash}X@{}}
\toprule
Task & \textbf{$\star$1} & \textbf{2} & \textbf{3} & \textbf{$\star$4} & \textbf{5} & \textbf{6} & \textbf{7} & \textbf{8} & \textbf{$\star$9} & \textbf{10} \\
\midrule
Answer & 781 & 592 & 926 & 592 & 926 & 592 & 926 & 926 & 592 & 781 \\
Turns & 5 & 9 & 8 & 10 & 8 & 10 & 8 & 8 & 10 & 5 \\
Reward & 0.90 & 0.82 & 0.84 & 0.80 & 0.84 & 0.80 & 0.84 & 0.84 & 0.80 & 0.90 \\
\bottomrule
\end{tabularx}
\normalsize
\par\smallskip
\par\smallskip\noindent\textit{Showing Task 1, Task 4 and Task 9 out of 10 tasks.}\par\smallskip
\noindent\begin{tabularx}{\linewidth}{@{}X@{\hspace{4pt}}X@{\hspace{4pt}}X@{}}
\begin{taskbox}{Task 1}{ClaudeColor}
\agentmsg{1}{I'll use a binary search strategy to find the number efficiently. {\\\relax}\  {\\\relax}\ [500]}
\gmmsg{The number is greater than 500.}
\thinkmsg{2}{Binary search: next guess is 750.}
\agentmsg{2}{[750]}
\gmmsg{The number is greater than 750.}
\par\smallskip\noindent\textcolor{gray}{\scriptsize\itshape \dots\ continues (showing 2 of 5 turns)}\par
\end{taskbox} & \begin{taskbox}{Task 4}{ClaudeColor}
\thinkmsg{1}{Let me start with binary search again.}
\agentmsg{1}{[500]}
\gmmsg{The number is greater than 500.}
\thinkmsg{2}{[750]}
\agentmsg{2}{[750]}
\gmmsg{The number is less than 750.}
\par\smallskip\noindent\textcolor{gray}{\scriptsize\itshape \dots\ continues (showing 2 of 10 turns)}\par
\end{taskbox} & \begin{taskbox}{Task 9}{ClaudeColor}
\thinkmsg{1}{Let me start with binary search again.}
\agentmsg{1}{[500]}
\gmmsg{The number is greater than 500.}
\thinkmsg{2}{[750]}
\agentmsg{2}{[750]}
\gmmsg{The number is less than 750.}
\par\smallskip\noindent\textcolor{gray}{\scriptsize\itshape \dots\ continues (showing 2 of 10 turns)}\par
\end{taskbox}
\\
\end{tabularx}
\end{anecdotebox}

\clearpage\subsection{Adaptation Breakdown}
\subsubsection{Adaptation Breakdown: Models ignore the hint of hidden pattern}
\label{app:adaptation_breakdown_models_ignore}

\begin{anecdotebox}{GPT-4o ignores the  hint that there is a pattern}{GPT4oColor}
\noindent\textbf{Configuration:} Environment: number\_guessing. Latent: set\_of\_3. Prompt: some\_info. Feedback: standard. Trajectory: 0.\par\smallskip
\textbf{Setup.} On set\_of\_3 the targets cycle through three fixed values. With \texttt{some\_info}, GPT-4o is told a pattern exists. Yet the opening guess is the midpoint of the search range and the agent never tries one of the previously-seen targets first.\par\smallskip
\textbf{What went wrong.} The set-of-3 hypothesis is never voiced or tested. Behavior is identical to the \texttt{no\_info} regime.\par\smallskip
\textbf{Per-task summary} (\textit{$\star$ marks tasks whose transcript is shown below}).\par
\noindent Agent task is to identify a hidden integer between 1 and 1000; after each guess, the game master replies whether the target is greater than, less than, or equal to the guess.\par
\par\noindent\textit{Hidden latent: set = \{592, 781, 926\}}\par\smallskip
\noindent\footnotesize
\begin{tabularx}{\linewidth}{@{}>{\bfseries}p{1.6cm}>{\raggedright\arraybackslash}X>{\raggedright\arraybackslash}X>{\raggedright\arraybackslash}X>{\raggedright\arraybackslash}X>{\raggedright\arraybackslash}X>{\raggedright\arraybackslash}X>{\raggedright\arraybackslash}X>{\raggedright\arraybackslash}X>{\raggedright\arraybackslash}X>{\raggedright\arraybackslash}X@{}}
\toprule
Task & \textbf{1} & \textbf{2} & \textbf{3} & \textbf{4} & \textbf{5} & \textbf{6} & \textbf{7} & \textbf{8} & \textbf{$\star$9} & \textbf{$\star$10} \\
\midrule
Answer & 781 & 592 & 926 & 592 & 926 & 592 & 926 & 926 & 592 & 781 \\
Turns & 5 & 10 & 8 & 10 & 8 & 10 & 8 & 8 & 10 & 5 \\
Reward & 0.90 & 0.80 & 0.84 & 0.80 & 0.84 & 0.80 & 0.84 & 0.84 & 0.80 & 0.90 \\
\bottomrule
\end{tabularx}
\normalsize
\par\smallskip
\par\smallskip\noindent\textit{Showing Task 9 and Task 10 out of 10 tasks.}\par\smallskip
\setlength{\columnsep}{6pt}
\begin{multicols}{2}\raggedcolumns
\begin{taskbox}{Task 9}{GPT4oColor}
\agentmsg{1}{Great job so far! Let's begin Game 9 using the same effective strategy. {\\\relax}\  {\\\relax}\ My first guess is: [500]}
\gmmsg{The number is greater than 500.}
\agentmsg{2}{Alright, the number is greater than 500. Let's try the middle of the new range. {\\\relax}\  {\\\relax}\ My next guess is: [750]}
\gmmsg{The number is less than 750.}
\agentmsg{3}{Great! The number is between 501 and 749. {\\\relax}\  {\\\relax}\ My next guess is: [625]}
\gmmsg{The number is less than 625.}
\par\smallskip\noindent\textcolor{gray}{\scriptsize\itshape \dots\ continues (showing turns 1--7 of 10)}\par
\end{taskbox}
\columnbreak
\begin{taskbox}{Task 10}{GPT4oColor}
\agentmsg{1}{Great job so far! Let's start Game 10 with the same effective strategy. {\\\relax}\  {\\\relax}\ My first guess is: [500]}
\gmmsg{The number is greater than 500.}
\agentmsg{2}{Alright, the number is greater than 500. Let's try the middle of the new range. {\\\relax}\  {\\\relax}\ My next guess is: [750]}
\gmmsg{The number is greater than 750.}
\agentmsg{3}{Got it! The number is greater than 750. Let's narrow it down further. {\\\relax}\  {\\\relax}\ My next guess is: [875]}
\gmmsg{The number is less than 875.}
\par\smallskip\noindent\textcolor{gray}{\scriptsize\itshape \dots\ continues (showing turns 1--3 of 5)}\par
\end{taskbox}
\end{multicols}
\end{anecdotebox}

\newpage

\subsubsection{Adaptation Breakdown: Models fail to act upon the recognized pattern}
\label{app:adaptation_breakdown_models_fail_to_act}
\begin{anecdotebox}{Gemini recognizes the pattern but does not act upon it}{GeminiColor}
\noindent\textbf{Category:} Vague hints under-used\par
\noindent\textbf{Model:} Gemini 2.5 Flash\par
\noindent\textbf{Configuration:} Environment: number\_guessing. Latent: dynamic\_range. Prompt: some\_info. Feedback: standard. Trajectory: 1.\par\smallskip
\textbf{Setup.} The \texttt{some\_info} hint is meant to nudge the model to look for cross-task patterns. Gemini's behavior across the trajectory does not reflect any such search.\par\smallskip
\textbf{What went wrong.} No reference to the hint, no statistics over previous targets, and no narrowing of the search range across the ten tasks.\par\smallskip
\textbf{Per-task summary} (\textit{$\star$ marks tasks whose transcript is shown below}).\par
\noindent Agent task is to identify a hidden integer between 365 and 10365; after each guess, the game master replies whether the target is greater than, less than, or equal to the guess.\par
\par\noindent\textit{Hidden latent: range covers approx. [1461, 2278]}\par\smallskip
\noindent\footnotesize
\begin{tabularx}{\linewidth}{@{}>{\bfseries}p{1.6cm}>{\raggedright\arraybackslash}X>{\raggedright\arraybackslash}X>{\raggedright\arraybackslash}X>{\raggedright\arraybackslash}X>{\raggedright\arraybackslash}X>{\raggedright\arraybackslash}X>{\raggedright\arraybackslash}X>{\raggedright\arraybackslash}X>{\raggedright\arraybackslash}X>{\raggedright\arraybackslash}X@{}}
\toprule
Task & \textbf{1} & \textbf{2} & \textbf{3} & \textbf{4} & \textbf{5} & \textbf{6} & \textbf{$\star$7} & \textbf{8} & \textbf{9} & \textbf{$\star$10} \\
\midrule
Answer & 2233 & 2278 & 1511 & 1820 & 1983 & 1811 & 2048 & 1461 & 1958 & 1663 \\
Turns & 13 & 8 & 13 & 12 & 14 & 12 & 13 & 11 & 13 & 13 \\
Reward & 0.74 & 0.84 & 0.74 & 0.76 & 0.72 & 0.76 & 0.74 & 0.78 & 0.74 & 0.74 \\
\bottomrule
\end{tabularx}
\normalsize
\par\smallskip
\par\smallskip\noindent\textit{Showing Task 7 and Task 10 out of 10 tasks.}\par\smallskip
\setlength{\columnsep}{6pt}
\begin{multicols}{2}\raggedcolumns
\begin{taskbox}{Task 7}{GeminiColor}
\thinkmsg{1}{**Analyzing Initial Parameters** I've begun the process by defining the initial boundaries: 365 to 10365. My strategy remains binary search. I'm calculating the midpoint to make my first guess within the provided range, aiming for an efficient approach to finding the number. {\\\relax}\ **Initiating First Guess** Having crunched the numbers, my first guess is in. Based on the range and my binary search strategy, I'm going with 5365. Now, I eagerly await feedback to refine my search.}
\agentmsg{1}{[5365]}
\gmmsg{The number is less than 5365.}
\agentmsg{2}{Okay, the number is less than 5365. {\\\relax}\  {\\\relax}\ The range is now between 365 and 5364. You have 29 guesses remaining. What's your next guess? {\\\relax}\ [2864]}
\gmmsg{The number is less than 2864.}
\par\smallskip\noindent\textcolor{gray}{\scriptsize\itshape \dots\ continues (showing turns 1--2 of 13)}\par
\end{taskbox}
\columnbreak
\begin{taskbox}{Task 10}{GeminiColor}
\thinkmsg{1}{**Estimating Initial Value** {\\\relax}\  {\\\relax}\ I'm calculating the initial guess for the game's final round. I've determined that a binary search is the way to go, which is an efficient strategy. Starting with the given range, I need to find the midpoint. To do this, I must first calculate the range's size and then determine the midpoint using the provided formula. {\\\relax}\  {\\\relax}\  {\\\relax}\ **Analyzing Previous Results** {\\\relax}\  {\\\relax}\ I've reviewed the previous nine games. The numbers, ranging from 1461 to 2278, are well below the upper bound of 10365, although I'll continue using binary search. The midpoint calculation is still correct, and my guess will remain at 5365.}
\agentmsg{1}{[5365]}
\gmmsg{The number is less than 5365.}
\par\smallskip\noindent\textcolor{gray}{\scriptsize\itshape \dots\ continues (showing turns 1 of 13)}\par
\end{taskbox}
\end{multicols}
\end{anecdotebox}

\newpage

\begin{anecdotebox}{Claude recognizes the hint but does not act upon it strongly}{ClaudeColor}
\noindent\textbf{Model:} Claude Sonnet 4.6\par
\noindent\textbf{Configuration:} Environment: bandits. Latent: ping\_pong. Prompt: some\_info. Feedback: standard. Trajectory: 0.\par\smallskip
\textbf{Setup.} On ping\_pong with \texttt{some\_info}, Claude's openings do not engage with the alternation hypothesis either — its cumulative reward (6.5--7.7) is in the same range as under \texttt{no\_info}.\par\smallskip
\textbf{What went wrong.} Same uniform exploration each task; no reference to prior winners.\par\smallskip
\textbf{Per-task summary} (\textit{$\star$ marks tasks whose transcript is shown below}).\par
\noindent Agent task is to identify the button with the highest reward probability among 5 buttons (red, blue, green, yellow, purple); each turn the agent may press a button to observe a 0/1 reward, or commit via [select \textless{}button\textgreater{}].\par
\par\noindent\textit{Hidden latent: best button alternates: red (first) -- purple (last)}\par\smallskip
\noindent\footnotesize
\begin{tabularx}{\linewidth}{@{}>{\bfseries}p{1.6cm}>{\raggedright\arraybackslash}X>{\raggedright\arraybackslash}X>{\raggedright\arraybackslash}X>{\raggedright\arraybackslash}X>{\raggedright\arraybackslash}X>{\raggedright\arraybackslash}X>{\raggedright\arraybackslash}X>{\raggedright\arraybackslash}X>{\raggedright\arraybackslash}X>{\raggedright\arraybackslash}X@{}}
\toprule
Task & \textbf{1} & \textbf{2} & \textbf{3} & \textbf{4} & \textbf{5} & \textbf{6} & \textbf{$\star$7} & \textbf{8} & \textbf{9} & \textbf{$\star$10} \\
\midrule
Answer & red & purple & red & purple & red & purple & red & purple & red & purple \\
Turns & 9 & 10 & 10 & 15 & 20 & 11 & 19 & 8 & 11 & 11 \\
Reward & 0.86 & 0.85 & 0.85 & 0.00 & 0.70 & 0.83 & 0.00 & 0.88 & 0.83 & 0.83 \\
\bottomrule
\end{tabularx}
\normalsize
\par\smallskip
\par\smallskip\noindent\textit{Showing Task 7 and Task 10 out of 10 tasks.}\par\smallskip
\setlength{\columnsep}{6pt}
\begin{multicols}{2}\raggedcolumns
\begin{taskbox}{Task 7}{ClaudeColor}
\thinkmsg{1}{Let me review the pattern: {\\\relax}\ - Episode 1: Red won {\\\relax}\ - Episode 2: Purple won {\\\relax}\ - Episode 3: Red won {\\\relax}\ - Episode 4: ??? (I selected green, which was wrong - might have been red or something else) {\\\relax}\ - Episode 5: Red won {\\\relax}\ - Episode 6: Purple won {\\\relax}\ Pattern seems like Red and Purple alternate or dominate. Let me start by testing red and purple, then the others.}
\agentmsg{1}{ [red]}
\par\smallskip\noindent\textcolor{gray}{\scriptsize\itshape \dots\ after 17 more turns \dots}\par\smallskip
\thinkmsg{19}{Updated stats:  - Red: 1/3 (33\%) - Blue: 0/1 (0\%)  - Green: 2/6 (33\%)  - Yellow: 3/7 (43\%) - Purple: 0/1 (0\%)  Yellow is still leading at 43\%. I've used 18 turns now. Yellow is the clear leader compared to others.}
\agentmsg{19}{ [select yellow]}
\gmmsg{You selected 'yellow' as your final answer on turn 19. Wrong!}
\end{taskbox}
\columnbreak
\begin{taskbox}{Task 10}{ClaudeColor}
\thinkmsg{1}{Final episode! Pattern summary: {\\\relax}\ - Episode 1: Red \checkmark{} {\\\relax}\ - Episode 2: Purple \checkmark{} {\\\relax}\ - Episode 3: Red \checkmark{} {\\\relax}\ - Episode 4: Unknown (wrong) {\\\relax}\ - Episode 5: Red \checkmark{} {\\\relax}\ - Episode 6: Purple \checkmark{} {\\\relax}\ - Episode 7: Unknown (wrong) {\\\relax}\ - Episode 8: Purple \checkmark{} {\\\relax}\ - Episode 9: Red \checkmark{}   {\\\relax}\ Red and purple dominate. Let me test both quickly and focus on getting a clear winner.}
\agentmsg{1}{ [red]}
\par\smallskip\noindent\textcolor{gray}{\scriptsize\itshape \dots\ after 9 more turns \dots}\par\smallskip
\agentmsg{11}{Updated stats:  - Red: 1/2 (50\%)  - Blue: 1/2 (50\%)  - Green: 0/1 (0\%)  - Yellow: 0/1 (0\%)  - Purple: 3/4 (75\%).  Purple is clearly the leader at 75\%! Selecting now.  [select purple]}
\gmmsg{You selected 'purple' as your final answer on turn 11. Correct!}
\end{taskbox}
\end{multicols}
\end{anecdotebox}

\newpage
\subsubsection{Adaptation Breakdown: Models fail to identify the pattern}
\label{app:adaptation_breakdown_models_fail_identify_pattern}

\begin{anecdotebox}{Gemini tries hard but is not able to find the simple hidden pattern - biases towards that red is the favorite}{GeminiColor}
\noindent\textbf{Configuration:} Environment: bandits. Latent: ping\_pong. Prompt: some\_info. Feedback: standard. Trajectory: 4.\par\smallskip
\textbf{Setup.} Second Gemini trajectory on bandits/ping\_pong with \texttt{some\_info}. Cumulative reward 3.78 — half of GPT-4o on the same configuration. Several tasks score 0 due to commits on the wrong-side arm.\par\smallskip
\textbf{What went wrong.} The agent commits aggressively on one arm and accepts the alternation tax.\par\smallskip
\textbf{Per-task summary} (\textit{$\star$ marks tasks whose transcript is shown below}).\par
\noindent Agent task is to identify the button with the highest reward probability among 5 buttons (red, blue, green, yellow, purple); each turn the agent may press a button to observe a 0/1 reward, or commit via [select \textless{}button\textgreater{}].\par
\par\noindent\textit{Hidden latent: best button alternates: red (first) -- purple (last)}\par\smallskip
\noindent\footnotesize
\begin{tabularx}{\linewidth}{@{}>{\bfseries}p{1.6cm}>{\raggedright\arraybackslash}X>{\raggedright\arraybackslash}X>{\raggedright\arraybackslash}X>{\raggedright\arraybackslash}X>{\raggedright\arraybackslash}X>{\raggedright\arraybackslash}X>{\raggedright\arraybackslash}X>{\raggedright\arraybackslash}X>{\raggedright\arraybackslash}X>{\raggedright\arraybackslash}X@{}}
\toprule
Task & \textbf{1} & \textbf{2} & \textbf{3} & \textbf{4} & \textbf{$\star$5} & \textbf{6} & \textbf{7} & \textbf{8} & \textbf{9} & \textbf{$\star$10} \\
\midrule
Answer & red & purple & red & purple & red & purple & red & purple & red & purple \\
Turns & 26 & 28 & 26 & 24 & 24 & 24 & 24 & 24 & 24 & 24 \\
Reward & 0.61 & 0.00 & 0.61 & 0.64 & 0.00 & 0.64 & 0.64 & 0.64 & 0.00 & 0.00 \\
\bottomrule
\end{tabularx}
\normalsize
\setlength{\columnsep}{6pt}
\begin{multicols}{2}\raggedcolumns
\begin{taskbox}{Task 5}{GeminiColor}
\thinkmsg{1}{**Considering Game Strategies**  I'm analyzing the data from Game 1, specifically the success rates for each color. Red seems promising at 61.5\%, while blue is at 50\%. The other colors, unfortunately, show no success. This initial data shapes my strategy for Game 5, where I'll weigh the odds based on these early insights. {\\\relax}\   **Refining Color Probabilities**  I am now focusing on the probabilities, updating them with data from Game 2 and Game 3. Red still seems decent, but its success rate dropped in Game 2. Purple showed up in Game 3 with a 36\% success rate, which is interesting. I'll need to consider this in light of the larger picture. I want to calculate an overall ranking for all colors before Game 5, which seems imminent.}
\agentmsg{1}{[red]}
\gmmsg{You pressed the red button and received a reward of 0.0.}
\agentmsg{2}{[blue]}
\par\smallskip\noindent\textcolor{gray}{\scriptsize\itshape \dots\ continues (showing turns 1--2 of 24)}\par
\end{taskbox}
\columnbreak
\begin{taskbox}{Task 10}{GeminiColor}
\thinkmsg{1}{**Analyzing Game Outcomes** Okay, I'm starting to analyze the results from the first game of ten. So far, the initial pull was heavily red, with subsequent pulls including both red and blue. I'm noting the color distribution and sequence to identify any emerging patterns within this one game, and then I'll look at the bigger picture! {\\\relax}\   **Reviewing Game Results**  I've just finished reviewing the results of the second game. The initial pulls showed red and green, and the subsequent results saw them alternating, but red's performance was not as strong this time. However, green was even weaker. I'm focusing on these fluctuations in performance to look for patterns across all games. I'll need to examine all the games and their scores to identify any significant performance trends.}
\agentmsg{1}{[red]}
\gmmsg{You pressed the red button and received a reward of 0.0.}
\agentmsg{2}{[blue]}
\par\smallskip\noindent\textcolor{gray}{\scriptsize\itshape \dots\ continues (showing turns 1--2 of 24)}\par
\end{taskbox}
\end{multicols}
\end{anecdotebox}

\newpage
\begin{anecdotebox}{Claude tries hard but fails to figure out the pattern}{ClaudeColor}
\noindent\textbf{Configuration:} Environment: secretary. Latent: threshold\_06. Prompt: some\_info. Feedback: standard. Trajectory: 1.\par\smallskip
\textbf{What went wrong.} Claude fails to recognize the parttern. \par\smallskip
\textbf{Per-task summary} (\textit{$\star$ marks tasks whose transcript is shown below}).\par
\noindent Agent task is to pick the maximum of 10 values revealed one at a time; on each turn the agent chooses [accept] (claim the current value and end the task) or [continue] (skip).\par
\par\noindent\textit{Hidden latent: max value \textgreater{} 0.6; all others \textless{} 0.6}\par\smallskip
\noindent\footnotesize
\begin{tabularx}{\linewidth}{@{}>{\bfseries}p{1.6cm}>{\raggedright\arraybackslash}X>{\raggedright\arraybackslash}X>{\raggedright\arraybackslash}X>{\raggedright\arraybackslash}X>{\raggedright\arraybackslash}X>{\raggedright\arraybackslash}X>{\raggedright\arraybackslash}X>{\raggedright\arraybackslash}X>{\raggedright\arraybackslash}X>{\raggedright\arraybackslash}X@{}}
\toprule
Task & \textbf{1} & \textbf{2} & \textbf{3} & \textbf{4} & \textbf{5} & \textbf{6} & \textbf{7} & \textbf{8} & \textbf{$\star$9} & \textbf{$\star$10} \\
\midrule
Answer & 0.69 at 6 & 0.72 at 3 & 0.69 at 9 & 0.63 at 8 & 0.73 at 8 & 0.72 at 10 & 0.62 at 6 & 0.63 at 7 & 0.70 at 1 & 0.73 at 9 \\
Turns & 9 & 11 & 8 & 9 & 9 & 7 & 6 & 6 & 11 & 7 \\
Reward & 1.00 & 0.33 & 0.38 & 1.00 & 1.00 & 0.31 & 1.00 & 0.42 & 0.13 & 0.37 \\
\bottomrule
\end{tabularx}
\normalsize
\setlength{\columnsep}{6pt}
\begin{multicols}{2}\raggedcolumns
\begin{taskbox}{Task 9}{ClaudeColor}
\thinkmsg{1}{Let me think about the pattern of where the maximum has appeared in previous rounds: {\\\relax}\  {\\\relax}\ Episode 1: Max was at position 6 (0.6937) - accepted correctly {\\\relax}\ Episode 2: Max was at position 3 (0.7158) - missed it, ended with 0.4755 {\\\relax}\ Episode 3: Max was somewhere after position 7 (0.5239 wasn't the max) {\\\relax}\ Episode 4: Max was at position 8 (0.6250) - accepted correctly {\\\relax}\ Episode 5: Max was at position 8 (0.7349) - accepted correctly {\\\relax}\ Episode 6: Max was somewhere after position 6 (0.4481 wasn't the max) {\\\relax}\ Episode 7: Max was at position 6 (0.6178) - accepted correctly {\\\relax}\ Episode 8: Max was somewhere after position 6 (0.5276 wasn't the max) {\\\relax}\   Now I'm at position 1 with a value of 0.7004, which is quite strong compared to the historical maximums I've seen. Following the secretary problem's optimal stopping rule would suggest skipping the first \textasciitilde{}37\% of candidates, but this current value is already in the upper range of what I've encountered before, so I need to weigh whether to accept it now or continue searching.}
\agentmsg{1}{[continue]}
\par\smallskip\noindent\textcolor{gray}{\scriptsize\itshape \dots\ continues (showing turn 1 of 11)}\par
\end{taskbox}
\columnbreak
\begin{taskbox}{Task 10}{ClaudeColor}
\thinkmsg{1}{Let me think about the patterns I've observed across the games: {\\\relax}\  {\\\relax}\ Game 1: Max was at position 6 (0.6937) - I accepted it \checkmark{} {\\\relax}\ Game 2: Max was at position 3 (0.7158) - I missed it, ended up accepting position 10 $\times$ {\\\relax}\ Game 3: Max was somewhere after position 7 - I accepted 0.5239 at position 7 $\times$ {\\\relax}\ Game 4: Max was at position 8 (0.6250) - I accepted it \checkmark{} {\\\relax}\ Game 5: Max was at position 8 (0.7349) - I accepted it \checkmark{}   {\\\relax}\ Game 6: Max was somewhere after position 6 - I accepted 0.4481 at position 6 $\times$ {\\\relax}\ Game 7: Max was at position 6 (0.6178) - I accepted it \checkmark{} {\\\relax}\ Game 8: Max was somewhere after position 6 - I accepted 0.5276 at position 6 $\times$ {\\\relax}\ Game 9: Max was at position 1 (0.7004) - I missed it $\times$ {\\\relax}\  Looking at where the maximum actually appeared---6, 3, unknown, 8, 8, unknown, 6, unknown, 1---I'm noticing the positions vary quite a bit, and I've been stopping too early in several cases. This is the final game, so I need to be more strategic about when to accept versus continue searching. {\\\relax}\   My approach is to skip the first few positions and then take the first value that exceeds everything I've seen so far. Right now at position 1 with 0.5315, I'm still in the observation phase, so I'll keep looking.}
\agentmsg{1}{[continue]}
\par\smallskip\noindent\textcolor{gray}{\scriptsize\itshape \dots\ continues (showing turns 1 of 7)}\par
\end{taskbox}
\end{multicols}
\end{anecdotebox}

\newpage

\subsubsection{Adaptation Breakdown: Model tries too hard to fit a pattern leading to suboptimal behavior}
\label{app:adaptation_breakdown_model_tries_hard}

\begin{anecdotebox}{Gemini trying hard to find the pattern - finds that it was wrong - abondons the binary search on the given task. And once failed on a task doent even attempt to find the pattern.}{GeminiColor}
\noindent\textbf{Configuration:} Environment: number\_guessing. Latent: two\_ranges. Prompt: some\_info. Feedback: standard. Trajectory: 0.\par\smallskip
\textbf{Setup.} On task 5 of two\_ranges, Gemini begins with binary search (3750, 4375, ...) but after crossing the midpoint switches to a `+1 each turn' hypothesis and increments linearly, exhausting all 30 turns without finding 5900. Episode reward: 0; trajectory cumulative: 6.74.\par\smallskip
\textbf{What went wrong.} The model commits to a wrong incrementing strategy after binary search — once committed, it does not revisit the binary-search hypothesis and wastes turns.\par\smallskip
\textbf{Per-task summary} (\textit{$\star$ marks tasks whose transcript is shown below}).\par
\noindent Agent task is to identify a hidden integer between 1 and 10000; after each guess, the game master replies whether the target is greater than, less than, or equal to the guess.\par
\par\noindent\textit{Hidden latent: targets drawn from two disjoint ranges}\par\smallskip
\noindent\footnotesize
\begin{tabularx}{\linewidth}{@{}>{\bfseries}p{1.6cm}>{\raggedright\arraybackslash}X>{\raggedright\arraybackslash}X>{\raggedright\arraybackslash}X>{\raggedright\arraybackslash}X>{\raggedright\arraybackslash}X>{\raggedright\arraybackslash}X>{\raggedright\arraybackslash}X>{\raggedright\arraybackslash}X>{\raggedright\arraybackslash}X>{\raggedright\arraybackslash}X@{}}
\toprule
Task & \textbf{1} & \textbf{2} & \textbf{3} & \textbf{4} & \textbf{$\star$5} & \textbf{6} & \textbf{7} & \textbf{8} & \textbf{9} & \textbf{$\star$10} \\
\midrule
Answer & 2721 & 2729 & 2702 & 2501 & 5900 & 2715 & 6233 & 2805 & 5821 & 2844 \\
Turns & 12 & 11 & 13 & 12 & 30 & 13 & 13 & 13 & 13 & 13 \\
Reward & 0.76 & 0.78 & 0.74 & 0.76 & 0.00 & 0.74 & 0.74 & 0.74 & 0.74 & 0.74 \\
\bottomrule
\end{tabularx}
\normalsize
\end{anecdotebox}
\begin{anecdotebox}{Transcript}{GeminiColor}
\setlength{\columnsep}{6pt}
\begin{multicols}{2}\raggedcolumns
\begin{taskbox}{Task 5}{GeminiColor}
\thinkmsg{1}{\textit{(strategy at start of episode)}\space {\\\relax}\ **Analyzing Game Patterns**  I've just finished analyzing the first four game rounds. The initial number guesses were 2721 (12 turns), 2729 (11 turns), 2702 (13 turns), and 2501 (12 turns). All guesses have fallen between 2501 and 2729. I am beginning to look at if there's any pattern there. {\\\relax}\ {\\\relax}\   **Refining Initial Guesses**  I've got a clearer picture of the number range: [2501, 2729]. The average is around 2663.25. While binary search is generally optimal, I'm now weighing the risk/reward of deviating. Should I aim for the middle of this observed sub-range (maybe 2600-2700) to speed up if the pattern continues, or stick with a binary search-type approach given the "might" in "might follow a pattern" condition? Considering max turns are 30, and the initial range of 10000. {\\\relax}\ {\\\relax}\ **Re-evaluating Optimal Strategy** I'm now zeroing in on an optimal initial guess, based on the accumulated data.\ldots{}}
\noindent\textcolor{gray}{\scriptsize\itshape \dots\ first 8 turns of this task omitted}\par\smallskip
\agentmsg{9}{[4995]}
\gmmsg{The number is greater than 4995.}
\agentmsg{10}{[4998]}
\gmmsg{The number is greater than 4998.}
\agentmsg{11}{[4999]}
\gmmsg{The number is greater than 4999.}
\agentmsg{12}{[5000]}
\gmmsg{The number is greater than 5000.}
\agentmsg{13}{[5001]}
\gmmsg{The number is greater than 5001.}
\agentmsg{14}{[5002]}
\par\smallskip\noindent\textcolor{gray}{\scriptsize\itshape \dots\ continues (showing turns 9--14 of 30)}\par
\end{taskbox}
\columnbreak
\begin{taskbox}{Task 10}{GeminiColor}
\thinkmsg{1}{**Analyzing Game Trends** {\\\relax}\  {\\\relax}\ I'm now starting Game 10. Based on the data from the previous nine games, I'm focusing on the number of turns and final scores. The scores are consistently low, under 5000. Each game is between 11-13 turns. I'll make my initial guess soon, keeping these trends in mind. {\\\relax}\  {\\\relax}\  {\\\relax}\ **Evaluating Recent Results** {\\\relax}\  {\\\relax}\ I've just finished reviewing the outcomes from Games 1-9. I'm seeing a trend: most scores cluster around 2500-3000, but there are also a couple of higher scores, in the\ldots{}}
\agentmsg{1}{I'm thinking of a number between 1 and 10000. You have 30 guesses. What's your guess? {\\\relax}\ [5000]}
\gmmsg{The number is less than 5000.}
\agentmsg{2}{[2500]}
\gmmsg{The number is greater than 2500.}
\agentmsg{3}{[3750]}
\gmmsg{The number is less than 3750.}
\agentmsg{4}{[3125]}
\gmmsg{The number is less than 3125.}
\agentmsg{5}{[2812]}
\par\smallskip\noindent\textcolor{gray}{\scriptsize\itshape \dots\ continues (showing turns 1--7 of 13)}\par
\end{taskbox}
\end{multicols}
\end{anecdotebox}

\newpage

\begin{anecdotebox}{Claude wrongly reasons about a failed task and tries too hard to reason for the pattern}{ClaudeColor}
\noindent\textbf{Model:} Claude Sonnet 4.6\par
\noindent\textbf{Configuration:} Environment: bandits. Latent: ping\_pong. Prompt: some\_info. Feedback: standard. Trajectory: 4.\par\smallskip
\textbf{Setup.} On ping\_pong with \texttt{some\_info}, Claude's trajectory~4 cumulative reward is 4.76 — well below its best ping\_pong trajectory (7.71). Several tasks score 0 because the agent commits to one arm and the latent alternates the answer.\par\smallskip
\textbf{What went wrong.} After a couple of wins on one arm, Claude commits and stops rotating. The alternation latent then punishes the very next task. Reasoning is verbose; the policy is brittle.\par\smallskip
\textbf{Per-task summary} (\textit{$\star$ marks tasks whose transcript is shown below}).\par
\noindent Agent task is to identify the button with the highest reward probability among 5 buttons (red, blue, green, yellow, purple); each turn the agent may press a button to observe a 0/1 reward, or commit via [select \textless{}button\textgreater{}].\par
\par\noindent\textit{Hidden latent: best button alternates: red (first) -- purple (last)}\par\smallskip
\noindent\footnotesize
\begin{tabularx}{\linewidth}{@{}>{\bfseries}p{1.6cm}>{\raggedright\arraybackslash}X>{\raggedright\arraybackslash}X>{\raggedright\arraybackslash}X>{\raggedright\arraybackslash}X>{\raggedright\arraybackslash}X>{\raggedright\arraybackslash}X>{\raggedright\arraybackslash}X>{\raggedright\arraybackslash}X>{\raggedright\arraybackslash}X>{\raggedright\arraybackslash}X@{}}
\toprule
Task & \textbf{1} & \textbf{2} & \textbf{3} & \textbf{4} & \textbf{5} & \textbf{6} & \textbf{$\star$7} & \textbf{8} & \textbf{9} & \textbf{$\star$10} \\
\midrule
Answer & red & purple & red & purple & red & purple & red & purple & red & purple \\
Turns & 8 & 11 & 9 & 15 & 18 & 15 & 14 & 16 & 12 & 18 \\
Reward & 0.88 & 0.83 & 0.00 & 0.78 & 0.73 & 0.78 & 0.00 & 0.76 & 0.00 & 0.00 \\
\bottomrule
\end{tabularx}
\normalsize
\par\smallskip
\par\smallskip\noindent\textit{Showing Task 7 and Task 10 out of 10 tasks.}\par\smallskip
\setlength{\columnsep}{6pt}
\begin{multicols}{2}\raggedcolumns
\begin{taskbox}{Task 7}{ClaudeColor}
\thinkmsg{1}{Pattern so far: Game 1=Red, Game 2=Purple, Game 3=Green, Game 4=Purple, Game 5=Red, Game 6=Purple. {\\\relax}\  Purple has won 3 times, Red 2 times, Green 1 time. Blue and Yellow haven't won yet. {\\\relax}\  Let me explore all buttons systematically again.}
\agentmsg{1}{  [red]}
\gmmsg{You pressed the red button and received a reward of 0.0.}
\thinkmsg{2}{Let me explore all buttons to get initial data.}
\agentmsg{2}{[blue]}
\par\smallskip\noindent\textcolor{gray}{\scriptsize\itshape \dots\ continues (showing turns 1--2 of 14)}\par
\end{taskbox}
\columnbreak
\begin{taskbox}{Task 10}{ClaudeColor}
\thinkmsg{1}{This is very frustrating. I've been getting unlucky with buttons that appear to have high probabilities but aren't the actual highest. The issue seems to be that multiple buttons can have decent probabilities, and I need more samples to differentiate.  Let me think about what I know: - Game 1: Red won - Game 2: Purple won  - Game 3: Green won (I picked yellow with 3/3 and was wrong)  - Game 4: Purple won  - Game 5: Red won - Game 6: Purple won - Game 7: Unknown winner (I picked yellow with 2/5,\ldots{}}
\agentmsg{1}{This is the last game. I've noticed that even buttons with 4/4 results can be wrong - there might be a button with an even higher probability. I need to be very thorough this time. Let me test each button multiple times. [red]}
\gmmsg{You pressed the red button and received a reward of 0.0.}
\par\smallskip\noindent\textcolor{gray}{\scriptsize\itshape \dots\ continues (showing turns 1 of 18)}\par
\end{taskbox}
\end{multicols}
\end{anecdotebox}

\clearpage
\subsection{Adaptation Miscalibration}
\label{app:failure-anecdotes-miscalibrated-adaptation}

\begin{anecdotebox}{GPT 4o does not know how to use the latent information}{GPT4oColor}
\noindent\textbf{Category:} Full\_info confusion\par
\noindent\textbf{Model:} GPT-4o\par
\noindent\textbf{Configuration:} Environment: number\_guessing. Latent: two\_ranges. Prompt: full\_info. Feedback: standard. Trajectory: 0.\par\smallskip
\textbf{Setup.} The \texttt{full\_info} prompt describes the disjoint-ranges structure verbosely. GPT-4o under \texttt{full\_info} scores below its own \texttt{some\_info} performance — tasks 8 and 9 both end at zero reward despite GPT-4o correctly hypothesising both candidate ranges at turn 1.\par\smallskip
\textbf{What went wrong.} GPT-4o opens both episodes with the right two-range hypothesis ($\sim$2501--3000 and $\sim$5900--6399), but after the first feedback it commits to a single range and exhausts the 30-turn budget there. By the time it crosses to the other range the budget is already gone. Task 8 (target 2805, lower range) is searched in the upper range until turn 27; Task 9 (target 5821, upper range) is searched in the lower range until turn 26.\par\smallskip
\textbf{Per-task summary} (\textit{$\star$ marks tasks whose transcript is shown below}).\par
\noindent Agent task is to identify a hidden integer between 1 and 10000; after each guess, the game master replies whether the target is greater than, less than, or equal to the guess.\par
\par\noindent\textit{Hidden latent: targets drawn from two disjoint ranges}\par\smallskip
\noindent\footnotesize
\begin{tabularx}{\linewidth}{@{}>{\bfseries}p{1.6cm}>{\raggedright\arraybackslash}X>{\raggedright\arraybackslash}X>{\raggedright\arraybackslash}X>{\raggedright\arraybackslash}X>{\raggedright\arraybackslash}X>{\raggedright\arraybackslash}X>{\raggedright\arraybackslash}X>{\raggedright\arraybackslash}X>{\raggedright\arraybackslash}X>{\raggedright\arraybackslash}X@{}}
\toprule
Task & \textbf{1} & \textbf{2} & \textbf{3} & \textbf{4} & \textbf{5} & \textbf{6} & \textbf{7} & \textbf{$\star$8} & \textbf{$\star$9} & \textbf{10} \\
\midrule
Answer & 2721 & 2729 & 2702 & 2501 & 5900 & 2715 & 6233 & 2805 & 5821 & 2844 \\
Turns & 12 & 12 & 15 & 10 & 11 & 9 & 9 & 30 & 30 & 5 \\
Reward & 0.76 & 0.76 & 0.70 & 0.80 & 0.78 & 0.82 & 0.82 & 0.00 & 0.00 & 0.90 \\
\bottomrule
\end{tabularx}
\normalsize
\setlength{\columnsep}{6pt}
\begin{multicols}{2}\raggedcolumns
\begin{taskbox}{Task 8}{GPT4oColor}
\agentmsg{1}{Based on previous games, we've identified two potential ranges: 2501 to 3000 and 5900 to 6399. Let's start by guessing the middle of these ranges to quickly determine which range the number might fall into: [2750]}
\gmmsg{The number is greater than 2750.}
\par\smallskip\noindent\textcolor{gray}{\scriptsize\itshape \dots\ after 28 more turns \dots}\par\smallskip
\par\noindent\textcolor{gray}{\scriptsize Intermediate guesses (turns 2--29): \texttt{6150, 6025, 5962, 5930, 5915, 5907, 5903, 5901, 5900, 5899, 5895, 5880, 5850, 5800, 5700, 5600, 5500, 5400, 5300, 5200, 5100, 5000, 4500, 4250, 4000, 3500, 3250, 3000}}\par\smallskip
\agentmsg{30}{The number is less than 3000. Let's narrow it down further within the known range: [2750]}
\gmmsg{The number is greater than 2750. {\\\relax}\ You've run out of guesses.}
\end{taskbox}
\columnbreak
\begin{taskbox}{Task 9}{GPT4oColor}
\agentmsg{1}{Based on previous games, we've identified two potential ranges: 2501 to 3000 and 5900 to 6399. Let's start with the middle of these ranges to quickly determine which range the number might fall into:  [2750]}
\gmmsg{The number is greater than 2750.}
\par\smallskip\noindent\textcolor{gray}{\scriptsize\itshape \dots\ after 28 more turns \dots}\par\smallskip
\par\noindent\textcolor{gray}{\scriptsize Intermediate guesses (turns 2--28): \texttt{2875, 2938, 2970, 2985, 2993, 2997, 2999, 3000, 6200, 2750, 2875, 2938, 2970, 2985, 2993, 2997, 2999, 3000, 3100, 3500, 4000, 5000, 5500, 6000, 5750, 5875, 5812}}\par\smallskip
\agentmsg{30}{The number is between 5813 and 5874. Let's narrow it down further: [5843]}
\gmmsg{The number is less than 5843. {\\\relax}\ You've run out of guesses.}
\end{taskbox}
\end{multicols}
\end{anecdotebox}

\newpage
\begin{anecdotebox}{GPT-4o full info underperforms as compared to som info on same trajectory (some=5.93 vs full=2.21)}{GPT4oColor}
\noindent\textbf{Category:} Full\_info confusion\par
\noindent\textbf{Model:} GPT-4o\par
\noindent\textbf{Configuration:} Environment: secretary. Latent: increasing\_position. Prompt and feedback: see sub-sections below. Trajectory: 4.\par\smallskip
\textbf{Setup.} On increasing\_position the optimal accept-position grows by one each task. Under \texttt{some\_info} GPT-4o scores 5.93 cumulative; under \texttt{full\_info} on the same trajectory it scores only 2.21. The verbose description of the latent does not produce an incrementing accept policy — instead it triggers premature accepts on early indices.\par\smallskip
\textbf{What went wrong.} Full\_info names the latent in plain English yet GPT-4o does not implement the corresponding policy. Each task is solved with the same heuristic, often accepting too early; cumulative reward is roughly half the some\_info baseline.\par\smallskip
\textbf{Per-task summary} (\textit{$\star$ marks tasks whose transcript is shown below}).\par
\par\smallskip\noindent\textbf{full\_info}\par
\noindent Agent task is to pick the maximum of 10 values revealed one at a time; on each turn the agent chooses [accept] (claim the current value and end the task) or [continue] (skip and see the next).\par
\noindent\textit{Hidden latent: max position grows by 1 each task (task k -\textgreater{} position k)}\par\smallskip
\noindent\footnotesize
\begin{tabularx}{\linewidth}{@{}>{\bfseries}p{1.6cm}>{\raggedright\arraybackslash}X>{\raggedright\arraybackslash}X>{\raggedright\arraybackslash}X>{\raggedright\arraybackslash}X>{\raggedright\arraybackslash}X>{\raggedright\arraybackslash}X>{\raggedright\arraybackslash}X>{\raggedright\arraybackslash}X>{\raggedright\arraybackslash}X>{\raggedright\arraybackslash}X@{}}
\toprule
Task & \textbf{$\star$1} & \textbf{2} & \textbf{3} & \textbf{4} & \textbf{5} & \textbf{$\star$6} & \textbf{7} & \textbf{8} & \textbf{$\star$9} & \textbf{10} \\
\midrule
Answer & 1.00 at 1 & 1.00 at 2 & 0.88 at 3 & 0.97 at 4 & 0.81 at 5 & 0.96 at 6 & 1.00 at 7 & 0.92 at 8 & 1.00 at 9 & 1.00 at 10 \\
Turns & 8 & 8 & 8 & 8 & 4 & 4 & 4 & 4 & 4 & 4 \\
Reward & 0.47 & 0.47 & 0.13 & 0.16 & 0.44 & 0.09 & 0.29 & 0.07 & 0.07 & 0.03 \\
\bottomrule
\end{tabularx}
\normalsize
\par\smallskip
\par\smallskip\noindent\textbf{some\_info}\par
\noindent Agent task is to pick the maximum of 10 values revealed one at a time; on each turn the agent chooses [accept] (claim the current value and end the task) or [continue] (skip and see the next).\par
\noindent\textit{Hidden latent: max position grows by 1 each task (task k -\textgreater{} position k)}\par\smallskip
\noindent\footnotesize
\begin{tabularx}{\linewidth}{@{}>{\bfseries}p{1.6cm}>{\raggedright\arraybackslash}X>{\raggedright\arraybackslash}X>{\raggedright\arraybackslash}X>{\raggedright\arraybackslash}X>{\raggedright\arraybackslash}X>{\raggedright\arraybackslash}X>{\raggedright\arraybackslash}X>{\raggedright\arraybackslash}X>{\raggedright\arraybackslash}X>{\raggedright\arraybackslash}X@{}}
\toprule
Task & \textbf{$\star$1} & \textbf{2} & \textbf{3} & \textbf{4} & \textbf{5} & \textbf{$\star$6} & \textbf{7} & \textbf{8} & \textbf{$\star$9} & \textbf{10} \\
\midrule
Answer & 1.00 at 1 & 1.00 at 2 & 0.88 at 3 & 0.97 at 4 & 0.81 at 5 & 0.96 at 6 & 1.00 at 7 & 0.92 at 8 & 1.00 at 9 & 1.00 at 10 \\
Turns & 8 & 2 & 10 & 4 & 10 & 6 & 5 & 10 & 7 & 2 \\
Reward & 0.47 & 1.00 & 0.44 & 1.00 & 0.23 & 1.00 & 0.47 & 0.38 & 0.49 & 0.45 \\
\bottomrule
\end{tabularx}
\normalsize
\par\smallskip
\end{anecdotebox}

\newpage

\begin{anecdotebox}{Gemini on wordladder/hub\_word\_4letter, trajectory 2: full\_info collapses (some=4.85 vs full=2.63) - tries to go for the latent without being confident}{GeminiColor}
\noindent\textbf{Category:} Full\_info confusion\par
\noindent\textbf{Model:} Gemini 2.5 Flash\par
\noindent\textbf{Configuration:} Environment: wordladder. Latent: hub\_word\_4letter. Prompt and feedback: see sub-sections below. Trajectory: 2.\par\smallskip
\textbf{Setup.} On the same trajectory~2, Gemini's cumulative reward drops from 4.85 (some\_info) to 2.63 (full\_info). Per-task rewards under full\_info: \texttt{[0.91, 0.50, 0.25, 0.97, 0.00]}. The verbose hub-word framing destabilizes word selection.\par\smallskip
\textbf{What went wrong.} Gemini under full\_info submits words that fit the hub-word narrative but break the one-letter-change rule, accumulates invalid moves, and ends task~5 at zero reward.\par\smallskip
\textbf{Per-task summary} (\textit{$\star$ marks tasks whose transcript is shown below}).\par
\par\smallskip\noindent\textbf{full\_info}\par
\noindent Agent task is to transform a given start word into a given target word (4-letter words) by changing exactly one letter per turn; every intermediate word must be a valid English word.\par
\noindent\textit{Hidden latent: all optimal paths pass through hub word = cork}\par\smallskip
\noindent\footnotesize
\begin{tabularx}{\linewidth}{@{}>{\bfseries}p{1.6cm}>{\raggedright\arraybackslash}X>{\raggedright\arraybackslash}X>{\raggedright\arraybackslash}X>{\raggedright\arraybackslash}X>{\raggedright\arraybackslash}X@{}}
\toprule
Task & \textbf{$\star$1} & \textbf{2} & \textbf{$\star$3} & \textbf{4} & \textbf{$\star$5} \\
\midrule
Answer & cord to farm & care to lock & boot to care & hand to work & boat to worm \\
Turns & 7 & 21 & 21 & 6 & 21 \\
Reward & 0.91 & 0.50 & 0.25 & 0.97 & 0.00 \\
\bottomrule
\end{tabularx}
\normalsize
\par\smallskip
\par\smallskip\noindent\textbf{some\_info}\par
\noindent Agent task is to transform a given start word into a given target word (4-letter words) by changing exactly one letter per turn; every intermediate word must be a valid English word.\par
\noindent\textit{Hidden latent: all optimal paths pass through hub word = cork}\par\smallskip
\noindent\footnotesize
\begin{tabularx}{\linewidth}{@{}>{\bfseries}p{1.6cm}>{\raggedright\arraybackslash}X>{\raggedright\arraybackslash}X>{\raggedright\arraybackslash}X>{\raggedright\arraybackslash}X>{\raggedright\arraybackslash}X@{}}
\toprule
Task & \textbf{$\star$1} & \textbf{2} & \textbf{$\star$3} & \textbf{4} & \textbf{$\star$5} \\
\midrule
Answer & cord to farm & care to lock & boot to care & hand to work & boat to worm \\
Turns & 4 & 8 & 9 & 4 & 5 \\
Reward & 1.00 & 0.94 & 0.91 & 1.00 & 1.00 \\
\bottomrule
\end{tabularx}
\normalsize
\par\smallskip
\end{anecdotebox}

\newpage
\begin{anecdotebox}{Full info}{GeminiColor}
\par\smallskip\noindent\textit{Showing Task 1, Task 3 and Task 5 out of 5 tasks.}\par\smallskip
\setlength{\columnsep}{6pt}
\begin{multicols}{3}\raggedcolumns
\begin{taskbox}{Task 1}{GeminiColor}
\thinkmsg{1}{**Discovering the Hub Word** {\\\relax}\  {\\\relax}\ I am now focusing on finding the "hub" word that links "cord" to "farm". The user wants me to discover it for the game. I think I am getting closer to the solution; if I can identify the pattern, this should be easier. {\\\relax}\  {\\\relax}\  {\\\relax}\ **Testing the Hub Hypothesis** {\\\relax}\  {\\\relax}\ I am now testing whether "form" functions as the hub word for this word ladder. The path is direct: "cord" to "corn" to "form," and then "form" to "farm." It seems quite efficient, at only three steps, which confirms its suitability as a potential intermediary. Now, I need to check how it will function within the broader sequence.}
\agentmsg{1}{[corn]}
\gmmsg{Nice! Keep going. {\\\relax}\ Word Ladder History: cord -\textgreater{} corn.  Target Word: farm}
\par\smallskip\noindent\textcolor{gray}{\scriptsize\itshape \dots\ continues (showing turns 1--1 of 7)}\par
\end{taskbox}
\columnbreak
\begin{taskbox}{Task 3}{GeminiColor}
\thinkmsg{1}{**Mapping the Path** {\\\relax}\  {\\\relax}\ I'm working on the word ladder, going from "boot" to "care" in Game 3. Starting with 'boot' and aiming for 'care', I'm planning my moves one letter at a time, aiming for optimal steps. The "boot" hint is guiding my choices! {\\\relax}\  {\\\relax}\  {\\\relax}\ **Identifying the Hub Word** {\\\relax}\  {\\\relax}\ I'm now focusing on identifying the intermediate "hub" word, drawing on previous games for clues. Game 1's path, `cord -\textgreater{} care -\textgreater{} farm`, suggests 'care'.  Game 2 has me starting at 'care' to reach 'lock'. That feels like my path may have a starting point and an ending point. I now suspect 'care' serves as the hub, and I'm developing a strategy to use this assumption in Game 3.}
\agentmsg{1}{[boar]}
\gmmsg{You attempted an invalid move. Reason: `boar` is not exactly one letter different from `boot`.}
\par\smallskip\noindent\textcolor{gray}{\scriptsize\itshape \dots\ continues (showing turns 1--1 of 21)}\par
\end{taskbox}
\columnbreak
\begin{taskbox}{Task 5}{GeminiColor}
\thinkmsg{1}{**Considering the Constraints** {\\\relax}\  {\\\relax}\ I'm focusing on the Word Ladder problem, specifically the 'boat' to 'worm' challenge. I'm reminded that solutions often converge at a crucial "hub" word. I'm actively seeking this hub, as the constraints dictate a one-letter change per step progression. This hub strategy will be central to resolving the final game in the series. {\\\relax}\  {\\\relax}\  {\\\relax}\ **Analyzing Past Word Ladders** {\\\relax}\  {\\\relax}\ I'm now diving into the previous games to spot potential "hub" words. While 'care' appeared in the initial game and as a target, I don't feel confident that it's a central hub word. It seems I didn't identify a clear hub in the initial word ladder. I'm reviewing my attempt to find some recurring themes in order to create the series-ending path.}
\agentmsg{1}{[boar]}
\gmmsg{Nice! Keep going. {\\\relax}\ Word Ladder History: boat -\textgreater{} boar.  Target Word: worm}
\par\smallskip\noindent\textcolor{gray}{\scriptsize\itshape \dots\ continues (showing turns 1--1 of 21)}\par
\end{taskbox}
\end{multicols}
\end{anecdotebox}

\newpage
\begin{anecdotebox}{Some Info}{GeminiColor}
\par\smallskip\noindent\textit{Showing Task 1, Task 3 and Task 5 out of 5 tasks.}\par\smallskip
\setlength{\columnsep}{6pt}
\begin{multicols}{3}\raggedcolumns
\begin{taskbox}{Task 1}{GeminiColor}
\thinkmsg{1}{**Initiating Word Transformation** {\\\relax}\  {\\\relax}\ I'm working on the Word Ladder game now. Starting with 'cord', I'm aiming for 'farm'. Considering the rules, my first step is to get closer by changing one letter. I've considered some options, and I think I'll change the first letter. {\\\relax}\  {\\\relax}\  {\\\relax}\ **Calculating Optimal Paths** {\\\relax}\  {\\\relax}\ I've evaluated several paths, focusing on single-letter changes to move from 'cord' to 'farm'. I've identified the shortest route: CORD, FORD, FORM, FARM, and validated it. The key was prioritizing valid English words with each transformation. I'm satisfied with this solution.}
\agentmsg{1}{[ford]}
\gmmsg{Nice! Keep going. {\\\relax}\ Word Ladder History: cord -\textgreater{} ford.  Target Word: farm}
\par\smallskip\noindent\textcolor{gray}{\scriptsize\itshape \dots\ continues (showing turns 1--1 of 4)}\par
\end{taskbox}
\columnbreak
\begin{taskbox}{Task 3}{GeminiColor}
\thinkmsg{1}{**Mapping the Path** {\\\relax}\  {\\\relax}\ I've begun mapping a word path from "boot" to "care," focusing on single-letter changes. My current path looks like this: boot -\textgreater{} coot. I'm actively brainstorming the next word to progress towards the target "care", while ensuring the changes are valid English words. {\\\relax}\  {\\\relax}\  {\\\relax}\ **Refining the Word Path** {\\\relax}\  {\\\relax}\ I've discarded the initial, lengthy path and am now focusing on "boot -\textgreater{} moot -\textgreater{} root -\textgreater{} soot -\textgreater{} shot -\textgreater{} chat -\textgreater{} cart -\textgreater{} care." The strategy has shifted towards minimizing steps. The current path is eight steps, which looks promising. Now, I'm analyzing it for potential shortcuts and optimizations, keeping an eye out for ways to reach "care" more directly.}
\agentmsg{1}{[moot]}
\gmmsg{Nice! Keep going. {\\\relax}\ Word Ladder History: boot -\textgreater{} moot.  Target Word: care}
\par\smallskip\noindent\textcolor{gray}{\scriptsize\itshape \dots\ continues (showing turns 1--1 of 9)}\par
\end{taskbox}
\columnbreak
\begin{taskbox}{Task 5}{GeminiColor}
\thinkmsg{1}{**Charting a Path Forward** {\\\relax}\  {\\\relax}\ I've got the start word "boat" and the end word "worm". So far, I've got a good start: boat -\textgreater{} moat. Now, I'm thinking, what's the next logical step to get closer to "worm"? I'm looking for a valid word that changes only one letter from "moat". I'm considering all possibilities. {\\\relax}\  {\\\relax}\  {\\\relax}\ **Refining the Approach** {\\\relax}\  {\\\relax}\ I'm back to the initial step, boat to moat. Now, I'm stuck again! I'm struggling to get from "moat" to "worm" in single-letter steps. I need a word between "moat" and "worm". I'm rethinking my strategy. Maybe I should consider other words from "boat."}
\agentmsg{1}{[moat]}
\gmmsg{Nice! Keep going. {\\\relax}\ Word Ladder History: boat -\textgreater{} moat.  Target Word: worm}
\par\smallskip\noindent\textcolor{gray}{\scriptsize\itshape \dots\ continues (showing turns 1--1 of 5)}\par
\end{taskbox}
\end{multicols}
\end{anecdotebox}

\newpage

\begin{anecdotebox}{Claude on bandits/ping\_pong, trajectory 0: full\_info hurts (some=6.65 vs full=4.31)}{ClaudeColor}
\noindent\textbf{Category:} Full\_info confusion\par
\noindent\textbf{Model:} Claude Sonnet 4.6\par
\noindent\textbf{Configuration:} Environment: bandits. Latent: ping\_pong. Prompt and feedback: see sub-sections below. Trajectory: 0.\par\smallskip
\textbf{Setup.} Claude on \texttt{traj\_0000}: cumulative reward drops from 6.65 (some\_info) to 4.31 (full\_info). The full description of the alternation latent leads Claude to commit to a single arm rather than rotate.\par\smallskip
\textbf{What went wrong.} Verbose alternation reasoning translates into over-commitment, not into a rotating policy. Cumulative reward is well below the some\_info baseline.\par\smallskip
\textbf{Per-task summary} (\textit{$\star$ marks tasks whose transcript is shown below}).\par
\par\smallskip\noindent\textbf{full\_info}\par
\noindent Agent task is to identify the button with the highest reward probability among 5 buttons (red, blue, green, yellow, purple); each turn the agent may press a button to observe a 0/1 reward, or commit via [select \textless{}button\textgreater{}].\par
\noindent\textit{Hidden latent: best button alternates: red (first) -- purple (last)}\par\smallskip
\noindent\footnotesize
\begin{tabularx}{\linewidth}{@{}>{\bfseries}p{1.6cm}>{\raggedright\arraybackslash}X>{\raggedright\arraybackslash}X>{\raggedright\arraybackslash}X>{\raggedright\arraybackslash}X>{\raggedright\arraybackslash}X>{\raggedright\arraybackslash}X>{\raggedright\arraybackslash}X>{\raggedright\arraybackslash}X>{\raggedright\arraybackslash}X>{\raggedright\arraybackslash}X@{}}
\toprule
Task & \textbf{$\star$1} & \textbf{2} & \textbf{3} & \textbf{4} & \textbf{5} & \textbf{$\star$6} & \textbf{7} & \textbf{$\star$8} & \textbf{9} & \textbf{10} \\
\midrule
Answer & red & purple & red & purple & red & purple & red & purple & red & purple \\
Turns & 9 & 17 & 10 & 12 & 10 & 11 & 14 & 12 & 4 & 13 \\
Reward & 0.86 & 0.00 & 0.85 & 0.00 & 0.85 & 0.00 & 0.00 & 0.00 & 0.94 & 0.80 \\
\bottomrule
\end{tabularx}
\normalsize
\par\smallskip
\par\smallskip\noindent\textbf{some\_info}\par
\noindent Agent task is to identify the button with the highest reward probability among 5 buttons (red, blue, green, yellow, purple); each turn the agent may press a button to observe a 0/1 reward, or commit via [select \textless{}button\textgreater{}].\par
\noindent\textit{Hidden latent: best button alternates: red (first) -- purple (last)}\par\smallskip
\noindent\footnotesize
\begin{tabularx}{\linewidth}{@{}>{\bfseries}p{1.6cm}>{\raggedright\arraybackslash}X>{\raggedright\arraybackslash}X>{\raggedright\arraybackslash}X>{\raggedright\arraybackslash}X>{\raggedright\arraybackslash}X>{\raggedright\arraybackslash}X>{\raggedright\arraybackslash}X>{\raggedright\arraybackslash}X>{\raggedright\arraybackslash}X>{\raggedright\arraybackslash}X@{}}
\toprule
Task & \textbf{$\star$1} & \textbf{2} & \textbf{3} & \textbf{4} & \textbf{5} & \textbf{$\star$6} & \textbf{7} & \textbf{$\star$8} & \textbf{9} & \textbf{10} \\
\midrule
Answer & red & purple & red & purple & red & purple & red & purple & red & purple \\
Turns & 9 & 10 & 10 & 15 & 20 & 11 & 19 & 8 & 11 & 11 \\
Reward & 0.86 & 0.85 & 0.85 & 0.00 & 0.70 & 0.83 & 0.00 & 0.88 & 0.83 & 0.83 \\
\bottomrule
\end{tabularx}
\normalsize
\end{anecdotebox}

\newpage
\begin{anecdotebox}{Full Info}{ClaudeColor}
\setlength{\columnsep}{6pt}
\begin{multicols}{3}\raggedcolumns
\begin{taskbox}{Task 1}{ClaudeColor}
\agentmsg{1}{I'll start by exploring each button to gather initial data, then focus on the most promising ones. {\\\relax}\  {\\\relax}\ Let me explore all 5 buttons first to get a baseline. {\\\relax}\  {\\\relax}\ [red]}
\gmmsg{You pressed the red button and received a reward of 1.0.}
\thinkmsg{2}{Good start! Red gave a reward. Let me explore the others.}
\agentmsg{2}{Good start with red! Let me try the others. {\\\relax}\  {\\\relax}\ [blue]}
\gmmsg{You pressed the blue button and received a reward of 1.0.}
\par\smallskip\noindent\textcolor{gray}{\scriptsize\itshape \dots\ continues (showing turns 1--2 of 9)}\par
\end{taskbox}
\columnbreak
\begin{taskbox}{Task 6}{ClaudeColor}
\thinkmsg{1}{Pattern so far: {\\\relax}\ - Game 1: Red \checkmark{} {\\\relax}\ - Game 2: not yellow {\\\relax}\ - Game 3: Red \checkmark{} {\\\relax}\ - Game 4: not green {\\\relax}\ - Game 5: Red \checkmark{} {\\\relax}\  Interesting - odd games (1, 3, 5) are all red. Even games (2, 4) are unknown. Game 6 is even. Let me think about what the pattern might be for even games.  Game 2: not yellow  Game 4: not green  Game 6: maybe blue or purple?  The hint says "The best color changes in a predictable pattern across games." Let me think about this differently. Maybe the pattern cycles through colors in order: red, blue, green, yellow, purple, red, blue, green, yellow, purple...    Wait, that doesn't work since red keeps appearing in odd positions. Let me reconsider - maybe odd games are always red, and even games cycle through the other colors. So game 2 could be blue, green, or purple (not yellow), game 4 could be blue, yellow, or purple (not green), and game 6 would be one of the remaining options. I should focus on testing blue and purple for game 6 since those seem most likely. I'm realizing I don't have enough data to pin down the pattern yet, so I should just test the remaining options in game 6 systematically---starting with blue and purple since those are still viable candidates.}
\par\smallskip\noindent\textcolor{gray}{\scriptsize\itshape \dots\ after 9 more turns \dots}\par\smallskip
\agentmsg{11}{Excellent! Green is 4/4 (100\%)! Very compelling evidence.  Updated scores: - Red: 0/1 (0\%)  - Blue: 0/1 (0\%)  - Green: 4/4 (100\%)  - Yellow: 0/1 (0\%) - Purple: 2/3 (67\%)  Green is clearly dominant here!  [select green]}
\gmmsg{
Wrong!
}
\end{taskbox}
\columnbreak
\begin{taskbox}{Task 8}{ClaudeColor}
\thinkmsg{1}{Let me reconsider the pattern. Looking at what I know: {\\\relax}\ - Game 1: Red \checkmark{}  - Game 2: Not yellow  - Game 3: Red \checkmark{} - Game 4: Not green - Game 5: Red \checkmark{} - Game 6: Not green - Game 7: Not blue  Wait, the problem says "The best color changes in a predictable pattern across games." But I've been wrong about the even games. Let me think harder The odd games (1, 3, 5) were RED.  But game 7 was also wrong, which breaks my assumption that odd games are always red. Maybe the pattern cycles through a set of colors rather than alternating between two. If it's something like red, blue, green, yellow repeating, then game 1 would be red, game 2 blue, game 3 green---but that doesn't match what I know either. I need to reconsider whether red was actually correct for games 1, 3, and 5, or if I just happened to guess right those times. I'm realizing game 7 breaks the odd-equals-red pattern since I guessed blue and was wrong. Let me test a simpler alternating hypothesis: red, blue, red, blue, red, blue, red, blue. .....  OK so possible patterns:  1. Alternating Red/Blue: R, B, R, B, R, B, R, B, R, B -\textgreater{} Game 8: Blue  2. All odd = Red ..... I'm confident in the even/odd pattern I've identified, so I'll commit to Blue for game 8.}
\agentmsg{1}{
[blue]}
\par\smallskip\noindent\textcolor{gray}{\scriptsize\itshape \dots\ after 10 more turns \dots}\par\smallskip
\agentmsg{12}{Updated scores:  - Red: 1/2 (50\%)  - Blue: 1/2 (50\%)  - Green: 0/1 (0\%)  - Yellow: 3/4 (75\%) - Purple: 1/2 (50\%). Yellow is clearly the strongest at 75\% with 4 samples! 
[select yellow]}
\gmmsg{Wrong!
}
\end{taskbox}
\end{multicols}
\end{anecdotebox}
\begin{anecdotebox}{Some Info}{ClaudeColor}
\setlength{\columnsep}{6pt}
\begin{multicols}{3}\raggedcolumns
\begin{taskbox}{Task 1}{ClaudeColor}
\agentmsg{1}{I'll start by exploring each button to get initial data, then focus on the most promising ones. {\\\relax}\  {\\\relax}\ [red]}
\gmmsg{You pressed the red button and received a reward of 1.0.}
\thinkmsg{2}{Good start! Red gave a reward. Let me explore the other buttons.}
\agentmsg{2}{Good start! Red gave a reward. Let me explore the others. {\\\relax}\  {\\\relax}\ [blue]}
\gmmsg{You pressed the blue button and received a reward of 1.0.}
\par\smallskip\noindent\textcolor{gray}{\scriptsize\itshape \dots\ continues (showing turns 1--2 of 9)}\par
\end{taskbox}
\columnbreak
\begin{taskbox}{Task 6}{ClaudeColor}
\thinkmsg{1}{Let me review what I've learned so far:  - Episode 1: Red won - Episode 2: Purple won  - Episode 3: Red won  - Episode 4: Wrong (selected green, likely red was correct) - Episode 5: Red won.  Red seems to be the dominant button. Let me start with red and see if I can confirm quickly.}
\agentmsg{1}{Looking at the pattern: Red has won episodes 1, 3, 5 - it seems to be the dominant button. Let me start with red. {\\\relax}\  {\\\relax}\ [red]}
\gmmsg{You pressed the red button and received a reward of 0.0.}
\thinkmsg{2}{Red started with a loss this time. Let me explore the others.}
\agentmsg{2}{Red missed this time. Let me explore the others. {\\\relax}\  {\\\relax}\ [blue]}
\gmmsg{You pressed the blue button and received a reward of 0.0.}
\par\smallskip\noindent\textcolor{gray}{\scriptsize\itshape \dots\ continues (showing turns 1--2 of 11)}\par
\end{taskbox}
\columnbreak
\begin{taskbox}{Task 8}{ClaudeColor}
\thinkmsg{1}{Let me think about the pattern more carefully. The pattern seems to be: Episode 1: Red {\\\relax}\ Episode 2: Purple {\\\relax}\ Episode 3: Red {\\\relax}\ Episode 4: ? (not green)  Episode 5: Red {\\\relax}\ Episode 6: Purple {\\\relax}\ Episode 7: ? (not yellow) {\\\relax}\   Maybe the pattern alternates between two buttons, or maybe there's a specific sequence. Looking at the odd episodes, they're consistently red, while the even episodes I've confirmed are purple. But episodes 4 and 7 don't fit this simple alternation. Maybe blue works for episode 4, and green for episode 7 since it had better odds than yellow. I should approach this more systematically instead of guessing---the winning button likely has a very high success rate that I'm just missing when I don't test it early enough.  For episode 8, I'll explore more carefully and only commit to a button once I have solid evidence like 4+ successes, rather than selecting when the data is still ambiguous.}
\agentmsg{1}{ Let me explore carefully.  [red]}
\par\smallskip\noindent\textcolor{gray}{\scriptsize\itshape \dots\ after 6 more turns \dots}\par\smallskip
\thinkmsg{8}{Purple is 3/3 (100\%)! I'm very confident. Let me select purple now.}
\agentmsg{8}{[select purple]}
\gmmsg{You selected 'purple' as your final answer on turn 8. Correct!}
\end{taskbox}
\end{multicols}
\end{anecdotebox}

\newpage

\end{appendix}

\ifdefined\useorstyle




\else
\newpage

\appendix

\fi


\end{document}